\pdfoutput=1
\RequirePackage[T1]{fontenc}
\documentclass[10pt,journal,compsoc]{IEEEtran}
\usepackage{amsmath,amsfonts,amssymb}
\usepackage{array}
\usepackage[nocompress]{cite}
\usepackage[caption=false,font=normalsize,labelfont=sf,textfont=sf]{subfig}
\usepackage{textcomp}
\usepackage{stfloats}
\usepackage{url}
\usepackage{verbatim}
\usepackage{graphicx}
\usepackage{placeins}
\usepackage{booktabs}
\usepackage{tabularx}
\usepackage{longtable}
\usepackage{multicol}
\usepackage[table]{xcolor}
\usepackage[edges]{forest}
\definecolor{famO}{HTML}{0072B2}
\definecolor{famAP}{HTML}{E69F00}
\definecolor{famR}{HTML}{009E73}
\definecolor{famTV}{HTML}{CC79A7}
\definecolor{famH}{HTML}{6E6E6E}
\providecommand{\rrmfamily}[1]{\textbf{#1}}
\newsavebox{\rrmfindingsbox}
\newenvironment{rrmfindings}{%
  \par\addvspace{4pt}%
  \setlength{\fboxsep}{4pt}%
  \begin{lrbox}{\rrmfindingsbox}%
  \begin{minipage}{\dimexpr\columnwidth-2\fboxsep\relax}%
  \textbf{Findings.}%
  \begin{list}{\textbullet}{%
    \setlength{\leftmargin}{0.9em}\setlength{\labelwidth}{0.5em}%
    \setlength{\labelsep}{0.4em}\setlength{\itemsep}{1pt}%
    \setlength{\parsep}{0pt}\setlength{\topsep}{1pt}%
    \setlength{\partopsep}{0pt}}}%
  {\end{list}\end{minipage}\end{lrbox}%
  \noindent\colorbox{black!5}{\usebox{\rrmfindingsbox}}\par\addvspace{4pt}}
\newsavebox{\rrmpromptbox}
\newcommand{\rrmtok}[1]{{\ttfamily\fontsize{7}{9.5}\selectfont #1}}
\newcommand{\rrmpromptrule}[2]{\rlap{\color{black!5}\vrule width\columnwidth
  height #1 depth\dimexpr #2+0.3pt\relax}}
\newcommand{\rrmpromptpad}[1]{\nointerlineskip
  \setbox0\hbox to \columnwidth{\rrmpromptrule{#1}{0pt}\hss}\dp0=0pt \box0 }
\newcommand{\rrmpromptgap}{%
  \setbox0\vbox to 2.5pt{\hbox to \columnwidth{\rrmpromptrule{2.5pt}{0pt}\hss}\vss}%
  \cleaders\box0\vskip 2.5pt }
\newcommand{\rrmpromptlinepen}{100}
\newcommand{\rrmpromptkeep}{\def\rrmpromptlinepen{10000}}
\newcount\rrmpromptline
\newcommand{\rrmpromptpar}[1]{%
  \rrmpromptsep
  \setbox\rrmpromptbox\vbox{\hsize=\dimexpr\columnwidth-8pt\relax
    \parindent=0pt \parskip=0pt \leftskip=0pt \rightskip=0pt
    \lineskiplimit=-\maxdimen \noindent\strut #1\par}%
  \splittopskip=\ht\strutbox \splitmaxdepth=\dp\strutbox
  \rrmpromptline=0
  \loop\ifvoid\rrmpromptbox\else
    \setbox0\vsplit\rrmpromptbox to \ht\strutbox
    \ht0=\ht\strutbox \dp0=\dp\strutbox
    \advance\rrmpromptline by 1
    \setbox0\hbox to \columnwidth{%
      \rrmpromptrule{\ht\strutbox}{\dp\strutbox}\hskip4pt\box0\hss}%
    \dp0=\dp\strutbox
    \nointerlineskip\box0
    \ifvoid\rrmpromptbox\else
      \ifnum\rrmpromptline=1 \nobreak
      \else\ifdim\ht\rrmpromptbox>\ht\strutbox \penalty\rrmpromptlinepen\relax
      \else \nobreak \fi\fi
    \fi
  \repeat
  \def\rrmpromptlinepen{100}%
  \def\rrmpromptsep{\rrmpromptgap}}
\newenvironment{rrmprompt}[1]{%
  \par\addvspace{5pt plus 3pt minus 1.5pt}%
  \footnotesize\vbadness=10000 \vfuzz=\maxdimen
  \rrmpromptpad{4pt}%
  \def\rrmpromptsep{\nobreak}%
  \rrmpromptpar{\textbf{#1.}}%
  \def\rrmpromptsep{\nobreak\rrmpromptgap}}%
  {\nobreak\rrmpromptpad{4pt}\par\addvspace{5pt plus 3pt minus 1.5pt}}
\makeatletter
\def\paragraph{\@startsection{paragraph}{4}{\z@}{0ex plus 0.1ex minus 0.1ex}%
  {0ex}{\normalfont\normalsize\itshape}}
\makeatother
\newcolumntype{L}[1]{>{\raggedright\arraybackslash}p{#1}}
\newcolumntype{Y}{>{\raggedright\arraybackslash}X}
\def\BibTeX{{\rm B\kern-.05em{\sc i\kern-.025em b}\kern-.08em
    T\kern-.1667em\lower.7ex\hbox{E}\kern-.125emX}}
\usepackage{balance}
\usepackage{etoolbox}
\makeatletter
\newcommand{\rrmheadingbreaks}{\raggedright\hyphenpenalty\@M\exhyphenpenalty\@M}
\patchcmd{\section}{\scshape}{\scshape\rrmheadingbreaks}{}%
  {\PackageWarning{main}{section heading patch failed}}
\patchcmd{\subsection}{\sffamily\bfseries}{\sffamily\bfseries\rrmheadingbreaks}{}%
  {\PackageWarning{main}{subsection heading patch failed}}
\patchcmd{\subsubsection}{\sffamily\itshape}{\sffamily\itshape\rrmheadingbreaks}{}%
  {\PackageWarning{main}{subsubsection heading patch failed}}
\makeatother
\usepackage[hidelinks]{hyperref}
\newif\ifRRMColorLinks
\RRMColorLinkstrue
\definecolor{RRMCiteColor}{HTML}{3E5C76}
\definecolor{RRMRefColor}{HTML}{6D5B72}
\ifRRMColorLinks
  \hypersetup{colorlinks=true,
    citecolor=RRMCiteColor,linkcolor=RRMRefColor,urlcolor=RRMCiteColor}
\fi
\hypersetup{
  pdftitle={Robotic Reward Models in the Foundation-Model Era: A Survey and
    Capability Benchmark},
  pdfauthor={Guoqiang Liang, Yiqi Lin, Shangchen Zhou, Wenzheng Zeng,
    Yanzhe Chen, Zechen Bai, Mike Zheng Shou},
}
\begin{document}
\bstctlcite{IEEEexample:BSTcontrol}

\title{Robotic Reward Models in the Foundation-Model Era: A Survey and Capability Benchmark}

\author{Guoqiang Liang$^{*1}$ \quad Yiqi Lin$^{*1}$ \quad Shangchen Zhou$^{2}$ \quad Wenzheng Zeng$^{1}$ \\ \quad Yanzhe Chen$^{1}$ \quad Zechen Bai$^{1}$  \quad Mike Zheng Shou$^{\dagger1}$ \\
$^{1}$National University of Singapore 
\quad
$^{2}$Nanyang Technological University \\
\thanks{$^*$ Equal contribution}
\thanks{$^\dagger$ Corresponding author}
}
\markboth{Robotic Reward Models in the Foundation-Model Era}%
{Liang \MakeLowercase{\textit{et al.}}: Robotic Reward Models in the Foundation-Model Era}

\IEEEtitleabstractindextext{%
\begin{abstract}

Robotic reward models (RRMs) use foundation models to replace per-task reward engineering with reusable, task-conditioned evaluation. 
Yet under one name, their outputs answer different questions, from whether a task succeeded to which execution is better.
This survey organizes 114 methods into five semantic families by the question their native outputs answer: Outcome, Absolute Progress, Relative, Temporal Value, and Hybrid. 
Within this taxonomy, we review how model design determines what an evaluator can judge, how raw experience becomes supervision, and how existing evaluation practice tests the result. 
The review exposes a gap: agreement with demonstration order and downstream policy gains do not show whether a reward tracks task fulfillment. 
We therefore introduce RRM-Bench-1K, 1,117 real-robot videos of closed-loop executions with task-state and camera-evidence labels, and four contrast suites that separate the credited condition from its usual cues. 
A unified study of 27 evaluators shows that the strongest final scores rank whole executions well enough for selection and filtering. 
Dense credit, however, follows the progression of an execution more closely than its task state, and no evaluator signals when the decisive view is missing.
How much fidelity a reward model needs therefore depends on the decision it drives.
We outline directions from real-failure supervision to metrics that separate task state from chronology, and open-source RRM-Hub, a toolbox in which 14 existing RRMs can be readily applied and new evaluators tested on the same capabilities.
Project page: \url{https://ethanliang99.github.io/awesome-robot-reward-model}.
\end{abstract}

}
\maketitle
\IEEEdisplaynontitleabstractindextext

\begin{figure*}[!t]
\centering
    \includegraphics[width=1\linewidth]{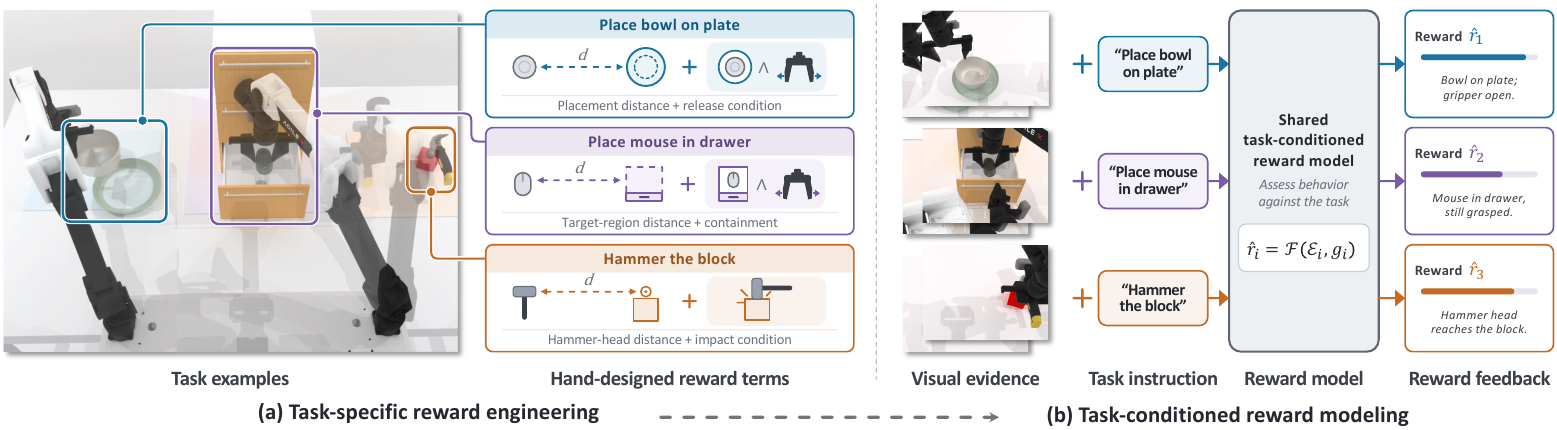}
    \vspace{-2mm}
    \caption{From per-task reward engineering to task-conditioned reward
modeling. (a)~Each task receives its own hand-designed reward, a distance
$d$ to the target plus a task-specific success condition, such as the bowl
resting on the plate with the gripper open. (b)~One shared evaluator
$\mathcal{F}$ maps visual evidence $\mathcal{E}_i$ and instruction $g_i$
to a reward $\hat{r}_i$ for every task.}
    \label{fig:teaser}
\vspace{-2mm}
\end{figure*}

\IEEEraisesectionheading{\section{Introduction}\label{sec:intro}}
\IEEEPARstart{F}{oundation} models let robot policies address more tasks and environments, as vision--language--action (VLA) models
such as $\pi_0$~\cite{DBLP:journals/corr/abs-2410-24164} and
$\pi_{0.5}$~\cite{DBLP:journals/corr/abs-2504-16054} illustrate. Every
task a policy attempts must also be judged: improving the policy,
selecting its data, and monitoring its execution all turn on whether the
robot's behavior satisfies the task. Conventional reward design supplies
this judgment task by task, with manually specified success criteria and
shaping terms~\cite{DBLP:conf/corl/YuQHJHFL19}
(Fig.~\ref{fig:teaser}(a)), an effort that grows with every task a
general policy adds. Robotic reward models (RRMs) replace it with
reusable, task-conditioned evaluators that use foundation models as
representations, supervisors, or judges to assess robot behavior against a
task specification (Fig.~\ref{fig:teaser}(b)).

Early RRMs derived rewards from pretrained foundation models.
ZSRM~\cite{pmlr-v162-mahmoudieh22a},
RoboCLIP~\cite{DBLP:conf/nips/SontakkeZAPBSFI23}, and
VLM-RM~\cite{DBLP:conf/iclr/RocamondeMNPL24} score the compatibility
between observations and a task description or demonstration, and use it as a reward or threshold it into a binary goal reward.
RL-VLM-F~\cite{DBLP:conf/icml/WangSZXBHE24} distills pairwise judgments of
a vision--language model (VLM) into a task reward. Recent evaluators such
as ReWiND~\cite{DBLP:journals/corr/abs-2505-10911},
Robo-Dopamine~\cite{DBLP:journals/corr/abs-2512-23703},
RoboReward~\cite{DBLP:journals/corr/abs-2601-00675}, and
Robometer~\cite{DBLP:journals/corr/abs-2603-02115} are trained on
multi-task robot data with supervision from successful and failed
executions, partial completion, temporal relations, and trajectory
preferences. They now serve several uses: stage and progress
estimation (SARM~\cite{DBLP:journals/corr/abs-2509-25358}), detection of
planning and execution errors
(Guardian~\cite{DBLP:journals/corr/abs-2512-01946}), value estimation for
reinforcement learning (ViVa~\cite{DBLP:journals/corr/abs-2604-08168}), and
candidate-action scoring and selection
(RoVer~\cite{DBLP:journals/corr/abs-2510-10975}).

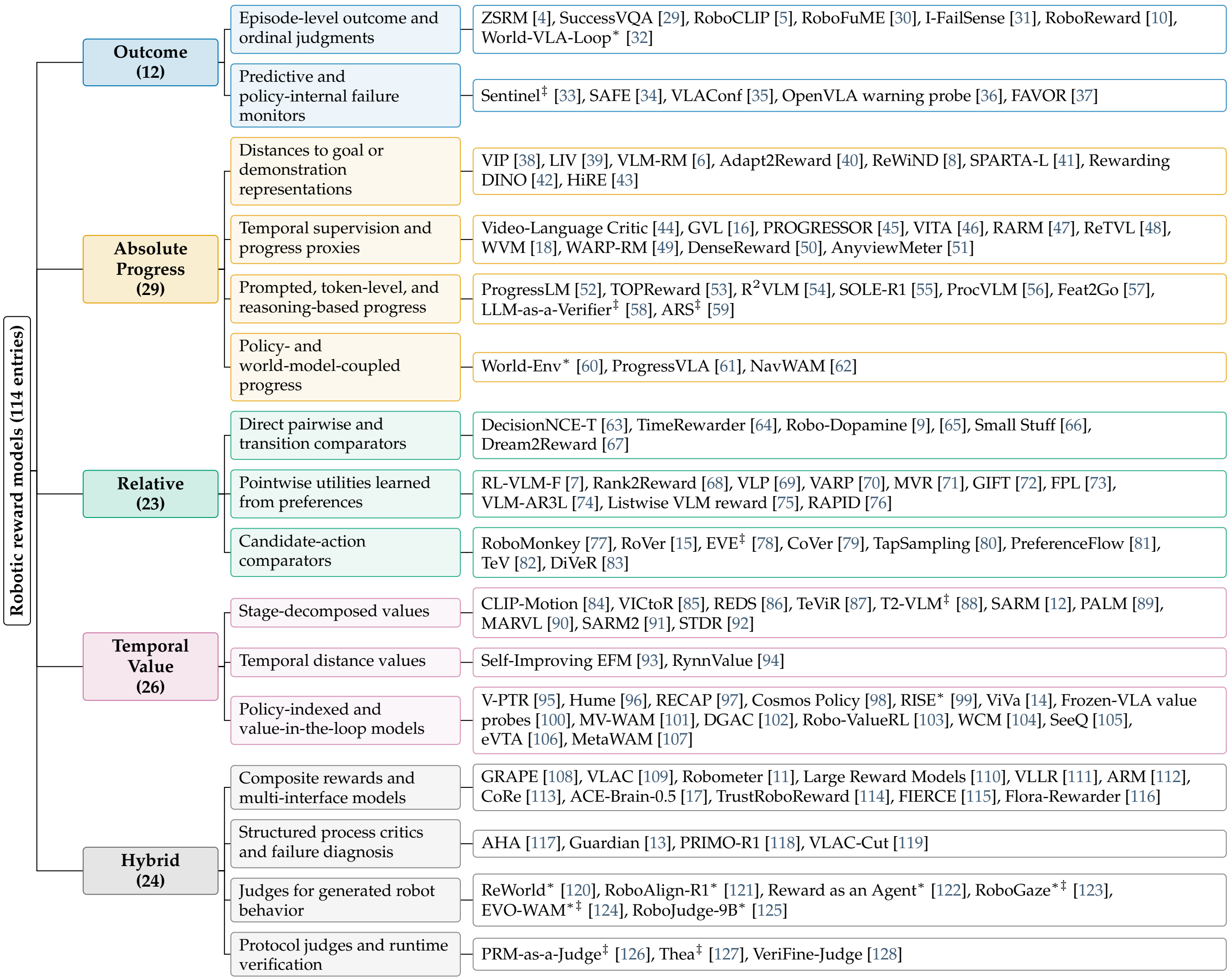
\begin{figure*}[!t]
\centering
\providecolor{famO}{HTML}{0072B2}
\providecolor{famAP}{HTML}{E69F00}
\providecolor{famR}{HTML}{009E73}
\providecolor{famTV}{HTML}{CC79A7}
\providecolor{famH}{HTML}{6E6E6E}
\begin{forest}
  forked edges,
  for tree={
    grow=east, reversed=true, anchor=base west,
    parent anchor=east, child anchor=west, base=center,
    draw, rounded corners=1.5pt, line width=0.4pt,
    font=\fontsize{7}{8.2}\selectfont, inner xsep=3.5pt, inner ysep=2.5pt,
    l sep=5pt, s sep=4pt, fork sep=2.5pt, edge={line width=0.4pt},
  },
  [{Robotic reward models (114 entries)}, rotate=90, anchor=center, child anchor=north, parent anchor=south, font=\fontsize{7.5}{8.6}\selectfont\bfseries
    [{\parbox{1.75cm}{\centering\fontsize{7.5}{8.6}\selectfont\bfseries Outcome\\(12)}}, fill=famO!18, draw=famO
      [{\parbox{3.15cm}{\raggedright Episode-level outcome and ordinal judgments}}, fill=famO!7, draw=famO!80
        [{\parbox{10.9cm}{\raggedright ZSRM~\cite{pmlr-v162-mahmoudieh22a}, SuccessVQA~\cite{du2023successvqa}, RoboCLIP~\cite{DBLP:conf/nips/SontakkeZAPBSFI23}, RoboFuME~\cite{robofume2023}, I-FailSense~\cite{DBLP:journals/corr/abs-2509-16072}, RoboReward~\cite{DBLP:journals/corr/abs-2601-00675}, World-VLA-Loop$^{\ast}$~\cite{DBLP:journals/corr/abs-2602-06508}}}, draw=famO!80]
      ]
      [{\parbox{3.15cm}{\raggedright Predictive and policy-internal failure monitors}}, fill=famO!7, draw=famO!80
        [{\parbox{10.9cm}{\raggedright Sentinel$^{\ddagger}$~\cite{agia2024sentinel}, SAFE~\cite{gu2025safe}, VLAConf~\cite{vlaconf2026}, OpenVLA warning probe~\cite{mahato2026openvlafailure}, FAVOR~\cite{favor2026}}}, draw=famO!80]
      ]
    ]
    [{\parbox{1.75cm}{\centering\fontsize{7.5}{8.6}\selectfont\bfseries Absolute Progress\\(29)}}, fill=famAP!18, draw=famAP
      [{\parbox{3.15cm}{\raggedright Distances to goal or demonstration representations}}, fill=famAP!7, draw=famAP!80
        [{\parbox{10.9cm}{\raggedright VIP~\cite{DBLP:conf/iclr/MaSJBK023}, LIV~\cite{DBLP:conf/icml/MaK0BJ23}, VLM-RM~\cite{DBLP:conf/iclr/RocamondeMNPL24}, Adapt2Reward~\cite{DBLP:conf/eccv/YangCQWWLGH24}, ReWiND~\cite{DBLP:journals/corr/abs-2505-10911}, SPARTA-L~\cite{sparta2025}, Rewarding DINO~\cite{DBLP:journals/corr/abs-2603-16978}, HiRE~\cite{hire2026}}}, draw=famAP!80]
      ]
      [{\parbox{3.15cm}{\raggedright Temporal supervision and progress proxies}}, fill=famAP!7, draw=famAP!80
        [{\parbox{10.9cm}{\raggedright Video-Language Critic~\cite{DBLP:journals/tmlr/AlakuijalaMWFKM25}, GVL~\cite{DBLP:conf/iclr/MaHFSLXK0DXBJ0Z25}, PROGRESSOR~\cite{DBLP:conf/iccv/AyalewZWJMW25}, VITA~\cite{vita2025}, RARM~\cite{DBLP:journals/corr/abs-2606-22027}, ReTVL~\cite{DBLP:journals/corr/abs-2606-24633}, WVM~\cite{DBLP:journals/corr/abs-2606-24742}, WARP-RM~\cite{DBLP:journals/corr/abs-2606-28320}, DenseReward~\cite{DBLP:journals/corr/abs-2607-13033}, AnyviewMeter~\cite{anyviewmeter2026}}}, draw=famAP!80]
      ]
      [{\parbox{3.15cm}{\raggedright Prompted, token-level, and reasoning-based progress}}, fill=famAP!7, draw=famAP!80
        [{\parbox{10.9cm}{\raggedright ProgressLM~\cite{progresslm2026}, TOPReward~\cite{DBLP:journals/corr/abs-2602-19313}, R$^{2}$VLM~\cite{r2vlm2026}, SOLE-R1~\cite{DBLP:journals/corr/abs-2603-28730}, ProcVLM~\cite{DBLP:journals/corr/abs-2605-08774}, Feat2Go~\cite{feat2go2026}, LLM-as-a-Verifier$^{\ddagger}$~\cite{llmverifier2026}, ARS$^{\ddagger}$~\cite{ars2026}}}, draw=famAP!80]
      ]
      [{\parbox{3.15cm}{\raggedright Policy- and world-model-coupled progress}}, fill=famAP!7, draw=famAP!80
        [{\parbox{10.9cm}{\raggedright World-Env$^{\ast}$~\cite{DBLP:journals/corr/abs-2509-24948}, ProgressVLA~\cite{DBLP:journals/corr/abs-2603-27670}, NavWAM~\cite{DBLP:journals/corr/abs-2606-13494}}}, draw=famAP!80]
      ]
    ]
    [{\parbox{1.75cm}{\centering\fontsize{7.5}{8.6}\selectfont\bfseries Relative\\(23)}}, fill=famR!18, draw=famR
      [{\parbox{3.15cm}{\raggedright Direct pairwise and transition comparators}}, fill=famR!7, draw=famR!80
        [{\parbox{10.9cm}{\raggedright DecisionNCE-T~\cite{decisionnce2024}, TimeRewarder~\cite{DBLP:journals/corr/abs-2509-26627}, Robo-Dopamine~\cite{DBLP:journals/corr/abs-2512-23703,robodopamine2_2026}, Small Stuff~\cite{smallstuff2026}, Dream2Reward~\cite{dream2reward2026}}}, draw=famR!80]
      ]
      [{\parbox{3.15cm}{\raggedright Pointwise utilities learned from preferences}}, fill=famR!7, draw=famR!80
        [{\parbox{10.9cm}{\raggedright RL-VLM-F~\cite{DBLP:conf/icml/WangSZXBHE24}, Rank2Reward~\cite{rank2reward2024}, VLP~\cite{vlp2025}, VARP~\cite{varp2025}, MVR~\cite{DBLP:journals/corr/abs-2603-01694}, GIFT~\cite{gift2026}, FPL~\cite{DBLP:journals/corr/abs-2606-32027}, VLM-AR3L~\cite{vlmar3l2026}, Listwise VLM reward~\cite{listwisevlm2026}, RAPID~\cite{rapid2026}}}, draw=famR!80]
      ]
      [{\parbox{3.15cm}{\raggedright Candidate-action comparators}}, fill=famR!7, draw=famR!80
        [{\parbox{10.9cm}{\raggedright RoboMonkey~\cite{DBLP:journals/corr/abs-2506-17811}, RoVer~\cite{DBLP:journals/corr/abs-2510-10975}, EVE$^{\ddagger}$~\cite{eve2025}, CoVer~\cite{DBLP:journals/corr/abs-2602-12281}, TapSampling~\cite{DBLP:journals/corr/abs-2605-25547}, PreferenceFlow~\cite{preferenceflow2026}, TeV~\cite{tev2026}, DiVeR~\cite{diver2026}}}, draw=famR!80]
      ]
    ]
    [{\parbox{1.75cm}{\centering\fontsize{7.5}{8.6}\selectfont\bfseries Temporal Value\\(26)}}, fill=famTV!18, draw=famTV
      [{\parbox{3.15cm}{\raggedright Stage-decomposed values}}, fill=famTV!7, draw=famTV!80
        [{\parbox{10.9cm}{\raggedright CLIP-Motion~\cite{clipmotion2023}, VICtoR~\cite{victor2024}, REDS~\cite{reds2025}, TeViR~\cite{tevir2025}, T2-VLM$^{\ddagger}$~\cite{t2vlm2025}, SARM~\cite{DBLP:journals/corr/abs-2509-25358}, PALM~\cite{DBLP:journals/corr/abs-2601-07060}, MARVL~\cite{DBLP:journals/corr/abs-2602-15872}, SARM2~\cite{DBLP:journals/corr/abs-2606-10305}, STDR~\cite{stdr2026}}}, draw=famTV!80]
      ]
      [{\parbox{3.15cm}{\raggedright Temporal distance values}}, fill=famTV!7, draw=famTV!80
        [{\parbox{10.9cm}{\raggedright Self-Improving EFM~\cite{ghasemipour2025selfimproving}, RynnValue~\cite{huang2026rynnvalue}}}, draw=famTV!80]
      ]
      [{\parbox{3.15cm}{\raggedright Policy-indexed and value-in-the-loop models}}, fill=famTV!7, draw=famTV!80
        [{\parbox{10.9cm}{\raggedright V-PTR~\cite{vptr2023}, Hume~\cite{DBLP:journals/corr/abs-2505-21432}, RECAP~\cite{physicalintelligence2025pistar}, Cosmos Policy~\cite{DBLP:journals/corr/abs-2601-16163}, RISE$^{\ast}$~\cite{DBLP:journals/corr/abs-2602-11075}, ViVa~\cite{DBLP:journals/corr/abs-2604-08168}, Frozen-VLA value probes~\cite{frozenvlaprobes2026}, MV-WAM~\cite{DBLP:journals/corr/abs-2606-21088}, DGAC~\cite{DBLP:journals/corr/abs-2606-21406}, Robo-ValueRL~\cite{xia2026robovaluerl}, WCM~\cite{fei2026wcm}, SeeQ~\cite{seeq2026}, eVTA~\cite{evta2026}, MetaWAM~\cite{liang2026metawam}}}, draw=famTV!80]
      ]
    ]
    [{\parbox{1.75cm}{\centering\fontsize{7.5}{8.6}\selectfont\bfseries Hybrid\\(24)}}, fill=famH!18, draw=famH
      [{\parbox{3.15cm}{\raggedright Composite rewards and multi-interface models}}, fill=famH!7, draw=famH!80
        [{\parbox{10.9cm}{\raggedright GRAPE~\cite{zhang2024grape}, VLAC~\cite{DBLP:journals/corr/abs-2509-15937}, Robometer~\cite{DBLP:journals/corr/abs-2603-02115}, Large Reward Models~\cite{DBLP:journals/corr/abs-2603-16065}, VLLR~\cite{DBLP:journals/corr/abs-2604-00055}, ARM~\cite{DBLP:journals/corr/abs-2604-03037}, CoRe~\cite{DBLP:journals/corr/abs-2607-01721}, ACE-Brain-0.5~\cite{DBLP:journals/corr/abs-2607-04426}, TrustRoboReward~\cite{wang2026trustroboreward}, FIERCE~\cite{fierce2026}, Flora-Rewarder~\cite{leng2026flora}}}, draw=famH!80]
      ]
      [{\parbox{3.15cm}{\raggedright Structured process critics and failure diagnosis}}, fill=famH!7, draw=famH!80
        [{\parbox{10.9cm}{\raggedright AHA~\cite{DBLP:conf/iclr/DuanPKWTYKFMG25}, Guardian~\cite{DBLP:journals/corr/abs-2512-01946}, PRIMO-R1~\cite{DBLP:journals/corr/abs-2603-15600}, VLAC-Cut~\cite{zhai2026help}}}, draw=famH!80]
      ]
      [{\parbox{3.15cm}{\raggedright Judges for generated robot behavior}}, fill=famH!7, draw=famH!80
        [{\parbox{10.9cm}{\raggedright ReWorld$^{\ast}$~\cite{DBLP:journals/corr/abs-2601-12428}, RoboAlign-R1$^{\ast}$~\cite{DBLP:journals/corr/abs-2605-03821}, Reward as an Agent$^{\ast}$~\cite{DBLP:journals/corr/abs-2606-19990}, RoboGaze$^{\ast\ddagger}$~\cite{robogaze2026}, EVO-WAM$^{\ast\ddagger}$~\cite{evowam2026}, RoboJudge-9B$^{\ast}$~\cite{robojudge2026}}}, draw=famH!80]
      ]
      [{\parbox{3.15cm}{\raggedright Protocol judges and runtime verification}}, fill=famH!7, draw=famH!80
        [{\parbox{10.9cm}{\raggedright PRM-as-a-Judge$^{\ddagger}$~\cite{DBLP:journals/corr/abs-2603-21669}, Thea$^{\ddagger}$~\cite{wang2026thea}, VeriFine-Judge~\cite{verifine2026}}}, draw=famH!80]
      ]
    ]
  ]
\end{forest}
\vspace{-3mm}
\caption{Taxonomy of the 114 methods in Table~\ref{tab:rrm-method-comparison}
by semantic family and design route
(Sections~\ref{sec:models-outcome}--\ref{sec:models-hybrid}), with the
number of methods per family in parentheses. $^{\ast}$~evaluates generated
behavior; $^{\ddagger}$~marks a protocol or harness.}
\label{fig:taxonomy}
\end{figure*}

Under the common name \emph{reward}, these evaluators return success
probabilities, completion percentages, preference utilities, expected
returns, and structured diagnoses. Each answers a different question: a
success probability says whether the task is done, a preference says which
execution is better, and an expected return says what the policy can still
achieve. These differences matter once a prediction
becomes a control signal: ranking candidates, shaping a policy, and
deciding when to stop place different demands on the same score. We
therefore organize RRMs by the question their \emph{native output}
answers, where a native output is the quantity a model returns before any
downstream transformation. The questions define five semantic families:
\emph{Outcome}, \emph{Absolute Progress}, \emph{Relative}, \emph{Temporal
Value}, and \emph{Hybrid} (Section~\ref{sec:rrm_families};
Fig.~\ref{fig:taxonomy}).

These output differences call for different tests of evaluator quality,
and the common tests answer narrower questions than they are read to
answer. Agreement with frame order on successful
demonstrations~\cite{DBLP:conf/iclr/MaHFSLXK0DXBJ0Z25} asks whether later frames score higher, a condition that any score rising with time
also meets; it does not establish whether the score tracks task
fulfillment when progress stalls or is
reversed~\cite{DBLP:journals/corr/abs-2607-04426,DBLP:journals/corr/abs-2606-24742}.
Downstream policy performance reflects the evaluator, learning algorithm,
and control system together~\cite{DBLP:journals/corr/abs-2606-24742}.
Neither test separates task fulfillment from the cues that usually
accompany it: motion, a completed manipulation, elapsed time, and a view
that shows the outcome. On a real robot, Tian et
al.~\cite{DBLP:journals/corr/abs-2606-01036} find that learned reward
models over-reward unsafe or only superficially task-like behavior that
human evaluators reject, and
CriticHack~\cite{critichack2026} shows that optimizing a policy against a
frozen learned reward can raise both task success and wrong-object failures.

Current benchmarks rarely contain the executions in which these cues part
from the task. Four such cases arise in closed-loop executions
(Fig.~\ref{fig:bench-contrasts}): (A) motion without its effect, (B) a
wrong destination or another broken constraint, (C) an achieved condition
that is later lost, and (D) a missing decisive
view~\cite{DBLP:journals/corr/abs-2512-23703}. A test set sampled from
natural executions holds few instances of the first three and leaves the
fourth unlabeled (Section~\ref{sec:bench-suites}), and no existing RRM
benchmark draws all its failures from natural executions or presents one
execution under both sufficient and reduced views
(Table~\ref{tab:benchmarks}). RRM-Bench-1K therefore builds a contrast
suite for each case from real executions: (A) and (B) probe proxy
validity, (C) temporal validity, and (D) evidence validity. Each suite
fixes an ordering or an abstention between paired judgments rather than a
target value, so every family is tested through its native output.

Relying on a reward model takes three answers: what its output credits,
whether its evidence suffices, and which decisions it can drive. This
article is organized around them. Five families answer the first, four
suites test the first two, and the third depends on the decision: the fidelity it needs grows with the optimization pressure it puts on the reward.

Adjacent surveys cover foundation models in
robotics~\cite{DBLP:journals/ijrr/FirooziTTMSLZSKHIDWLS25}, robotic video
world models~\cite{DBLP:journals/corr/abs-2601-07823}, VLA
models~\cite{ma2026vlasurvey}, reward models in deep reinforcement
learning~\cite{yu2025rewardsurvey}, and reward and process reward models
for large language models~\cite{zhong2025rmsurvey,zheng2026prmsurvey}.
Concurrent surveys organize progress reward models for robot
learning~\cite{DBLP:journals/corr/abs-2607-21655} by input--output
interface, signal construction, and data and benchmarks, and verifiers for
robot policies~\cite{wan2026nofreechecker} by who supplies the judgment.
Unlike these surveys, this article spans all five families, organizes
them by the question their native outputs answer, and evaluates them
empirically on the proposed RRM-Bench-1K.

This article makes three contributions:
\begin{itemize}
\item \textbf{A survey and semantics-first taxonomy of foundation-era RRMs.}
We organize 114 methods by semantic family and design route, explaining
which quantity each route estimates and which behavior differences change
its output. Recording each method's output, query evidence, learning
mechanism, and training data (Table~\ref{tab:rrm-method-comparison})
shows that one semantics arises from several mechanisms and that the
evidence interface bounds what a result can establish.
\item \textbf{RRM-Bench-1K and a unified empirical study.} We annotate
1{,}117 real-robot videos of closed-loop executions on five embodiments,
from up to three cameras, with task-state labels for the conditions that
hold at each moment and camera-evidence labels for whether a view suffices
to judge the task. Four contrast suites separate motion from its effect, a
satisfied constraint from a broken one, an achieved condition from one
later lost, and a decisive view from its absence. A unified protocol
evaluates 27 evaluators on these suites and measures endpoint judgment,
dense credit, and evidence use separately instead of merging them
into one score (Table~\ref{tab:bench-controls}). The gaps it exposes
motivate six research directions (Section~\ref{sec:future}).
\item \textbf{RRM-Hub: an open-source toolbox.} Common adapters for 14
existing RRMs preserve each model's native outputs and supported inputs,
so they can be applied readily and new evaluators tested under the same
protocols. Stored raw outputs and per-capability reports make comparisons
reproducible.
\end{itemize}

Section~\ref{sec:definition_scope} defines RRMs, their five families, and
the scope; Sections~\ref{sec:models} and~\ref{sec:data} review model
designs, data foundations, and evaluation practice;
Sections~\ref{sec:rrm-bench} and~\ref{sec:experiments} present
RRM-Bench-1K, RRM-Hub, and the unified study; and
Section~\ref{sec:future} discusses open challenges.

\section{Definitions and Scope}
\label{sec:definition_scope}

\subsection{Reward Design without Foundation Models}
\label{sec:earlier-reward-design}

Before foundation models, each robot task or domain received its own
reward (Fig.~\ref{fig:teaser}(a)): specified by hand, computed by
geometric task oracles, recovered through inverse reinforcement
learning~\cite{ng2000irl,abbeel2004apprenticeship,ziebart2008maxent,ho2016gail,fu2018airl},
or learned from human preferences~\cite{christiano2017preferences},
success examples~\cite{fu2018vice,singh2019endtoend}, or self-supervised
video~\cite{sermanet2018tcn}. This line continues in evaluators
without foundation models: expert-video prediction
(VIPER~\cite{escontrela2023viper}, Diffusion
Reward~\cite{DBLP:conf/eccv/HuangJZX24}), goal distance over object
graphs (GORDON~\cite{protopapa2026gordon}), progress from video
(ProgAgent~\cite{DBLP:journals/corr/abs-2603-07784}) or tactile maps
(TaRL~\cite{tarl2026}), success--failure discrimination (Success
Visitation Matching~\cite{DBLP:journals/corr/abs-2606-23640}),
preference-aligned features (RAPL~\cite{rapl2024}), and, beside a
pretrained policy, value-based action ranking
(V-GPS~\cite{nakamoto2024vgps}) or failure verification
(VLA-Corrector~\cite{vlacorrectorsong2026}). Eureka~\cite{ma2024eureka},
Text2Reward~\cite{xie2024text2reward}, and Language to
Rewards~\cite{yu2023l2r} prompt a large language model to write per-task
reward code, which Hybrid evaluators such as
CoRe~\cite{DBLP:journals/corr/abs-2607-01721} and
GRAPE~\cite{zhang2024grape} combine with other judgments.

\subsection{Task-Conditioned Evaluation and Five Semantic Families}
\label{sec:rrm_definition}
\label{sec:rrm_families}

A robotic reward model is a task-conditioned evaluator that draws on
CLIP~\cite{DBLP:conf/icml/RadfordKHRGASAM21} or a later foundation model
for representation, supervision, or the judgment itself. It
assesses robot behavior against a task specification $g\in\mathcal{G}$,
given as language, a goal image, a reference demonstration, or a
structured goal, supplied with each query or fixed in the evaluator. It
reads execution evidence $e\in\mathcal{E}_{\iota}$ admitted by an
interface $\iota$: single observations, temporal windows, trajectory
prefixes, complete rollouts, paired executions, or synchronized camera
views, optionally with actions, proprioception, or simulator state. For an
evaluative query $q$, such as terminal success, progress, pairwise
preference, or value, the evaluator is a map
\begin{equation}
    \hat{y}^{(q)}=F_{\theta}^{(q)}(g,e),
    \qquad \hat{y}^{(q)}\in\mathcal{Y}_{q},
\label{eq:rrm-definition}
\end{equation}
whose native output space $\mathcal{Y}_{q}$ holds a label, probability,
score, ordering, distribution, or structured judgment.

Outputs of the same form can carry different meanings: a scalar may encode
visual compatibility, task progress, a preference utility, or an expected
return. We therefore assign each evaluator to the family whose question
its native output answers; what it is conditioned on, the specification
$g$ and the evidence that $\iota$ admits, is recorded for every entry but
does not decide the family (Section~\ref{sec:models-crosscutting}).
Outcome, Absolute Progress, Relative, and Temporal Value each answer one
question, and Hybrid composes them (Fig.~\ref{fig:taxonomy}); within each
family, Section~\ref{sec:models} groups methods by design route, that is,
by how they produce the output.

\emph{Outcome} (is the task done, or will it fail?) judges whether a
declared event has occurred or will occur within a horizon:
terminal success, end-of-episode grades, and failure forecasts such as
SAFE's running failure score~\cite{gu2025safe}
(Fig.~\ref{fig:outcome-semantics}). \emph{Absolute Progress} (how far
along is it?) covers pointwise estimates of advancement on a whole-task
scale, anchored by a goal, reference, learned scale, or task procedure,
such as the progress $p_t\in[0,1]$ of
ProcVLM~\cite{DBLP:journals/corr/abs-2605-08774}
(Fig.~\ref{fig:progress-semantics}). \emph{Relative} (which is better?)
covers quantities defined by ordering under a fixed task or criterion,
from a direct comparator's judgment of a pair or candidate set, such as
the signed hop between two states that the Robo-Dopamine
GRM~\cite{DBLP:journals/corr/abs-2512-23703} returns
(Fig.~\ref{fig:relative-semantics}), to a pointwise utility defined by
comparisons. \emph{Temporal Value} (which stage, and how well is it
going?) covers targets measured against a reference the execution itself sets: the stage it has reached, the time it still needs, such as
the seconds to completion predicted by RynnValue~\cite{huang2026rynnvalue}
(Fig.~\ref{fig:temporal-semantics}), or the return expected under the
policy that continues it, as in $V^{\pi}$ or $Q^{\pi}$; the target, not
temporal input, defines the family, so a whole-task progress estimate reading the full video history is Absolute Progress. \emph{Hybrid} (what
happened, and on what evidence?) covers evaluators that compose judgments from several families or as the fields of a structured verification judgment, as
Robometer~\cite{DBLP:journals/corr/abs-2603-02115} does with progress, success, and preference from one backbone
(Fig.~\ref{fig:hybrid-semantics}).

Table~\ref{tab:rrm-method-comparison} lists the family, output, query evidence, learning mechanism, and training data of all 114 methods.

\subsection{Scope and Corpus}
\label{sec:rrm_scope}

A method enters the corpus when it meets four criteria: (i)~its evaluative
use is validated on robot behavior, robot observations, or a behavior
representation; (ii)~its output answers a task-conditioned question about
success, progress, preference, value, behavior quality, or verification
and is used or evaluated as a decision signal; (iii)~CLIP or a later
foundation model contributes to the evaluator through representations,
adaptation, supervision, distillation (including a standalone model
trained on foundation-model labels), or direct judgment; and (iv)~the work
designs or evaluates the evaluator rather than only applying one. A
policy, video generator, or world model qualifies only through a component
that judges task fulfillment and can be queried or ablated on its own; its
action likelihood, reachability, or predicted observations describe what
is probable or feasible, not whether the task is fulfilled.
\section{Model Designs and Evaluative Semantics}
\label{sec:models}

Architecture and semantics are separate axes, so we first trace the evidence-to-decision path that all evaluators share and then review the
five families along the design routes of Fig.~\ref{fig:taxonomy}.
Figs.~\ref{fig:outcome-semantics}--\ref{fig:hybrid-semantics} show one
representative method per family on simulated RoboTwin~2.0
executions~\cite{chen2026robotwin2}, with its framework in (a) and its native output, with illustrative values, in (b);
ProcVLM~\cite{DBLP:journals/corr/abs-2605-08774},
Robo-Dopamine~\cite{DBLP:journals/corr/abs-2512-23703},
RynnValue~\cite{huang2026rynnvalue}, and
Robometer~\cite{DBLP:journals/corr/abs-2603-02115} are also evaluated in
Section~\ref{sec:experiments}.

\subsection{From Evidence to Decisions}
\label{sec:models-pipeline}

An evaluator encodes the task specification and the evidence its interface
admits, such as a camera set, an embodiment descriptor, a candidate
action, or a predicted future, and a query-specific head returns the
native output. A downstream rule then consumes the prediction:
SAFE~\cite{gu2025safe} (Fig.~\ref{fig:outcome-semantics}) raises an alarm
when its failure score \(s_t\) first exceeds a threshold \(\eta_t\)
calibrated on successful rollouts, shaping uses a progress estimate
\(p_t\) as the potential of a shaped reward~\cite{ng1999shaping}, and
selection ranks candidates by a preference utility. Each rule places its
own demand on the evaluator: selection needs only the order of the
candidates, whereas shaping exposes the reward to a policy that optimizes
against it (Section~\ref{sec:future-decisions}).

This pipeline separates \emph{semantics} (the estimated quantity),
\emph{form} (its encoding), and \emph{role} (its use), and the family
follows semantics alone, whatever a method calls its output. For
methods with several outputs, we set aside policy outputs and
deterministic transformations and assign the family the remaining outputs share. A rationale, confidence estimate, or abstention gate
attached to one judgment keeps that judgment's family, as in ProcVLM
(Fig.~\ref{fig:progress-semantics}), whereas independently meaningful
judgments from different families make a model Hybrid, as in Robometer
(Fig.~\ref{fig:hybrid-semantics}).

\subsection{Outcome Models}
\label{sec:models-outcome}

\begin{figure}[!t]
\centering
\includegraphics[width=\columnwidth]{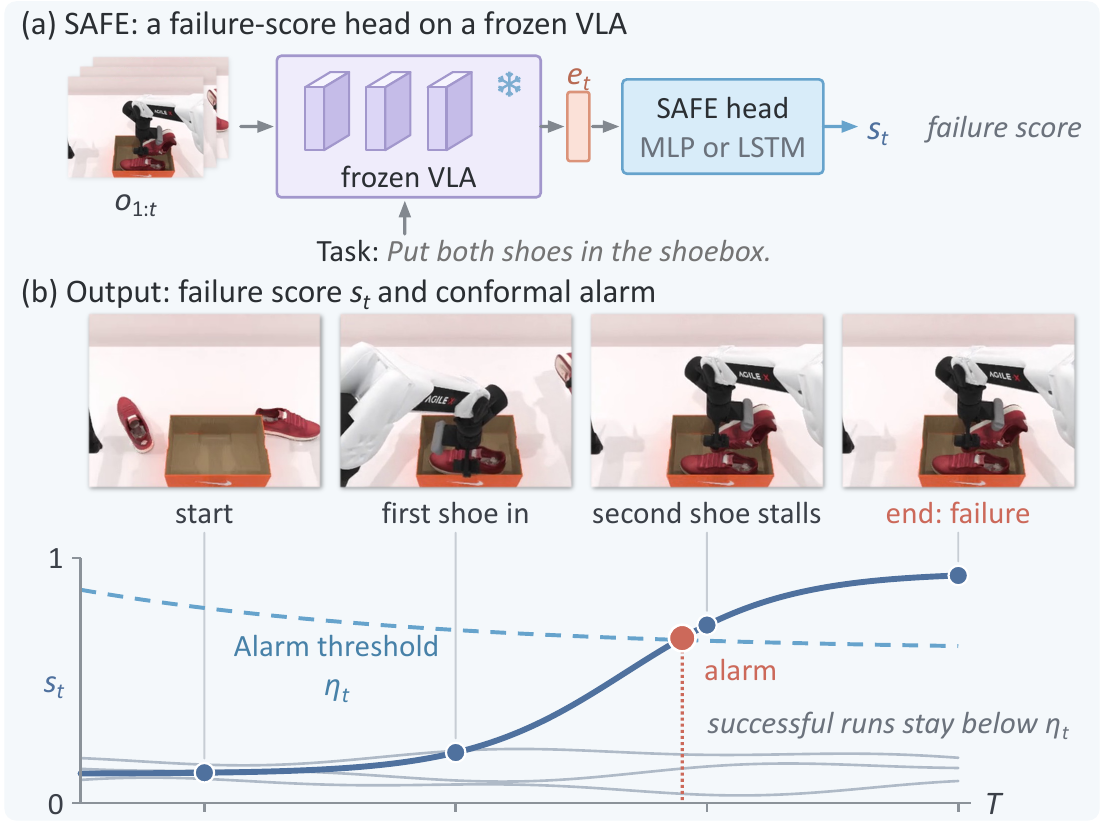}
\caption{SAFE~\cite{gu2025safe}, a representative Outcome model. (a)~A
light head on a frozen VLA maps its features $e_t$ to a running failure
score $s_t$. (b)~$s_t$ rises as the second shoe stalls and crosses the
conformal threshold $\eta_t$, which raises the alarm before the execution
fails.}
\label{fig:outcome-semantics}
\end{figure}

Outcome models judge whether a declared event, such as success or failure,
has occurred or will occur within a horizon. Episode-level judgments need
the decisive state in view, which suits acceptance and filtering;
predictive monitors such as SAFE (Fig.~\ref{fig:outcome-semantics})
forecast the event from signals that precede it, which suits stopping and
intervention.

\subsubsection{Episode-Level Outcome and Ordinal Judgments}

These judges read a foundation representation out into a verdict or a
grade and differ in what the readout learns to separate: positive and
negative goal descriptions in CLIP~\cite{DBLP:conf/icml/RadfordKHRGASAM21}
space (ZSRM~\cite{pmlr-v162-mahmoudieh22a}), whether a video shows the
task accomplished (SuccessVQA~\cite{du2023successvqa}, fine-tuned from
Flamingo~\cite{alayrac2022flamingo}), or demonstrations and unsuccessful
or synthetic data, judged from an image or a whole trajectory
(RoboFuME~\cite{robofume2023},
I-FailSense~\cite{DBLP:journals/corr/abs-2509-16072}); proprietary
embodied-reasoning models such as Gemini Robotics
ER~2~\cite{deepmind2026roboticser2} list success detection among their
specialties. Grades and similarities extend the verdict: a 1--5
end-of-episode grade
(RoboReward~\cite{DBLP:journals/corr/abs-2601-00675}), similarity to a
demonstration (RoboCLIP~\cite{DBLP:conf/nips/SontakkeZAPBSFI23}), or a
thresholded reward regressed from generated-video latents
(World-VLA-Loop~\cite{DBLP:journals/corr/abs-2602-06508}). A verdict can
reject only the failed endings that its negative examples cover, so
rejecting a wrong object or destination requires such negatives or
grounding in the instruction.

\subsubsection{Predictive and Policy-Internal Failure Monitors}

A forecast's main design choice is what the monitor reads. Policy-internal
monitors read the acting policy's features, which signal failure before it
shows in the scene but must be refit per policy: a failure score
learned from outcome-labeled rollouts (SAFE~\cite{gu2025safe}), a probe of
frozen OpenVLA~\cite{kim2024openvla} activations that flags steps within
15 steps of failure~\cite{mahato2026openvlafailure}, or calibrated feature
support (VLAConf~\cite{vlaconf2026}). External monitors read the execution
and apply across policies, asking a frontier VLM whether the robot will finish in time (Sentinel~\cite{agia2024sentinel}) or
detecting task-breaking deviations from anchors a world action model generates (FAVOR~\cite{favor2026}).

\paragraph{Evaluation implications}
An Outcome claim needs an operational event, a declared horizon, and evidence showing the event; grades call for ordinal metrics,
probabilities for calibration, and online monitors for alarm timing against the decisive event (Fig.~\ref{fig:outcome-semantics}(b)).
Endpoint and dense judgment are separate capabilities: RoboReward filters
completed executions well yet detects regressions worst
(Section~\ref{sec:exp-proxy}).

\FloatBarrier
\subsection{Absolute Progress Models}
\label{sec:models-progress}

\begin{figure}[!t]
\centering
\includegraphics[width=\columnwidth]{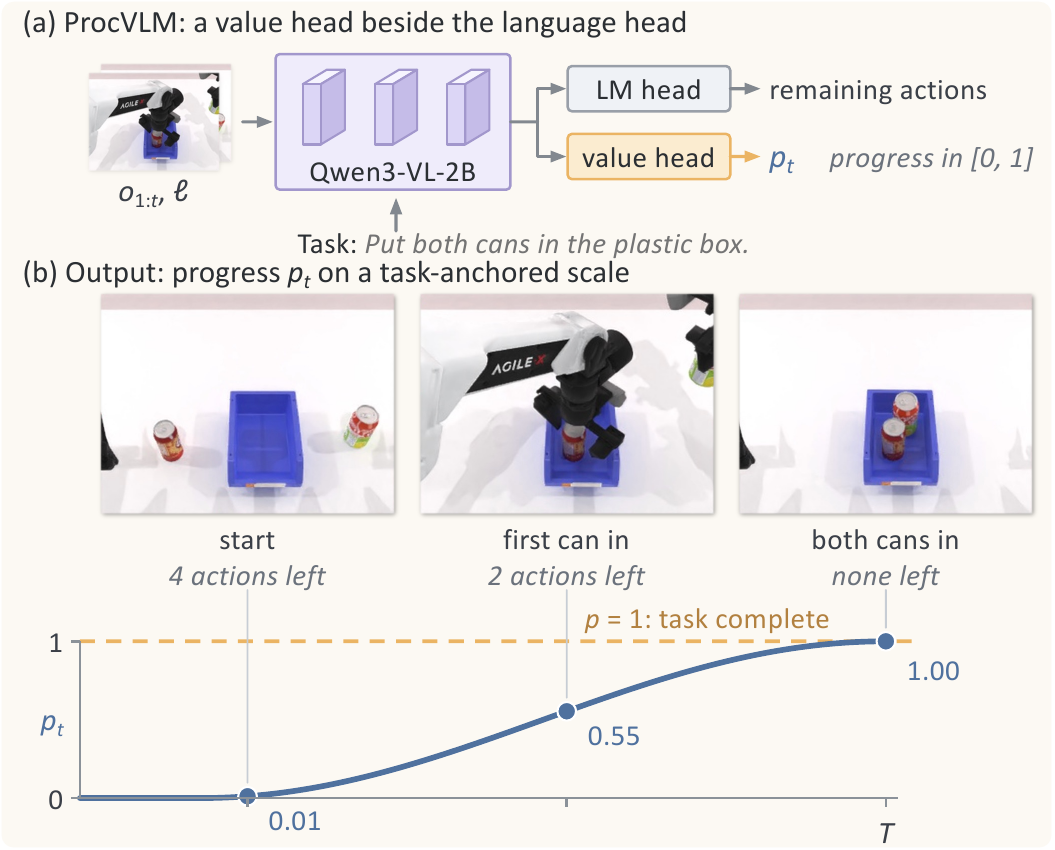}
\caption{ProcVLM~\cite{DBLP:journals/corr/abs-2605-08774}, a
representative Absolute Progress model. (a)~Given observations $o_{1:t}$
and instruction $\ell$, the language-model head first generates the
remaining actions, and the value head then reads this generated procedure
to output progress $p_t\in[0,1]$. (b)~$p_t$ rises as fewer actions remain
and reaches the anchor $p=1$ when both cans are in the box.}
\label{fig:progress-semantics}
\end{figure}

Absolute Progress models answer how far an execution has advanced on an
anchored scale; ProcVLM (Fig.~\ref{fig:progress-semantics}), for instance,
anchors progress to the remaining actions. Routes differ in where the anchor comes from: a goal or demonstration representation,
demonstration time, task semantics read by a VLM, or the system that
produces the behavior. The anchor decides how a score treats a step away from the goal configuration while the task advances, such as a
retraction to regrasp: a representation distance can read it as a
regression, a time-supervised estimate can credit it whether or not it
serves the task, and a procedure-based estimate can credit it if the
generated procedure contains it.

\subsubsection{Distances to Goal or Demonstration Representations}

The most direct anchor is a goal or demonstration in a representation
space, as in the goal-conditioned value representation of
VIP~\cite{DBLP:conf/iclr/MaSJBK023}, the language--visual goal alignment
of LIV~\cite{DBLP:conf/icml/MaK0BJ23}, and the zero-shot image--text
compatibility of VLM-RM~\cite{DBLP:conf/iclr/RocamondeMNPL24}. How the
alignment is learned decides which deviations the score registers: learned
failure prompts (Adapt2Reward~\cite{DBLP:conf/eccv/YangCQWWLGH24}), a
position code only for the first frame so that time cannot be read as
progress (ReWiND~\cite{DBLP:journals/corr/abs-2505-10911}), a light head
on frozen DINOv3~\cite{simeoni2025dinov3} and text features that ranks
states like analytic rewards (Rewarding
DINO~\cite{DBLP:journals/corr/abs-2603-16978}), and anchors grounded in
tracked object masks (SPARTA~\cite{sparta2025}) or in successful and
failed states (HiRE~\cite{hire2026}), which respond to changes that a
global similarity averages away.

\subsubsection{Temporal Supervision and Progress Proxies}

Temporal order within demonstrations offers a second scale, and with it
the assumption that achievement advances with time, which holds on
successful demonstrations. GVL~\cite{DBLP:conf/iclr/MaHFSLXK0DXBJ0Z25}
counters the assumption by asking a frozen VLM for the completion of
shuffled frames; others learn the scale from demonstrations with online
refinement (PROGRESSOR~\cite{DBLP:conf/iccv/AyalewZWJMW25}), a ranking
objective (Video-Language
Critic~\cite{DBLP:journals/tmlr/AlakuijalaMWFKM25}), reference alignment
(RARM~\cite{DBLP:journals/corr/abs-2606-22027}), distributional outputs
(WARP-RM~\cite{DBLP:journals/corr/abs-2606-28320}), or test-time updates
(VITA~\cite{vita2025}), and AnyviewMeter~\cite{anyviewmeter2026} adapts
Robometer to multiview input. Labels that fall teach the score to register
setbacks: chunks of normalized progress \(t/T\) whose frames are kept,
repeated, or reversed (World Value Models,
WVM~\cite{DBLP:journals/corr/abs-2606-24742}), preferences around
annotated retries (ReTVL~\cite{DBLP:journals/corr/abs-2606-24633}), or
simulated trajectories with failures and recoveries
(DenseReward~\cite{DBLP:journals/corr/abs-2607-13033}).

\subsubsection{Prompted, Token-Level, and Reasoning-Based Progress}

VLM-based estimators, prompted or fine-tuned, reach the anchor through
task semantics and differ in how the score is read out: the probability of
a task-truth token (TOPReward~\cite{DBLP:journals/corr/abs-2602-19313}),
the expected score token over repeated queries, which also serves as a
shaping reward (LLM-as-a-Verifier~\cite{llmverifier2026}), or a score that
follows reasoning, as a process rationale
(SOLE-R1~\cite{DBLP:journals/corr/abs-2603-28730}), a generated procedure
(ProcVLM), or recurrent steps over long tasks (R$^2$VLM~\cite{r2vlm2026}).
Feat2Go~\cite{feat2go2026} learns binned progress from feature similarity
to subgoal states, ARS~\cite{ars2026} assembles progress from events that
VLM agents verify over a rollout, and ProgressLM~\cite{progresslm2026} can
return an explicit unanswerable response, which supports
evidence-sufficiency tests. A generated procedure can name the object,
destination, and order that a constraint check needs, but a wrong
decomposition yields a smooth score for the wrong procedure.

\subsubsection{Policy-Coupled and World-Model-Coupled Progress}

Progress can also be estimated inside the system that proposes or predicts
behavior, which shows the score what the system intends next but ties it
to that system: World-Env
(RehearseVLA)~\cite{DBLP:journals/corr/abs-2509-24948} scores imagined
video for reward and early termination,
NavWAM~\cite{DBLP:journals/corr/abs-2606-13494} predicts goal progress
with future observations, and
ProgressVLA~\cite{DBLP:journals/corr/abs-2603-27670} guides action
generation by the progress of predicted frames.

\paragraph{Evaluation implications}
An Absolute Progress claim rests on an anchor, a declared scale, and a
measured response to the task state. Time-supervised and goal-anchored
scores part ways at motion without its effect and at a lost achievement
(Fig.~\ref{fig:bench-contrasts}(A), (C)), so tests must decouple the task
state from elapsed time; where the two diverge, most evaluated models
follow elapsed time (Section~\ref{sec:exp-legacy}).

\FloatBarrier
\subsection{Relative Models}
\label{sec:models-preference}

\begin{figure}[!t]
\centering
\includegraphics[width=\columnwidth]{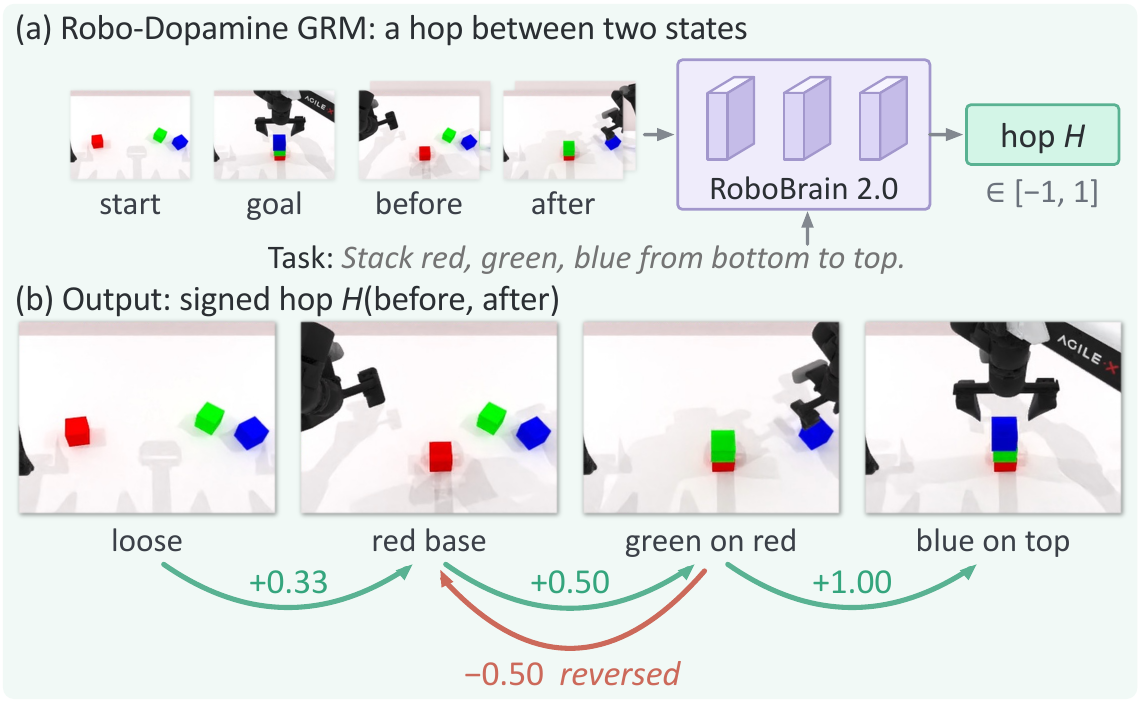}
\caption{The Robo-Dopamine GRM~\cite{DBLP:journals/corr/abs-2512-23703},
a representative Relative model. (a)~Given start and goal states, the GRM
compares a before and an after state and returns a signed hop
$H\in[-1,1]$. (b)~Hops are scaled to the distance that remains or to the
progress made, so chained hops keep progress in $[0,1]$; removing the
green block gives $-0.50$.}
\label{fig:relative-semantics}
\end{figure}

Relative models judge which of two executions, states, or actions better
satisfies the task. A direct comparator judges the items it is given,
often two states of one execution, so its sign registers a local change, a
regression included: the General Reward Model (GRM) of Robo-Dopamine
(Fig.~\ref{fig:relative-semantics}) returns a signed hop between a before
and an after state, and progress is chained from the hops. A utility
learned from preferences scores one item at a time, commonly through the
Bradley--Terry model~\cite{bradley1952rank}, in which the probability that
one item is preferred to another is a logistic function of their utility
difference, so its values carry meaning only as an ordering of items like
those compared in training.

\subsubsection{Direct Pairwise and Transition Comparators}

For transitions, the pair supplies the reference: a signed temporal distance between frames
(TimeRewarder~\cite{DBLP:journals/corr/abs-2509-26627}), hops between
multiview before/after states (Robo-Dopamine), refined with rollout
history, synthetic negatives, and a curriculum over hop sizes
(Robo-Dopamine~2.0~\cite{robodopamine2_2026}), or comparison with the instruction (DecisionNCE-T~\cite{decisionnce2024}), a predicted successful
transition (Dream2Reward~\cite{dream2reward2026}), or pairs of observed
and predicted candidates (a pair critic~\cite{smallstuff2026}).

\subsubsection{Pointwise Utilities Learned from Preferences}

The criterion that labels the training pairs defines the utility: VLM preferences on observation pairs
(RL-VLM-F~\cite{DBLP:conf/icml/WangSZXBHE24}), human comparisons under a
natural-language criterion (Freeform Preference Learning,
FPL~\cite{DBLP:journals/corr/abs-2606-32027}), multiview orderings from a
video--language model (MVR~\cite{DBLP:journals/corr/abs-2603-01694}), or
instruction-conditioned video scores (VLP~\cite{vlp2025}).
Rank2Reward~\cite{rank2reward2024} learns temporal order with a
Bradley--Terry head and inherits the time assumption of
Section~\ref{sec:models-progress}, and GIFT~\cite{gift2026} transfers a
maximum-entropy inverse-reinforcement-learning
reward~\cite{ziebart2008maxent} to new scenes, with
GPT-4o~\cite{openai2024gpt4o} inferring intent. Others make comparisons more informative: diverse queries that stop once the reward
stabilizes (RAPID~\cite{rapid2026}), trajectory sketches
(VARP~\cite{varp2025}), ranked lists (a listwise VLM reward~\cite{listwisevlm2026}), or joint pointwise and comparative pathways (VLM-AR3L~\cite{vlmar3l2026}).

\subsubsection{Candidate-Action Comparators}

When the alternatives are actions, ordering becomes a control primitive
for test-time selection. Scorers learn from synthetic preferences
(RoboMonkey~\cite{DBLP:journals/corr/abs-2506-17811}), predict a
refinement direction (RoVer~\cite{DBLP:journals/corr/abs-2510-10975}),
align instructions with actions
(CoVer~\cite{DBLP:journals/corr/abs-2602-12281}), or verify each candidate's progress
(TapSampling~\cite{DBLP:journals/corr/abs-2605-25547}), and some sit
beside a frozen policy as an energy verifier (TeV~\cite{tev2026}), a
success-visitation discriminator (DiVeR~\cite{diver2026}),
preference-guided flow matching
(PreferenceFlow~\cite{preferenceflow2026}), or zero-shot VLM verifiers
(EVE~\cite{eve2025}). Ranking gains depend on the proposal pool as well as the scorer, so best-of-\(N\) gains can come from sampling alone, and a
mismatched verifier can hurt: in the CoVer study, a RoboMonkey verifier
cut simulated success of a \(\pi_0\)~\cite{DBLP:journals/corr/abs-2410-24164} policy to under a fifth on red-teaming instructions.

\paragraph{Evaluation implications}
Relative models are judged by their choices: pairwise accuracy, ties,
ranking metrics, cycles among direct comparisons, and robustness to paraphrases and the candidate pool. Signed comparators must also give a
lost achievement a negative step (Fig.~\ref{fig:relative-semantics}(b));
Robo-Dopamine-GRM-3B does so when a shelf falls
(Fig.~\ref{fig:bench-results}) yet recalls under half the falls
(Table~\ref{tab:bench-main-results}).

\FloatBarrier
\subsection{Temporal Value Models}
\label{sec:models-value}

\begin{figure}[!t]
\centering
\includegraphics[width=\columnwidth]{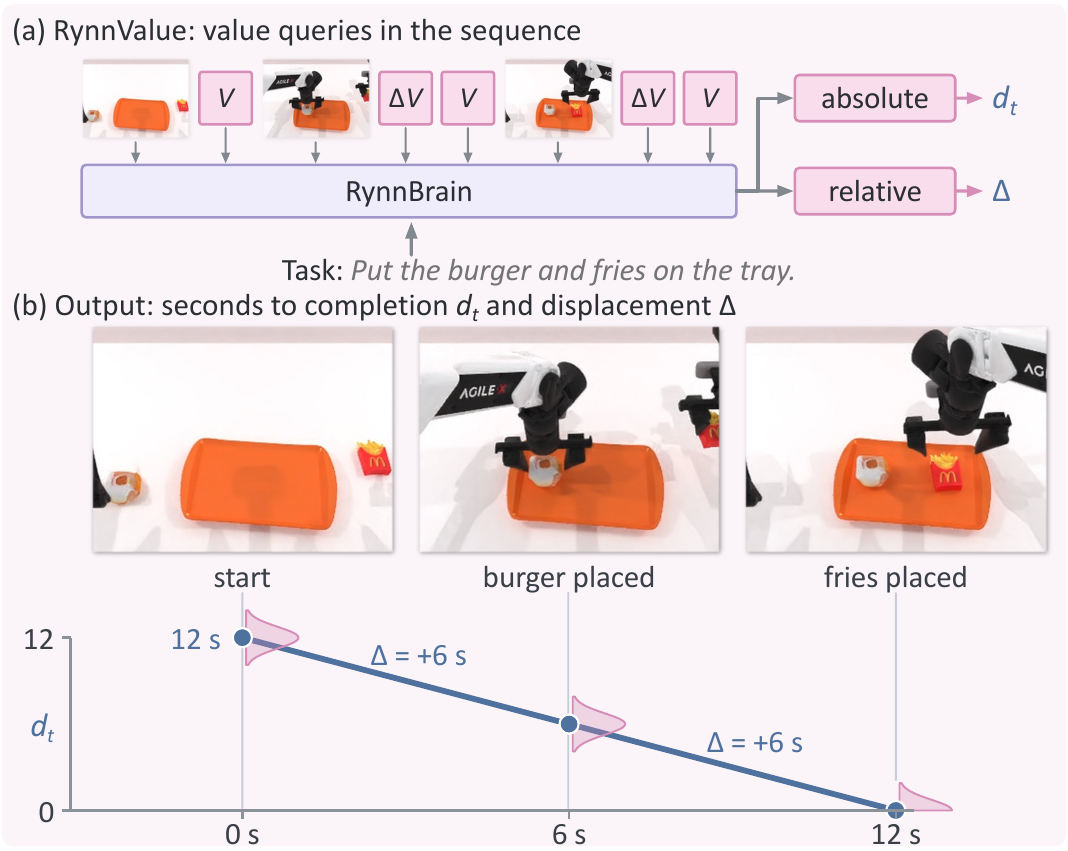}
\caption{RynnValue~\cite{huang2026rynnvalue}, a representative Temporal
Value model. (a)~Value ($V$) and displacement ($\Delta V$) query tokens
interleaved with the frames return the seconds to completion $d_t$ and
the signed displacement $\Delta$. (b)~Each placement advances the
execution by $\Delta=+6$~s, and $d_t$ falls from 12~s to 0~s.}
\label{fig:temporal-semantics}
\end{figure}

Temporal Value models estimate progress within a stage, the time left, or
the return still expected; RynnValue (Fig.~\ref{fig:temporal-semantics}),
for instance, counts the seconds to completion. A return-based value
\(V^\pi\), the discounted task reward expected from the current prefix,
also depends on the policy \(\pi\) that continues the execution, and
\(Q^\pi\) on the next action as well. These targets share temporal input
but answer different questions, so a stage error corrupts within-stage
progress, a speed change the time to go, and a policy change \(V^\pi\).

\subsubsection{Stage-Decomposed Values}

Stage decomposition replaces one global reference with a sequence of local
ones, and the designs differ in how they recognize the stage: with CLIP
(VICtoR~\cite{victor2024}, CLIP-Motion~\cite{clipmotion2023}), jointly
with within-stage progress
(SARM~\cite{DBLP:journals/corr/abs-2509-25358}), through primitive stages
shared across tasks (SARM2~\cite{DBLP:journals/corr/abs-2606-10305}), from
VLM-segmented demonstrations with grasp verification
(STDR~\cite{stdr2026}), or jointly with a policy's actions
(PALM~\cite{DBLP:journals/corr/abs-2601-07060}). Subtask structure can
also condition the reward (REDS~\cite{reds2025},
MARVL~\cite{DBLP:journals/corr/abs-2602-15872}) or be tracked over time as
VLM-proposed subgoals (T2-VLM~\cite{t2vlm2025}) or the farthest-reached
frame of a generated reference video (TeViR~\cite{tevir2025}). Local
references keep an early achievement from saturating a global score and
expose regressions within a stage, which makes stage assignment the
critical step.

\subsubsection{Temporal Distance Values}

Temporal distance can be counted in seconds or in control steps, so a slow
but correct execution has more time to go than a fast one in the same task
state. RynnValue learns its targets from timestamps, so speed, pauses, and
the completion cutoff all enter them, and suppresses temporal shortcuts
with random temporal sampling, order shuffling, and value-isolation
attention. Self-Improving Embodied Foundation
Models~\cite{ghasemipour2025selfimproving} count steps instead,
differencing the expected steps to an instructed goal into a shaped reward
and thresholding them into a success detector.

\subsubsection{Policy-Indexed and Value-in-the-Loop Models}

When the target is expected return, the policy becomes part of its
meaning. V-PTR~\cite{vptr2023} pretrains a value function on human video
for robot Q-learning, and RECAP~\cite{physicalintelligence2025pistar}, the
training recipe of \(\pi^{*}_{0.6}\), fits a VLM-initialized value
function to Monte Carlo returns and thresholds its advantages into an
indicator that conditions the policy, so its value is that of the
data-collecting policy; Robo-ValueRL~\cite{xia2026robovaluerl} conditions
such an estimator on visual history. Critics can also co-evolve with the policy or read
its representations, as a reward token refitted by temporal-difference
(TD) bootstrapping (eVTA~\cite{evta2026}), a Q-value for candidate chunks
(SeeQ~\cite{seeq2026}), or a linear probe of discounted outcomes
(frozen-VLA value probes~\cite{frozenvlaprobes2026}). Others act inside
the control loop: they repurpose a video generator as a value model
(ViVa~\cite{DBLP:journals/corr/abs-2604-08168}), rank sampled or
corrective actions (Hume~\cite{DBLP:journals/corr/abs-2505-21432},
DGAC~\cite{DBLP:journals/corr/abs-2606-21406}), or attach return or
progress values to the futures a world model generates (Cosmos
Policy~\cite{DBLP:journals/corr/abs-2601-16163},
RISE~\cite{DBLP:journals/corr/abs-2602-11075}, WCM~\cite{fei2026wcm},
MV-WAM~\cite{DBLP:journals/corr/abs-2606-21088},
MetaWAM~\cite{liang2026metawam}).

\paragraph{Evaluation implications}
A Temporal Value result is read against its declared target and the policy
or data distribution behind it. The target fixes the expected
response to reversal, stagnation, and retry: a lost achievement can leave
expected return high if recovery is likely, so RRM-Bench-1K tests only the sign of a policy-indexed value's response to a fall
(Section~\ref{sec:bench-conditions}).

\FloatBarrier
\subsection{Hybrid Models}
\label{sec:models-hybrid}

\begin{figure}[!t]
\centering
\includegraphics[width=\columnwidth]{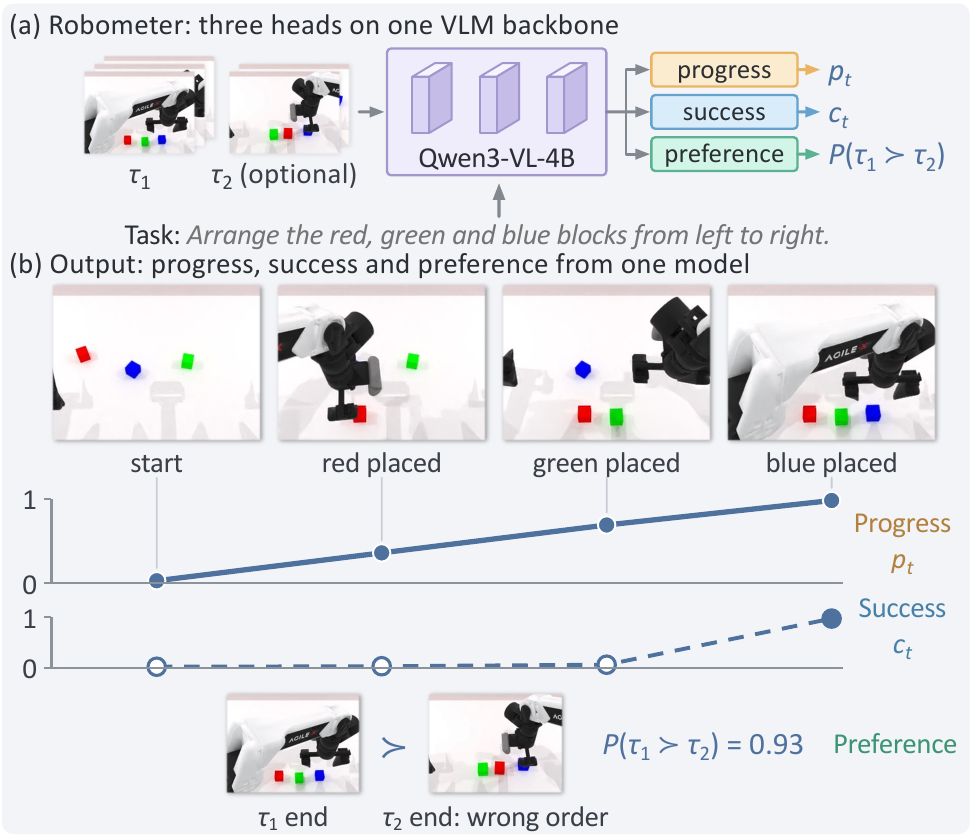}
\caption{Robometer~\cite{DBLP:journals/corr/abs-2603-02115}, a
representative Hybrid model. (a)~Three heads on one VLM backbone return
per-frame progress $p_t$, success $c_t$, and preference
$P(\tau_1\succ\tau_2)$ over two trajectories. (b)~$p_t$ rises with
each placed block, $c_t$ turns 1 only at the end, and $\tau_1$ is
preferred to a $\tau_2$ ending in the wrong order.}
\label{fig:hybrid-semantics}
\end{figure}

Hybrid models answer several evaluative questions at once, such as
progress and failure risk, or what failed and where; Robometer
(Fig.~\ref{fig:hybrid-semantics}), for instance, returns progress,
success, and a preference from three heads on one backbone. Hybrid designs
combine judgments through heads on a shared backbone, composite rewards,
or critics and judge ensembles that return a verdict with its cause and
evidence, and more outputs add information only when each passes its own
family's tests and their disagreements are resolved.

\subsubsection{Composite Rewards and Multi-Interface Models}

One strategy combines separate judgments across families: a VLM-preference
residual on a code-based formal reward
(CoRe~\cite{DBLP:journals/corr/abs-2607-01721}), a critic initialized with
VLM-estimated subgoal progress
(VLLR~\cite{DBLP:journals/corr/abs-2604-00055}), or rollouts ranked by
likelihood, success, and a foundation-model keypoint cost for preference
optimization (GRAPE~\cite{zhang2024grape}). Others keep each judgment
queryable, pairing progress with preference
(ACE-Brain-0.5~\cite{DBLP:journals/corr/abs-2607-04426}), a segment-level
advantage with a completion probability
(ARM~\cite{DBLP:journals/corr/abs-2604-03037}), a progress delta with a
done signal (VLAC~\cite{DBLP:journals/corr/abs-2509-15937}), an episode
grade with step-level action quality
(Flora-Rewarder~\cite{leng2026flora}), or observed
progress with a causal failure-risk head (FIERCE~\cite{fierce2026}). The
outputs must also agree: the temporal-contrast, progress, and completion
interfaces of Large Reward Models
(LRM)~\cite{DBLP:journals/corr/abs-2603-16065} can disagree, and
TrustRoboReward~\cite{wang2026trustroboreward} repairs contradictory
training labels through preference-ordered isotonic editing.

\subsubsection{Structured Process Critics and Failure Diagnosis}

A diagnosis ties the verdict to its cause.
AHA~\cite{DBLP:conf/iclr/DuanPKWTYKFMG25} and
Guardian~\cite{DBLP:journals/corr/abs-2512-01946}, tuned on failures
perturbed from or synthesized in simulation, judge from multiview images
whether a subtask or a plan succeeded and why, Guardian with a reasoning
trace; PRIMO-R1~\cite{DBLP:journals/corr/abs-2603-15600} learns by
supervised and then reinforcement fine-tuning to reason about progress in
video and transfers zero-shot to failure detection;
VLAC-Cut~\cite{zhai2026help} pairs signed progress with a diagnosis of
each change and cuts a rollout into segments to keep or discard.

\subsubsection{Judges for Generated Robot Behavior}

Judges of generated behavior separate realism from task completion, since a realistic video can still fail the task. ReWorld~\cite{DBLP:journals/corr/abs-2601-12428}
scores generated videos for physical realism, task completion,
embodiment plausibility, and visual quality,
RoboAlign-R1~\cite{DBLP:journals/corr/abs-2605-03821} distills such a
multidimensional teacher, and RoboJudge-9B~\cite{robojudge2026} rates
physical adherence and instruction alignment separately. Agentic judges
plan, vote, and reflect (Reward as an
Agent~\cite{DBLP:journals/corr/abs-2606-19990}), route subtask segments to
specialist VLMs returning time-localized failure events
(RoboGaze~\cite{robogaze2026}), or check inverse-dynamics consistency of generated video and actions (EVO-WAM~\cite{evowam2026}).

\subsubsection{Protocol Judges and Runtime Verification}

Protocol and action judges also return structured judgments.
PRM-as-a-Judge~\cite{DBLP:journals/corr/abs-2603-21669} aggregates
existing process reward models zero-shot into an
Outcome--Process--Diagnosis profile, and
VeriFine-Judge~\cite{verifine2026} distills a frontier-VLM teacher into a judge of instruction consistency and safety of proposed actions. At runtime, judgments decide whether to continue: Thea's \emph{Evaluation
as Exit Codes}~\cite{wang2026thea} checks the observation against an
explicit post-condition and returns \emph{process}, \emph{success}, or
\emph{failure} with evidence.

\paragraph{Evaluation implications}
Hybrid models call for field-wise evaluation: completion, failure type,
localization, confidence, and correction utility each need their own test,
explanations are checked against the actual failure, and composite rewards need each component measured directly.

\subsection{Cross-Cutting Design Dimensions}
\label{sec:models-crosscutting}

Semantics and mechanism vary independently: Absolute Progress comes from
embedding distances in VIP, demonstration time in VITA, token
probabilities in TOPReward, and generated procedures in ProcVLM, and one
backbone yields different evaluators under different queries
(Section~\ref{sec:exp-legacy}). Four further dimensions fix what a result
can establish. The \emph{evidence interface} bounds what is identifiable,
from goal compatibility in a single frame and the direction of change in a
before--after pair to an online alarm on a causal prefix and an outcome on
a complete rollout. \emph{Transfer} is shown by tests of the native evaluator, with any fitting, test-time adaptation, or hindsight
access stated. \emph{Provenance} determines which errors travel with a
model: frozen representations carry pretraining biases, adaptation its label shortcuts, and distillation the teacher's errors.
\emph{Coupling} determines what can be tested in isolation, since
policy-internal and world-model-coupled heads change with the system they
serve. Table~\ref{tab:rrm-method-comparison} records each method's evidence, provenance, and native output, and RRM-Bench-1K uses them to
assign evaluators to tests (Section~\ref{sec:bench-conditions}); reporting
the raw prediction with each derived signal keeps a frame index from passing as progress.

\section{Data Foundations and Evaluation Practice}
\label{sec:data}

Most evaluators learn from data, so how raw experience becomes targets decides what their judgment means. We follow this path to
supervision and use, then ask whether evaluations resolve the ambiguities training leaves.

\begin{table*}[!t]
\centering
\DeclareRobustCommand{\bmY}{\tikz[baseline=-0.55ex]
  \filldraw[line width=0.4pt] (0,0) circle (0.55ex);}
\DeclareRobustCommand{\bmP}{\tikz[baseline=-0.55ex]{
  \fill (0,0) -- (90:0.55ex) arc[start angle=90, end angle=270,
    radius=0.55ex] -- cycle;
  \draw[line width=0.4pt] (0,0) circle (0.55ex);}}
\DeclareRobustCommand{\bmN}{--}
\caption{RRM-Bench-1K and existing benchmarks for robotic
reward models. Scale follows each source's unit. \bmY: yes; \bmP: partial;
\bmN: no.}
\label{tab:benchmarks}
\begingroup
\footnotesize
\setlength{\tabcolsep}{3pt}
\renewcommand{\arraystretch}{1.0}
\newcommand{\bmH}[2]{\begin{tabular}[b]{@{}c@{}}#1\\#2\end{tabular}}
\begin{tabularx}{\textwidth}{@{}L{3.9cm}YL{3.4cm}ccccc@{}}
\toprule
Benchmark & Measurement target & Scale &
\bmH{Natural}{failures} & \bmH{Regression}{labels} & \bmH{Graded}{endpoints} &
\bmH{Paired}{evidence} & \bmH{Scope}{variants} \\
\midrule
ManiRewardBench~\cite{DBLP:journals/corr/abs-2602-19313} &
Stage progress; success detection & 653 episodes; 130 tasks &
\bmP & \bmN & \bmP & \bmN & \bmN \\
RBM-EVAL~\cite{DBLP:journals/corr/abs-2603-02115} &
Progress alignment; ranking & 571 OOD videos &
\bmP & \bmN & \bmY & \bmN & \bmN \\
PRIMO Bench~\cite{DBLP:journals/corr/abs-2603-15600} &
Progress under task and scene shift & 18k samples; humanoid set &
\bmN & \bmN & \bmN & \bmN & \bmN \\
RoboRewardBench~\cite{DBLP:journals/corr/abs-2601-00675} &
Episode-end progress grade (1--5) & 2{,}831 test items &
\bmP & \bmN & \bmY & \bmN & \bmN \\
RoboRMBench~\cite{jeung2026robormbench} &
Paraphrase invariance of rewards & 2{,}390 trajectories &
\bmP & \bmN & \bmY & \bmN & \bmN \\
Progress-Bench~\cite{progresslm2026} &
Progress from a demo; answerability & ${>}3$k instances &
\bmN & \bmN & \bmN & \bmP & \bmN \\
OpenGVL~\cite{opengvl2025} &
Temporal order of shuffled frames & 200 trajectories; 4 datasets &
\bmN & \bmN & \bmN & \bmN & \bmN \\
RoboPulse~\cite{DBLP:journals/corr/abs-2603-21669} &
Pairwise progress discrimination & 1{,}800 pairs; 816 tasks &
\bmN & \bmP & \bmN & \bmN & \bmN \\
Suboptimal-Value-Bench~\cite{DBLP:journals/corr/abs-2606-24742} &
Value under hesitation and retry & 800 trajectories; 15 tasks &
\bmP & \bmY & \bmN & \bmN & \bmN \\
RoboPulse++~\cite{prmjudge15_2026} &
Interval-level progress direction & 2{,}244 intervals &
\bmP & \bmY & \bmN & \bmN & \bmN \\
FailBench~\cite{navasardyan2026failbench} &
Binary success detection & 2{,}197 attempts; 14 sources &
\bmP & \bmN & \bmP & \bmN & \bmN \\
RoboValue~\cite{liu2026robovalue} &
Grounding, progress, recovery, and subtask credit &
2{,}792 test trajectories; 15 sim + 20 real tasks &
\bmN & \bmY & \bmP & \bmN & \bmN \\
\midrule
\textbf{RRM-Bench-1K (ours)} &
Fulfillment, retraction, and evidence & 1{,}117 videos &
\bmY & \bmY & \bmY & \bmY & \bmY \\
\bottomrule
\end{tabularx}
\endgroup
\end{table*}

\subsection{Sources of Raw Experience}
\label{sec:data-sources}

\paragraph{Robot demonstrations and failures}
Large teleoperation corpora such as Open
X-Embodiment~\cite{oneill2024openx}, DROID~\cite{khazatsky2024droid}, and
BridgeData~\cite{walke2023bridgedata} provide broad task coverage and are
widely used to train and evaluate reward
models~\cite{DBLP:conf/iclr/MaHFSLXK0DXBJ0Z25,DBLP:journals/corr/abs-2601-00675,DBLP:journals/corr/abs-2605-08774},
but they consist mostly of successful demonstrations, so the boundary
between acceptable, suboptimal, and failed behavior is thinly covered.
Recent collections add failures:
RoboMIND~\cite{DBLP:journals/corr/abs-2412-13877} releases about 5,000
real teleoperation failures annotated with their causes, and
RBM-1M~\cite{DBLP:journals/corr/abs-2603-02115} includes failed and
suboptimal executions among more than one million trajectories from 21
platforms.

\paragraph{Human and internet video}
Passive video extends semantic and temporal coverage beyond the available
robot trials: EPIC-KITCHENS~\cite{damen2022epic100} serves
LIV~\cite{DBLP:conf/icml/MaK0BJ23} and
PROGRESSOR~\cite{DBLP:conf/iccv/AyalewZWJMW25},
EgoDex~\cite{hoque2026egodex} serves
LRM~\cite{DBLP:journals/corr/abs-2603-16065} and
WVM~\cite{DBLP:journals/corr/abs-2606-24742}, action-free video serves
V-PTR~\cite{vptr2023} and VICtoR~\cite{victor2024},
VLAC~\cite{DBLP:journals/corr/abs-2509-15937} mixes human and robot video,
and EgoRecovery~\cite{DBLP:journals/corr/abs-2607-19745} records human
recovery behavior. Whether the corresponding robot action is feasible
depends on embodiment, viewpoint, and contact dynamics.

\paragraph{Simulation and autonomous rollouts}
Simulation makes failure coverage controllable:
RoboFAC~\cite{DBLP:journals/corr/abs-2505-12224},
FailSafe~\cite{DBLP:journals/corr/abs-2510-01642},
Guardian~\cite{DBLP:journals/corr/abs-2512-01946},
SOLE-R1~\cite{DBLP:journals/corr/abs-2603-28730}, and
DenseReward~\cite{DBLP:journals/corr/abs-2607-13033} generate failures
with privileged geometric or phase labels but inherit the simulator's
appearance, dynamics, and failure taxonomy. Policy rollouts yield errors
closer to the deployment distribution, with coverage set by the policy
that produced them, as in the rollouts that refine Cosmos
Policy~\cite{DBLP:journals/corr/abs-2601-16163},
World-VLA-Loop~\cite{DBLP:journals/corr/abs-2602-06508},
RISE~\cite{DBLP:journals/corr/abs-2602-11075}, and
ViVa~\cite{DBLP:journals/corr/abs-2604-08168}.

\paragraph{Human and model annotation}
Human annotation is sparse but captures qualities that proxies omit:
double-blind policy comparisons and critiques in
RoboArena~\cite{DBLP:journals/corr/abs-2506-18123}, stage
labels in SARM~\cite{DBLP:journals/corr/abs-2509-25358}, synchronized
keyframes in Robo-Dopamine~\cite{DBLP:journals/corr/abs-2512-23703},
progressive, regressive, or stagnant transitions in
ARM~\cite{DBLP:journals/corr/abs-2604-03037}, retry keypoints in
ReTVL~\cite{DBLP:journals/corr/abs-2606-24633}, and preferences along
annotator-named axes in FPL~\cite{DBLP:journals/corr/abs-2606-32027}.
VLMs scale up pairwise labels,
rationales, and subtask annotations, from
RL-VLM-F~\cite{DBLP:conf/icml/WangSZXBHE24} to the ProcCorpus of
ProcVLM~\cite{DBLP:journals/corr/abs-2605-08774}, trading annotation cost
against coverage and teacher error.

\subsection{From Trajectories to Supervision}
\label{sec:data-construction}

Label construction decides which differences the evaluator is taught to
preserve. Time and order labels teach stage and sequence, outcome labels finished versus failed endings, preferences a relative
ordering, and event or condition labels local change and constraints. Any
other distinction can only transfer from pretraining or other data.

\paragraph{Temporal and outcome-derived targets}
Normalized time, frame order, temporal distance, and discounted terminal
outcomes are inexpensive labels that a demonstration supplies with little
further
annotation~\cite{DBLP:journals/corr/abs-2509-15937,DBLP:journals/corr/abs-2509-26627,DBLP:journals/corr/abs-2604-08168,huang2026rynnvalue,vita2025}.
They equate achievement with elapsed time, which holds on successful
demonstrations, so motion without its effect and an achievement that is
later lost, the cases of suites (A) and (C), receive no label of their
own, and a return target further depends on the reward definition and the
policy.

\paragraph{Synthetic and natural negatives}
Because successful demonstrations dominate, negatives are often
synthesized from them, and each transformation teaches its own
distinction. Rewound videos, as in
ReWiND~\cite{DBLP:journals/corr/abs-2505-10911}, teach that backward
playback scores lower, although a real regression such as a dropped object
unfolds forward in time; clipped episodes, as in
RoboReward~\cite{DBLP:journals/corr/abs-2601-00675}, teach that a
truncated execution earns partial credit; and counterfactual instructions,
in RoboReward and CoVer~\cite{DBLP:journals/corr/abs-2602-12281}, teach
task-level grounding, whereas a wrong destination keeps the instruction
and changes one relation in the scene, the case of suite (B).
Perturbations in AHA~\cite{DBLP:conf/iclr/DuanPKWTYKFMG25} and
FailSafe~\cite{DBLP:journals/corr/abs-2510-01642} and world-model failures
in Dream2Fix~\cite{DBLP:journals/corr/abs-2603-13528} come closer to
physical errors but follow the perturbation model, so an evaluator can
learn its artifacts rather than the failure. Natural failures, such as the
RoboArena~\cite{DBLP:journals/corr/abs-2506-18123} rollouts that
RoboReward includes, supply the subtle errors of execution quality and
safety that synthetic negatives
underrepresent~\cite{DBLP:journals/corr/abs-2606-01036}, and few label
sets state whether decisive evidence is visible (suite (D)).

\paragraph{Preferences and intent}
Preferences replace a numeric target with a criterion: human judgments,
VLM labels, analytic rewards, distances to expert actions, and trajectory
order each give
\emph{better} its own meaning~\cite{DBLP:conf/icml/WangSZXBHE24,DBLP:journals/corr/abs-2506-17811,DBLP:journals/corr/abs-2603-16978,DBLP:journals/corr/abs-2606-32027}.
VARP~\cite{varp2025}, PreferenceFlow~\cite{preferenceflow2026}, and
GIFT~\cite{gift2026} draw preferences from trajectory sketches, human interventions, and inferred intent.

\paragraph{Stage and language constructions}
For long tasks, SARM~\cite{DBLP:journals/corr/abs-2509-25358},
SARM2~\cite{DBLP:journals/corr/abs-2606-10305}, REDS~\cite{reds2025},
VICtoR~\cite{victor2024}, and STDR~\cite{stdr2026} attach local progress
to inferred or annotated stages; a local label inherits its segmentation errors, so stage assignment and within-stage accuracy need separate
measurement. Question--answer, reasoning--score, and rubric formats fit labels to VLM training~\cite{DBLP:journals/corr/abs-2603-15600,DBLP:journals/corr/abs-2605-03821,progresslm2026}.

\subsection{Training Objectives, Use, and Provenance}
\label{sec:data-consumption}

Different targets share learning tools: regression, distributional
prediction, temporal-difference learning, contrastive alignment, ranking,
and token
prediction~\cite{DBLP:conf/iclr/MaSJBK023,DBLP:conf/icml/MaK0BJ23,DBLP:conf/icml/WangSZXBHE24,DBLP:journals/corr/abs-2603-15600,DBLP:journals/corr/abs-2606-24742},
with Bradley--Terry~\cite{bradley1952rank} and
Plackett--Luce~\cite{plackett1975permutations,luce1959choice} losses for
pairwise and listwise preferences~\cite{listwisevlm2026}. One loss serves
several targets, so target and interface define the native output.
The consuming decision sets demands on calibration, latency, and
error cost, and a new decision rule can reuse an existing evaluator: Reward-DAgger~\cite{rewarddagger2026} uses a frozen
Robometer~\cite{DBLP:journals/corr/abs-2603-02115} to decide when to
request human intervention, and
EVOLVE-VLA~\cite{DBLP:journals/corr/abs-2512-14666} accumulates the deltas
of VLAC~\cite{DBLP:journals/corr/abs-2509-15937} against milestone frames
into a progress value that rewards and ends test-time rollouts.

Scale does not remove assumptions built into labels: more successful
demonstrations can leave a subtle failure boundary unchanged, and
synthetic or VLM-generated labels can share shortcuts with their source models. High temporal correlation can then reflect a time-labeling rule, and agreement with a related VLM an inherited bias; the suites of Section~\ref{sec:bench-goal} separate them from task-state sensitivity.

\subsection{Existing Evaluation Practice}

\subsubsection{Protocols in Current Use}
\label{sec:eval-protocols}

A metric is informative when its target matches the claim: temporal correlation measures ordering in time, pairwise accuracy a
relative choice, outcome accuracy the recognition of an event, and scalar error agreement with a numerical target~\cite{DBLP:conf/iclr/MaHFSLXK0DXBJ0Z25,DBLP:journals/corr/abs-2601-00675,DBLP:journals/corr/abs-2603-02115,DBLP:journals/corr/abs-2603-15600}.
High value-order correlation (VOC) on successful videos therefore supports
a claim about ordering, whereas recognizing effects or reversals requires
tests that contain them, and downstream policy success reflects the evaluator with initialization, exploration, optimization, and the
candidate pool.

\subsubsection{Benchmarks and System-Level Resources}
\label{sec:eval-benchmarks}

Dedicated benchmarks broaden these protocols (Table~\ref{tab:benchmarks}):
beyond its columns, Progress-Bench~\cite{progresslm2026} tests
answerability, RoboPulse++~\cite{prmjudge15_2026} compares pair and
sequence inference, RoboValue~\cite{liu2026robovalue} branches seeded
failures by recovery outcome, and CriticHack~\cite{critichack2026}
stresses a frozen learned reward through policy optimization. Adjacent
resources evaluate failure explanation and correction
(RoboFail~\cite{DBLP:conf/corl/LiuBS23},
AHA~\cite{DBLP:conf/iclr/DuanPKWTYKFMG25},
RoboFAC~\cite{DBLP:journals/corr/abs-2505-12224},
ViFailback~\cite{DBLP:journals/corr/abs-2512-02787}), process and
streaming task understanding
(RoboProcessBench~\cite{DBLP:journals/corr/abs-2606-13040},
RoboChrono~\cite{robochrono2026}), and generated robot video
(RobotWorldBench~\cite{DBLP:journals/corr/abs-2605-03821},
RoboGazeBench~\cite{robogaze2026}, RoboJudge-Bench~\cite{robojudge2026}).
System-level resources supply realistic behavior from which evaluator
tests can be built: RoboArena~\cite{DBLP:journals/corr/abs-2506-18123}
aggregates double-blind pairwise preferences into policy rankings on one
embodiment, the DROID Franka setup;
RoboChallenge~\cite{DBLP:journals/corr/abs-2510-17950} provides
large-scale real-robot evaluation on single- and dual-arm robots and the
rollouts of RRM-Bench-1K;
WorldEval~\cite{DBLP:journals/corr/abs-2505-19017} and
dWorldEval~\cite{DBLP:journals/corr/abs-2604-22152} evaluate policies
through generated futures; and a deployment study on the SO-101
arm~\cite{DBLP:journals/corr/abs-2606-08881} characterizes naturally
occurring failures.

\subsubsection{Remaining Gaps}
\label{sec:eval-gaps}

Table~\ref{tab:benchmarks} compares the test sets on five properties:
\emph{natural failures} arise during the recorded executions,
\emph{regression labels} mark where progress is lost, \emph{graded
endpoints} distinguish zero, partial, and full fulfillment, \emph{paired
evidence} presents one execution under views labeled sufficient and reduced, and \emph{scope variants} narrow the task within an execution. Partial marks denote failures that are partly constructed or of
undocumented origin, regressions constructed by reversing frames, binary
endpoints, and evidence varied by editing or mismatching inputs. The other
test sets each fully provide at most one of these properties, and none
fully provides paired evidence or scope variants, so coverage stays thin
where convenient cues disagree with the task.
\section{RRM-Bench-1K: A Capability Benchmark}
\label{sec:rrm-bench}

\begin{figure*}[!t]
\centering
\includegraphics[width=\linewidth]{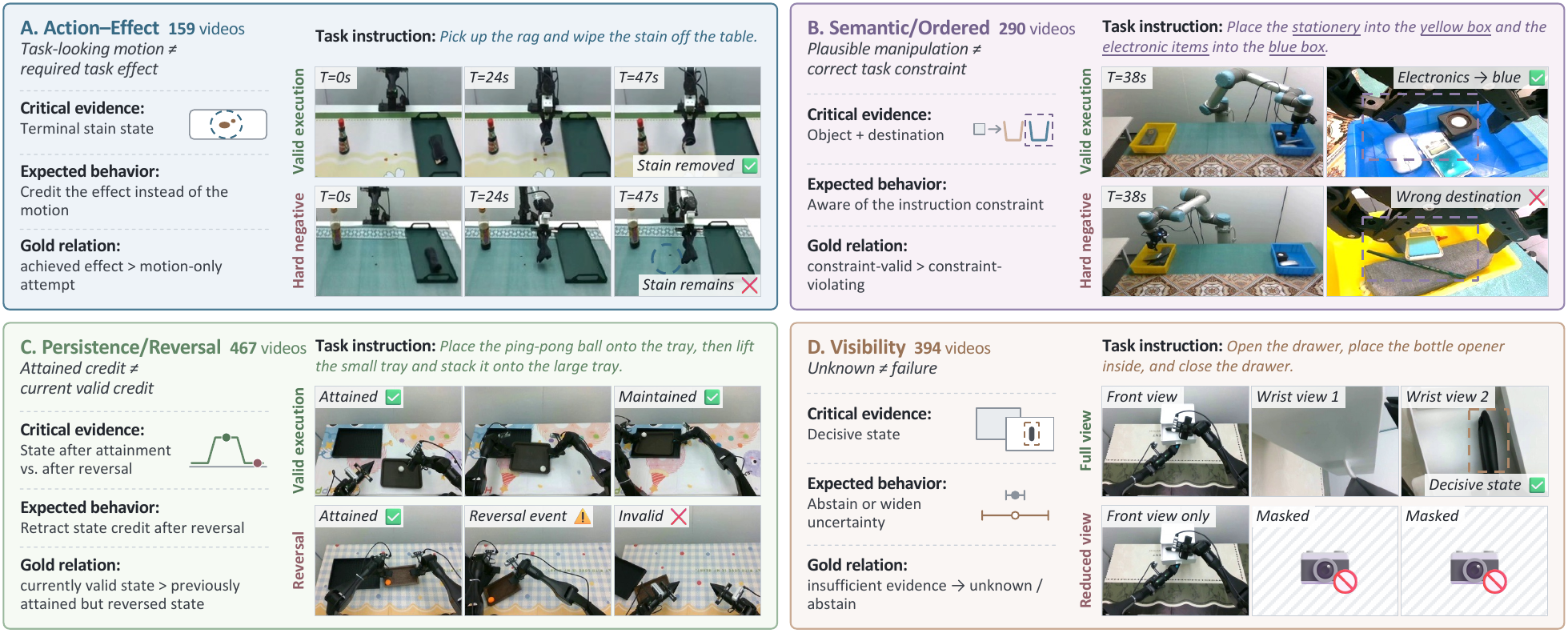}
\vspace{-2mm}
\caption{The four contrast suites of RRM-Bench-1K on real robot
executions. Each panel names the critical evidence, the expected behavior,
and the gold relation; the count after each suite name is its number of videos. (A) and (B) test proxy validity, (C) temporal
validity, and (D) evidence validity.}
\label{fig:bench-contrasts}
\vspace{-2mm}
\end{figure*}

Relying on a reward model requires that its score rise for the right reason
and rest on evidence that shows it. When a robot asked to wipe a stain
sweeps a rag across the table and stops while the stain remains
(Fig.~\ref{fig:bench-contrasts}(A)), a score that tracks motion, elapsed
time, or resemblance to successful videos still rises. RRM-Bench-1K is
built around such cases and combines the five properties of
Table~\ref{tab:benchmarks}. \textbf{Natural failures:} its 1{,}117 videos
record learned policies on a real robot platform, each execution kept
whole with its failures, partial completions, retries, and recoveries.
\textbf{Regression labels:} task-state labels record which task conditions
hold at each moment and withdraw credit when an achieved condition is
lost. \textbf{Graded endpoints:} the final task state grades every video
as zero, partial, or full fulfillment. \textbf{Paired evidence:} up to
three cameras record each execution, and each view is labeled by whether
it shows enough to judge the task. \textbf{Scope variants:} some videos
narrow the task scope within an execution. Four suites use these
properties.

\subsection{Four Decisive Cases and Their Contrast Suites}
\label{sec:bench-goal}
\label{sec:bench-suites}

Each suite separates the credited condition from a cue that usually
accompanies it in natural executions (Fig.~\ref{fig:bench-contrasts}). A
suite pairs an execution that earns credit with a counterpart that shares
the cue but not the credit, a hard negative in (A)--(C) and a reduced view
in (D). It names the critical evidence that separates them and fixes a gold
relation between their judgments, an ordering or an abstention rather than
a target value, so each family is tested through its native output
(Section~\ref{sec:bench-metrics}). Success and failure controls
anchor each suite. Cases (A) and (B) test
\emph{proxy validity}, whether credit follows the specified object,
destination, order, and effect; (C) tests \emph{temporal validity}, whether
credit lasts as long as the credited condition holds; and (D) tests
\emph{evidence validity}, whether the observations support the
evaluator's judgment.

\textbf{(A) Action--Effect} separates task-like motion from its effect,
which arrive together when an execution succeeds. It includes
executions whose motion resembles success while the required effect is
verifiably absent, such as wiping with an empty gripper or pressing a
switch that does not flip. The terminal effect is the critical evidence: an
execution that achieves it must outscore a motion-only attempt, and a score
that ranks the two alike follows motion or elapsed time.

\textbf{(B) Semantic/Ordered Constraint} separates a successful
manipulation from a correct one. It includes wrong objects or
destinations, omitted instances, and violated prerequisites, in which the
grasp, transport, and release can all succeed. The critical evidence is the
object and its destination read against the instruction, so only an
evaluator that checks them ranks the constraint-valid execution higher.

\textbf{(C) Persistence/Reversal} separates an achieved condition from one
that still holds. It pairs conditions that persist with ones achieved and later lost; conditions that were never
achieved form a separate case. Credit must fall at the reversal: an
evaluator that keeps it accumulates history, and one that withdraws it
tracks the current task state.

\textbf{(D) Visibility} separates what happened from what a camera shows.
It shows executions under reduced camera evidence, most beside a
sufficient view of the same execution, so the task state stays fixed while
the evidence changes. Without the decisive state in view, the expected
behavior is to abstain or widen uncertainty; an evaluator without such an
output is measured by how far its judgment degrades.

Natural executions rarely produce cases (A)--(C) on their own
(Supplementary Section~\ref{sec:supp-temporal}): the great majority of
task-state falls end a failed attempt or follow a drop or release failure, and
lost achievements, wrong destinations, and actions without effect make up
the rest. Case (D), by contrast, is common: over a third of the single
views give partial or insufficient evidence
(Section~\ref{sec:bench-reference}). On successful demonstrations each
pair of factors moves together, and a test set sampled at random would
contain few instances of (A)--(C) and leave (D) unlabeled. The suites
concentrate these cases while keeping every video a real execution, and
each isolates a capability that dense shaping and monitoring need. No
evaluator in Section~\ref{sec:experiments} handles all four.

\subsection{Data from Real Closed-Loop Executions}
\label{sec:bench-core}

RRM-Bench-1K is built on closed-loop executions recorded on RoboChallenge
Table30~\cite{DBLP:journals/corr/abs-2510-17950} and Table30
V2~\cite{robochallenge_v2_report}, a real-robot platform that runs learned
policies on standardized tasks and scores each execution with a rubric.
Because the policies act in closed loop, their failures arise naturally
and are recorded under one protocol. The benchmark covers 52 tasks on five
embodiments: three single-arm robots (ARX-5, UR5, and Franka) and two
dual-arm robots (Cobot Magic ALOHA and DOS-W1). Up to three platform
cameras record each execution; the camera views of whole executions and of
their scope variants (Section~\ref{sec:bench-reference}) form the 1{,}117
videos, and a video recorded alongside other cameras can also be scored
together with them as one multiview input.

\subsection{Task-State and Camera-Evidence Labels}
\label{sec:bench-reference}

We annotate each execution with task units, events, task-state labels, and
camera-evidence labels, and derive the scope variants and contrast suites
from these annotations. Annotation proceeds in three steps. First, to fix
what earns credit, we write for each task the \emph{task units} of its
instruction: object and destination constraints, repeated instances, and,
where declared, prerequisites, each with the rule by which it keeps its
credit. A placement that must persist loses its credit when undone,
whereas a required action stays credited after the robot moves on. Second,
to locate when the task state changes, annotators review each execution
frame by frame and mark the \emph{events}: completions, failed attempts,
losses, and reversals, with retries and re-achievements where they occur.
Each event is time-stamped and linked to a unit, and its type decides
whether credit is added, withdrawn, or unchanged. Third, to obtain a dense
reference without judging every frame, a rule-based compiler turns the
units and events into a score over time for each declared task scope,
either the full task or a scope variant that covers one object, one unit,
or a window of units under a narrower instruction, applying that scope's
credit, prerequisite, and reversal rules. One set of events thus serves
every scope of an execution (Supplementary
Section~\ref{sec:supp-settings}). The resulting label records the task
state rather than the motion performed. An action-history reference, which
accumulates credit for completed actions, is kept for comparison; the
two references end differently in over a third of the executions, and the
task-state reference falls at least once in over half.

Whether a condition holds and whether a camera shows it are labeled
separately. The state label is read from the views showing the decisive conditions, and each camera view is labeled as giving sufficient (V),
partial (P), or insufficient (X) evidence to judge the task. An effect
that no view can establish stays unresolved, excluded from state metrics. Most videos are labeled V and about a third P; X is rare.

\subsection{Evaluation Protocol and Metrics}
\label{sec:bench-conditions}
\label{sec:bench-metrics}

\emph{Eligibility and settings.} Each evaluator enters the tests its
native query and output support, Relative models through comparisons of
observation pairs. The direction test fixes only the sign of a change,
so a time-to-go or policy-indexed value
(Section~\ref{sec:models-value}) also enters it, and a terminal grade
enters on successive prefixes.
Specialized evaluators run through their documented interfaces with native
outputs; departures are given in
Supplementary Section~\ref{sec:supp-settings} and
Table~\ref{tab:supp-roster}.

\emph{Task fulfillment and false credit.} Endpoint judgments are compared
within each task, whose executions share an instruction but differ
in final completion. The study asks whether the highest final score selects
the most complete execution and how often a failed execution reaches the
median final score of fully completed ones. On failures of suites (A) and (B), this share is false credit.

\emph{Temporal response.} Following RoboPulse++~\cite{prmjudge15_2026},
direction classification labels each 1~FPS transition as rising or falling
and reports per-class precision, recall, and F1 with Macro-F1 and accuracy
(Table~\ref{tab:bench-main-results}). Every non-flat transition of the
task-state reference is scored, and an unchanged prediction is a miss for both classes. Event-level tests compare a model's recall of falls following a specific cause with its own rate of predicting a fall
anywhere; a model that predicts falls at random recalls every cause at that
rate.

\emph{Conventional metrics.} VOC correlates predicted values with elapsed time, averaged over successful executions, so it measures
temporal ordering where time and task state advance together. Kendall's
$\tau_a$ is the rank agreement of final predictions with endpoint
fulfillment within each task, averaged over tasks, so it measures endpoint
ranking (Table~\ref{tab:bench-controls}). Both use native outputs, with
VOC in the Pearson form of Robometer~\cite{DBLP:journals/corr/abs-2603-02115}
and $\tau_a$ as in Robometer and RynnValue~\cite{huang2026rynnvalue}.

\emph{Evidence and scope.} The paired-evidence test scores each P/X video
and its V counterpart on the same transitions against one task-state label, so only the camera changes
(Table~\ref{tab:bench-evidence-results}). Scope analysis compares each
scope variant with the full task of its execution through unit mappings, because a narrower scope can be complete while the full task is
not.

\emph{Controls.} Normalized elapsed time, ignoring video, and a
predictor that always answers rising are shortcut controls; the
action-history reference (Section~\ref{sec:bench-reference}) is a
label control.

\subsection{RRM-Hub Toolbox}
\label{sec:rrm-hub}

RRM-Hub is an open-source toolbox for testing a reward model on the
questions of this benchmark. A model enters through an adapter that
declares the queries, inputs, and native outputs it supports. The toolbox
runs each eligible test, keeps the raw outputs, and reports each
capability separately, so a new evaluator is compared with the published
results suite by suite. Existing RRMs run on new videos through the same interface. Adapters cover the 14 specialized methods of
Section~\ref{sec:exp-models} with their checkpoints and interfaces, and
the general-purpose VLMs and Jev~1.14~\cite{typesafe2026jev} are queried
with fixed prompts (Supplementary Section~\ref{sec:supp-settings}).

\begin{table*}[!t]
\centering
\caption{Progress-direction classification on RRM-Bench-1K. Bold marks the best single-view result in each panel. $^{\dagger}$Curve
interpolated to 1~FPS. $^{\ddagger}$Multiview input
(Section~\ref{sec:exp-evidence}); all other rows take single-view input. $^{\S}$Scored on
adjacent source frames.}
\label{tab:bench-main-results}
\footnotesize
\setlength{\tabcolsep}{4pt}
\begin{tabularx}{\textwidth}{@{}l!{\hspace{12pt}}
*{3}{>{\centering\arraybackslash}X}!{\hspace{12pt}}
*{3}{>{\centering\arraybackslash}X}!{\hspace{12pt}}
*{2}{>{\centering\arraybackslash}X}@{}}
\toprule
& \multicolumn{3}{c!{\hspace{12pt}}}{Rising}
& \multicolumn{3}{c!{\hspace{12pt}}}{Falling}
& \multicolumn{2}{c@{}}{Both classes} \\
\cmidrule(l{4pt}r{16pt}){2-4}\cmidrule(l{4pt}r{16pt}){5-7}\cmidrule(l{4pt}){8-9}
Model & Precision & Recall & F1 & Precision & Recall & F1 & Macro-F1 & Accuracy \\
\midrule
\rowcolor{black!6}\multicolumn{9}{l}{\textbf{\textit{Specialized evaluators, 2B--4B}}} \\
ProcVLM-2B~\cite{DBLP:journals/corr/abs-2605-08774}$^{\dagger}$ & 0.924 & 0.540 & 0.682 & 0.128 & 0.465 & 0.201 & 0.441 & 0.533 \\
VLAC-2B~\cite{DBLP:journals/corr/abs-2509-15937}$^{\dagger}$ & 0.922 & 0.374 & 0.532 & 0.118 & 0.312 & 0.171 & 0.352 & 0.368 \\
Robo-Dopamine-GRM-3B~\cite{DBLP:journals/corr/abs-2512-23703} & 0.908 & 0.412 & 0.566 & 0.135 & 0.401 & 0.202 & 0.384 & 0.411 \\
\quad Multiview$^{\ddagger}$ & 0.929 & 0.593 & 0.724 & 0.161 & 0.482 & 0.242 & 0.483 & 0.581 \\
Robo-Dopamine-GRM-2.0-4B-Preview~\cite{DBLP:journals/corr/abs-2512-23703} & 0.927 & 0.594 & 0.724 & 0.158 & \textbf{0.518} & \textbf{0.242} & 0.483 & 0.586 \\
\quad Multiview$^{\ddagger}$ & 0.940 & 0.730 & 0.822 & 0.219 & 0.567 & 0.316 & 0.569 & 0.713 \\
Robo-ValueRL~\cite{xia2026robovaluerl}$^{\ddagger}$ & 0.909 & 0.514 & 0.656 & 0.105 & 0.526 & 0.175 & 0.416 & 0.515 \\
Robometer-4B~\cite{DBLP:journals/corr/abs-2603-02115}$^{\dagger}$ & 0.916 & \textbf{0.871} & \textbf{0.893} & 0.181 & 0.263 & 0.214 & \textbf{0.554} & \textbf{0.812} \\
RoboReward-4B~\cite{DBLP:journals/corr/abs-2601-00675}$^{\dagger}$ & \textbf{0.937} & 0.263 & 0.410 & \textbf{0.209} & 0.044 & 0.072 & 0.241 & 0.241 \\
RynnValue-4B~\cite{huang2026rynnvalue}$^{\dagger}$ & 0.929 & 0.677 & 0.783 & 0.153 & 0.507 & 0.235 & 0.509 & 0.660 \\
\rowcolor{black!6}\multicolumn{9}{l}{\textbf{\textit{Specialized evaluators, 7B--8B}}} \\
PRIMO-R1-7B~\cite{DBLP:journals/corr/abs-2603-15600}$^{\dagger}$ & 0.914 & 0.571 & 0.703 & 0.135 & 0.181 & 0.154 & 0.429 & 0.531 \\
LRM-8B~\cite{DBLP:journals/corr/abs-2603-16065} (Contrastive)$^{\S}$ & 0.921 & 0.171 & 0.288 & 0.071 & \textbf{0.754} & 0.130 & 0.209 & 0.213 \\
LRM-8B~\cite{DBLP:journals/corr/abs-2603-16065} (Progress) & 0.924 & 0.155 & 0.265 & 0.127 & 0.148 & 0.136 & 0.201 & 0.154 \\
Robo-Dopamine-GRM-8B~\cite{DBLP:journals/corr/abs-2512-23703} & 0.921 & 0.456 & 0.610 & 0.141 & 0.422 & 0.211 & 0.411 & 0.453 \\
\quad Multiview$^{\ddagger}$ & 0.933 & 0.604 & 0.733 & 0.176 & 0.489 & 0.259 & 0.496 & 0.592 \\
Robo-Dopamine-GRM-2.0-8B-Preview~\cite{DBLP:journals/corr/abs-2512-23703} & \textbf{0.931} & 0.594 & 0.725 & 0.148 & 0.478 & 0.226 & 0.476 & 0.582 \\
\quad Multiview$^{\ddagger}$ & 0.938 & 0.727 & 0.819 & 0.212 & 0.552 & 0.307 & 0.563 & 0.709 \\
RynnValue-8B~\cite{huang2026rynnvalue}$^{\dagger}$ & 0.926 & 0.684 & \textbf{0.787} & \textbf{0.160} & 0.445 & \textbf{0.236} & \textbf{0.511} & 0.661 \\
SOLE-R1-8B~\cite{DBLP:journals/corr/abs-2603-28730} & 0.925 & \textbf{0.685} & \textbf{0.787} & 0.146 & 0.492 & 0.225 & 0.506 & \textbf{0.666} \\
\quad Multiview$^{\ddagger}$ & 0.934 & 0.745 & 0.829 & 0.181 & 0.515 & 0.267 & 0.548 & 0.723 \\
TOPReward~\cite{DBLP:journals/corr/abs-2602-19313} (Qwen3-VL-8B~\cite{qwen2025qwen3vl})$^{\dagger}$ & 0.913 & 0.558 & 0.693 & 0.111 & 0.503 & 0.182 & 0.437 & 0.552 \\
VLAC-8B~\cite{DBLP:journals/corr/abs-2509-15937}$^{\dagger}$ & 0.917 & 0.423 & 0.579 & 0.120 & 0.285 & 0.169 & 0.374 & 0.410 \\
\rowcolor{black!6}\multicolumn{9}{l}{\textbf{\textit{Specialized evaluators, 27B--30B}}} \\
Flora-Rewarder-27B~\cite{leng2026flora} & \textbf{0.930} & 0.748 & 0.829 & 0.172 & \textbf{0.472} & 0.252 & 0.541 & 0.721 \\
\quad Multiview$^{\ddagger}$ & 0.924 & 0.735 & 0.819 & 0.154 & 0.443 & 0.229 & 0.524 & 0.706 \\
VLAC-Cut-30B-A3B~\cite{zhai2026help} & 0.922 & \textbf{0.916} & \textbf{0.919} & \textbf{0.276} & 0.268 & \textbf{0.272} & \textbf{0.596} & \textbf{0.853} \\
\rowcolor{black!6}\multicolumn{9}{l}{\textbf{\textit{Specialized evaluators, proprietary backbone}}} \\
GVL~\cite{DBLP:conf/iclr/MaHFSLXK0DXBJ0Z25} (Gemini 3.8 Flash) & 0.928 & 0.730 & 0.817 & 0.167 & 0.422 & 0.240 & 0.528 & 0.700 \\
\midrule
\rowcolor{black!6}\multicolumn{9}{l}{\textbf{\textit{General-purpose VLMs, Pair style}}} \\
GPT-6.1 Sol~\cite{openai2026gpt61sol} & 0.929 & \textbf{0.920} & \textbf{0.925} & \textbf{0.326} & 0.355 & \textbf{0.340} & \textbf{0.632} & \textbf{0.865} \\
GPT-6 Luna~\cite{openai2026gpt6astra} & 0.917 & 0.640 & 0.754 & 0.123 & \textbf{0.466} & 0.195 & 0.474 & 0.623 \\
GPT-5.6 Terra~\cite{openai2026gpt56} & 0.914 & 0.908 & 0.911 & 0.200 & 0.212 & 0.206 & 0.558 & 0.840 \\
Claude Opus 5.5~\cite{anthropic2026opus55} & 0.925 & 0.900 & 0.912 & 0.264 & 0.331 & 0.294 & 0.603 & 0.844 \\
Claude Sonnet 5.5~\cite{anthropic2026sonnet55} & 0.914 & 0.886 & 0.900 & 0.183 & 0.236 & 0.206 & 0.553 & 0.822 \\
Claude Haiku 5.5~\cite{anthropic2026haiku55} & 0.923 & 0.702 & 0.798 & 0.145 & 0.464 & 0.221 & 0.509 & 0.679 \\
Gemini 3.8 Flash~\cite{deepmind2026gemini38flash} & \textbf{0.931} & 0.896 & 0.913 & 0.287 & 0.384 & 0.329 & 0.621 & 0.846 \\
\midrule
\rowcolor{black!6}\multicolumn{9}{l}{\textbf{\textit{General-purpose VLMs, Sequence style}}} \\
GPT-6.1 Sol~\cite{openai2026gpt61sol} & \textbf{0.969} & 0.598 & 0.740 & \textbf{0.736} & \textbf{0.244} & \textbf{0.366} & 0.553 & 0.563 \\
GPT-6 Luna~\cite{openai2026gpt6astra} & 0.954 & 0.419 & 0.582 & 0.442 & 0.176 & 0.252 & 0.417 & 0.395 \\
GPT-5.6 Terra~\cite{openai2026gpt56} & 0.953 & 0.394 & 0.557 & 0.679 & 0.129 & 0.217 & 0.387 & 0.368 \\
Claude Opus 5.5~\cite{anthropic2026opus55} & 0.948 & \textbf{0.849} & \textbf{0.895} & 0.697 & 0.237 & 0.354 & \textbf{0.624} & \textbf{0.789} \\
Claude Sonnet 5.5~\cite{anthropic2026sonnet55} & 0.932 & 0.769 & 0.843 & 0.529 & 0.112 & 0.184 & 0.514 & 0.705 \\
Claude Haiku 5.5~\cite{anthropic2026haiku55} & 0.935 & 0.542 & 0.686 & 0.322 & 0.075 & 0.122 & 0.404 & 0.496 \\
Gemini 3.8 Flash~\cite{deepmind2026gemini38flash} & 0.960 & 0.720 & 0.823 & 0.688 & 0.196 & 0.305 & 0.564 & 0.669 \\
\midrule
\rowcolor{black!6}\multicolumn{9}{l}{\textbf{\textit{Decision model}}} \\
Jev~1.14~\cite{typesafe2026jev} (Pair) & 0.916 & 0.743 & 0.820 & 0.138 & 0.350 & 0.198 & 0.509 & 0.704 \\
Jev~1.14~\cite{typesafe2026jev} (Sequence) & \textbf{0.929} & \textbf{0.751} & \textbf{0.831} & \textbf{0.183} & \textbf{0.427} & \textbf{0.256} & \textbf{0.543} & \textbf{0.719} \\
\bottomrule
\end{tabularx}
\end{table*}

\section{Unified Empirical Study}
\label{sec:experiments}

The study asks what each evaluator's output credits, at the endpoint and
under broken constraints (suites (A) and (B)) and as the task state
changes (C), and whether its evidence suffices (D), then turns to the query
protocol, task scope, and metrics that explain where these capabilities
part. Tables~\ref{tab:bench-main-results}
and~\ref{tab:bench-controls} and Fig.~\ref{fig:bench-results} report the
results, and Supplementary Section~\ref{sec:supp-study} gives settings
and per-model values.

\subsection{Evaluated Models}
\label{sec:exp-models}

Table~\ref{tab:bench-main-results} lists the 27 evaluators with their
sources. Its 19 specialized evaluators come from 14 methods whose native
outputs span all five families; Supplementary Section~\ref{sec:supp-settings} gives
their interfaces and settings. Robo-ValueRL takes only multiview input,
and six others are scored with it too (Section~\ref{sec:exp-evidence}).
Seven proprietary
general-purpose VLMs are queried in two styles with a fixed prompt:
Pair style presents two observations from one video, and Sequence
style the video's observations in temporal order. The
decision model Jev~1.14~\cite{typesafe2026jev} returns probabilities over fixed answers and is queried in both styles.

\subsection{Proxy Validity: Endpoint Selection and False Credit}
\label{sec:exp-proxy}

Endpoint judgments order executions more reliably than dense judgments track
change. Within each task whose executions differ in final
completion, the highest final score picks the most complete execution in
three quarters of cases or more for Sequence-queried GPT-6.1 Sol, Jev~1.14,
and Flora-Rewarder, against fewer than half at random (Supplementary
Section~\ref{sec:supp-proxy}). Final scores
also filter well: the top tenth of the executions, as ranked by Robometer,
RoboReward, or GPT-6.1 Sol, holds about twice the share of fully completed
executions that a random selection holds. RoboReward, with the lowest
falling F1 among specialized evaluators, filters this well: endpoint and
dense judgment are separate capabilities.

The failed executions of the Action--Effect and Semantic/Ordered
Constraint suites (Fig.~\ref{fig:bench-contrasts}(A)--(B)) divide the
evaluators. Under Flora-Rewarder, GPT-6.1 Sol, and Robometer, few reach the median final score of their task's completed executions;
under VLAC-Cut, TOPReward, and VLAC-8B, half or more do.
TOPReward also selects the most complete execution less often than
chance.

Broken constraints are the hardest case for dense judgments. On the
Semantic/Ordered Constraint suite, most evaluators in
Table~\ref{tab:bench-controls} score below the always-rising predictor on
Macro-F1, more than on any other suite (Supplementary
Section~\ref{sec:supp-proxy}). General-purpose VLMs recall falls after a drop or a release failure well above their overall rate of predicting a
fall, but barely exceed that rate for falls after a placement at a wrong destination. Specialized evaluators respond to wrong placements more
often, Robometer and VLAC-2B most clearly, yet each recalls at most half
of these falls. A drop looks like a setback, whereas a wrong placement
looks like progress unless its destination is checked against the
instruction, suggesting dense credit follows the appearance of
progress more closely than the instruction's constraints.

\begin{rrmfindings}
\item The strongest final scores select and filter whole executions well,
even from an evaluator that responds to few falls: endpoint judgment and
dense credit are separate capabilities.
\item Credit on the failed executions of suites (A) and (B) divides the
evaluators, and broken constraints are the hardest case: a wrong
placement looks like progress unless its destination is checked against
the instruction.
\end{rrmfindings}

\begin{figure*}[!t]
\centering
\includegraphics[width=0.96\linewidth]{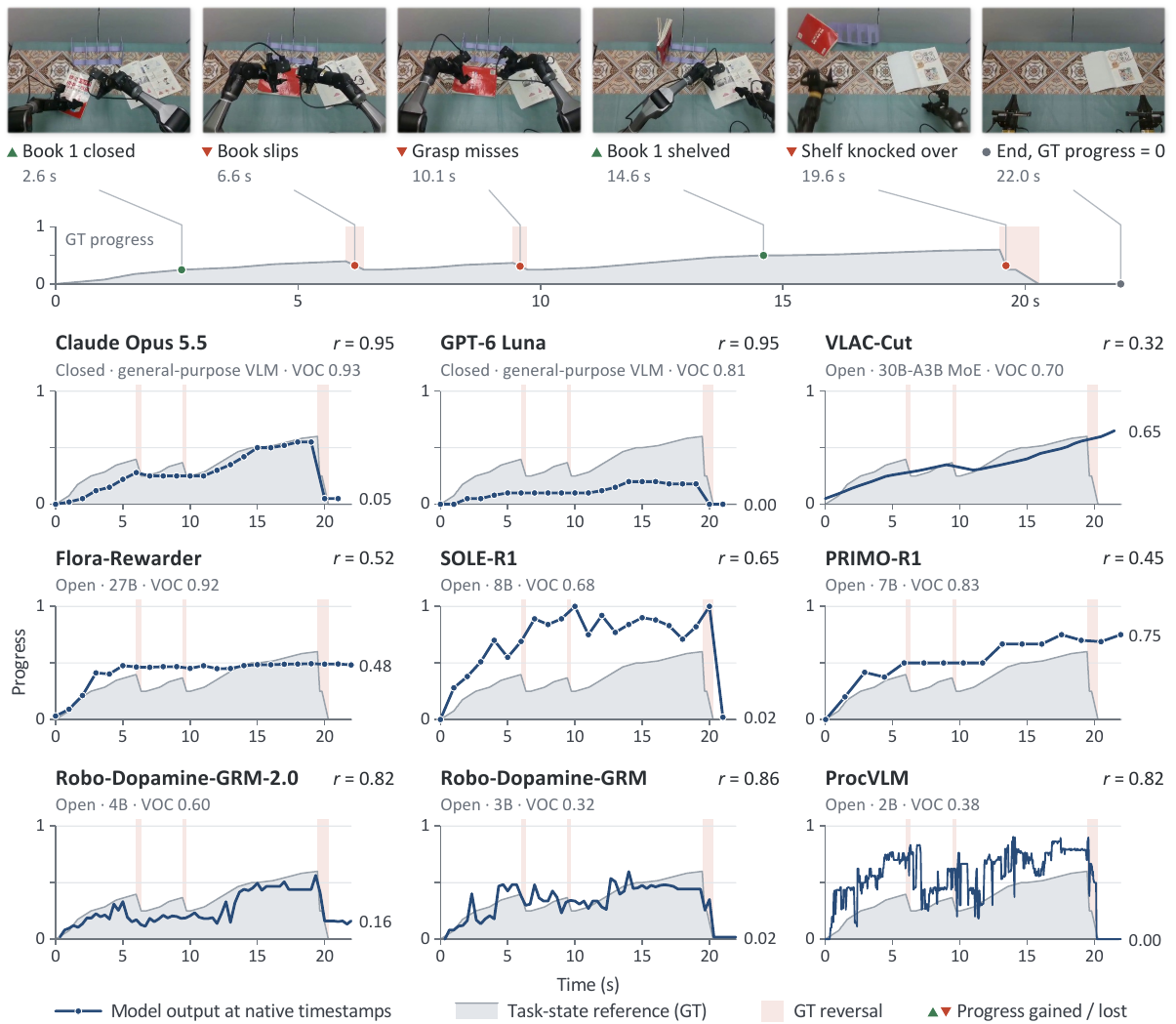}
\vspace{-1mm}
\caption{Reward profiles of nine evaluators on one execution of
put the books back. The task-state reference (GT, gray) drops to zero
when the robot knocks the shelf over at 19.6~s; red bands mark its
reversals, and $r$ is each model's Pearson correlation with it.}
\label{fig:bench-results}
\vspace{-3mm}
\end{figure*}

\subsection{Temporal Validity: Does Credit Track Task State?}
\label{sec:exp-temporal}

No evaluator tracks task-state falls reliably, and the responses take
three forms (Table~\ref{tab:bench-main-results}). RoboReward and
LRM-8B (Progress) leave their output unchanged on about three quarters of
the transitions; RoboReward's five grades are too coarse to register most
changes. Most specialized evaluators fluctuate: they predict falls more
often than the task state falls, and the median one predicts a fall on a
quarter of the transitions where the task state rises. Sequence-queried
VLMs rarely predict a fall and miss most falls.

Losses of an achieved condition, the cases of suite (C)
(Fig.~\ref{fig:bench-contrasts}(C)), are rare among the falls but
draw a clearer response, and
Persistence/Reversal is the suite on which the fewest evaluators fall below
the always-rising predictor. All VLM rows and most single-view
specialized evaluators recall lost achievements above their overall rate
of predicting a fall, and Robometer and Robo-Dopamine-GRM-2.0-8B-Preview
exceed that rate the most, recalling more than half of them. VLAC-Cut,
which leads the single-view specialized evaluators in falling F1, recalls few. In Fig.~\ref{fig:bench-results}, five of the nine evaluators
withdraw credit when the shelf falls, whereas PRIMO-R1 and VLAC-Cut
end higher than before.

Credit can also follow the action: where the action-history reference
advances but the task state does not, the three Pair-queried VLMs best in Macro-F1 predict more rises than where both are flat
(Supplementary Section~\ref{sec:supp-temporal}).

\begin{rrmfindings}
\item No evaluator tracks task-state falls reliably: outputs stay
unchanged, fluctuate with spurious falls, or rarely fall.
\item Lost achievements draw clearer responses than other falls, yet the
single-view specialized evaluator that detects falls best recalls few, so direction accuracy and sensitivity to lost achievements are separate
capabilities.
\end{rrmfindings}

\begin{table}[!t]
\centering
\caption{Conventional and direction metrics of the single-view evaluators. V and P/X give Macro-F1 under sufficient and
reduced camera evidence on the same transitions. Bold marks the best result in each panel. $^{\dagger}$Curve interpolated to
1~FPS.}
\label{tab:bench-controls}
\label{tab:bench-evidence-results}
\footnotesize
\setlength{\tabcolsep}{2.6pt}
\begin{tabularx}{\columnwidth}{@{}Yccccc@{}}
\toprule
& \multicolumn{2}{c}{Conventional} & \multicolumn{3}{c}{Direction} \\
\cmidrule(lr){2-3}\cmidrule(l){4-6}
Model & VOC & $\tau_a$ & Falling F1 & V & P/X \\
\midrule
\rowcolor{black!6}\multicolumn{6}{l}{\textbf{\textit{Specialized evaluators, 2B--4B}}} \\
ProcVLM-2B~\cite{DBLP:journals/corr/abs-2605-08774}$^{\dagger}$ & 0.382 & 0.175 & 0.201 & 0.469 & 0.407 \\
VLAC-2B~\cite{DBLP:journals/corr/abs-2509-15937}$^{\dagger}$ & 0.370 & 0.093 & 0.171 & 0.357 & 0.331 \\
Robo-Dopamine-GRM-3B~\cite{DBLP:journals/corr/abs-2512-23703} & 0.324 & 0.113 & 0.202 & 0.399 & 0.356 \\
Robo-Dopamine-GRM-2.0-4B-Preview~\cite{DBLP:journals/corr/abs-2512-23703} & 0.598 & 0.311 & \textbf{0.242} & 0.514 & 0.452 \\
Robometer-4B~\cite{DBLP:journals/corr/abs-2603-02115}$^{\dagger}$ & \textbf{0.832} & \textbf{0.473} & 0.214 & \textbf{0.582} & \textbf{0.527} \\
RoboReward-4B~\cite{DBLP:journals/corr/abs-2601-00675}$^{\dagger}$ & 0.807 & 0.314 & 0.072 & 0.254 & 0.189 \\
RynnValue-4B~\cite{huang2026rynnvalue}$^{\dagger}$ & 0.459 & 0.419 & 0.235 & 0.572 & 0.450 \\
\rowcolor{black!6}\multicolumn{6}{l}{\textbf{\textit{Specialized evaluators, 7B--8B}}} \\
PRIMO-R1-7B~\cite{DBLP:journals/corr/abs-2603-15600}$^{\dagger}$ & \textbf{0.828} & 0.285 & 0.154 & 0.431 & 0.426 \\
LRM-8B~\cite{DBLP:journals/corr/abs-2603-16065} (Progress) & 0.198 & 0.157 & 0.136 & 0.158 & 0.215 \\
Robo-Dopamine-GRM-8B~\cite{DBLP:journals/corr/abs-2512-23703} & 0.389 & 0.263 & 0.211 & 0.442 & 0.388 \\
Robo-Dopamine-GRM-2.0-8B-Preview~\cite{DBLP:journals/corr/abs-2512-23703} & 0.605 & 0.280 & 0.226 & 0.504 & 0.447 \\
RynnValue-8B~\cite{huang2026rynnvalue}$^{\dagger}$ & 0.522 & \textbf{0.390} & \textbf{0.236} & \textbf{0.570} & 0.441 \\
SOLE-R1-8B~\cite{DBLP:journals/corr/abs-2603-28730} & 0.683 & 0.196 & 0.225 & 0.494 & \textbf{0.498} \\
TOPReward~\cite{DBLP:journals/corr/abs-2602-19313} (Qwen3-VL-8B)$^{\dagger}$ & 0.379 & 0.081 & 0.182 & 0.434 & 0.427 \\
VLAC-8B~\cite{DBLP:journals/corr/abs-2509-15937}$^{\dagger}$ & 0.414 & 0.120 & 0.169 & 0.366 & 0.358 \\
\rowcolor{black!6}\multicolumn{6}{l}{\textbf{\textit{Specialized evaluators, 27B--30B}}} \\
Flora-Rewarder-27B~\cite{leng2026flora} & \textbf{0.915} & \textbf{0.572} & 0.252 & 0.551 & 0.511 \\
VLAC-Cut-30B-A3B~\cite{zhai2026help} & 0.695 & 0.261 & \textbf{0.272} & \textbf{0.588} & \textbf{0.561} \\
\rowcolor{black!6}\multicolumn{6}{l}{\textbf{\textit{Specialized evaluators, proprietary backbone}}} \\
GVL~\cite{DBLP:conf/iclr/MaHFSLXK0DXBJ0Z25} (Gemini 3.8 Flash) & 0.788 & 0.271 & 0.240 & 0.553 & 0.506 \\
\midrule
\rowcolor{black!6}\multicolumn{6}{l}{\textbf{\textit{General-purpose VLMs, Sequence style}}} \\
GPT-6.1 Sol~\cite{openai2026gpt61sol} & 0.898 & \textbf{0.695} & \textbf{0.366} & 0.559 & 0.491 \\
GPT-6 Luna~\cite{openai2026gpt6astra} & 0.805 & 0.444 & 0.252 & 0.466 & 0.337 \\
GPT-5.6 Terra~\cite{openai2026gpt56} & 0.878 & 0.632 & 0.217 & 0.389 & 0.294 \\
Claude Opus 5.5~\cite{anthropic2026opus55} & \textbf{0.929} & 0.646 & 0.354 & \textbf{0.618} & \textbf{0.557} \\
Claude Sonnet 5.5~\cite{anthropic2026sonnet55} & 0.889 & 0.559 & 0.184 & 0.493 & 0.474 \\
Claude Haiku 5.5~\cite{anthropic2026haiku55} & 0.807 & 0.332 & 0.122 & 0.393 & 0.369 \\
Gemini 3.8 Flash~\cite{deepmind2026gemini38flash} & 0.891 & 0.601 & 0.305 & 0.560 & 0.531 \\
\midrule
\rowcolor{black!6}\multicolumn{6}{l}{\textbf{\textit{Decision model, Sequence style}}} \\
Jev~1.14~\cite{typesafe2026jev} & 0.905 & 0.496 & 0.256 & 0.534 & 0.525 \\
\bottomrule
\end{tabularx}
\end{table}

\subsection{Evidence Validity: Does the View Matter?}
\label{sec:exp-evidence}

Hiding the decisive evidence (Fig.~\ref{fig:bench-contrasts}(D)) lowers
Macro-F1 for nearly every evaluator in
Table~\ref{tab:bench-evidence-results}, and no evaluated interface can report insufficient evidence. Views with no task evidence at all reveal whether an evaluator knows it cannot judge. Most
Pair-queried evaluators still report progress on them, close to the
always-rising predictor, whereas Sequence-queried VLMs hold their output
nearly constant, which a direction test against the hidden task state
counts as error. An unseen state is neither success nor failure and needs an output and a score of its own (Section~\ref{sec:future-evidence}).

Extra cameras help evaluators with separate camera slots.
With the video's camera in the primary slot and its execution's other cameras in the rest ($^{\ddagger}$ in Table~\ref{tab:bench-main-results}),
all four Robo-Dopamine checkpoints and SOLE-R1 gain. With three
synchronized cameras, a reduced primary view plus the other two beats the sufficient view alone for all five, so the others recover what the primary view hides. Flora-Rewarder, given views tiled in one frame, does not gain. Judgment also depends on the primary camera: swapping it changes the predicted direction more often than rerunning the input (Supplementary
Section~\ref{sec:supp-evidence}).

\begin{rrmfindings}
\item Hiding the decisive view lowers Macro-F1 for nearly every evaluator,
and where no view shows the task, most Pair-queried evaluators still
report progress: no evaluated interface can report insufficient evidence.
\item Evaluators with separate camera slots recover what a reduced view hides, yet their judgment changes with the primary camera.
\end{rrmfindings}

\subsection{Query Protocol, Scope, and Conventional Metrics}
\label{sec:exp-legacy}

The query is part of the evaluator. Switching from Pair to Sequence
queries raises falling precision and lowers falling recall for all seven
VLMs, Gemini 3.8 Flash's Macro-F1 rises from the GVL protocol to Sequence
and then to Pair queries, and the two styles of Jev~1.14 agree on about
two thirds of the transitions (Table~\ref{tab:bench-main-results}). A Pair
query sees two observations and a Sequence query the whole sampled video,
so context and output granularity change together; the results are
consistent with Pair queries exposing local change, real or spurious, and
Sequence queries smoothing it over the video.

Evaluators also differ in respecting the task scope. When a
scope variant ends complete while the full task is unfinished, Sequence-queried VLMs give it more credit than the full task at the same
moment in most pairs, whereas the Robo-Dopamine models, VLAC-Cut, and
SOLE-R1 do so in fewer than half (Supplementary
Section~\ref{sec:supp-metrics}).

VOC rewards chronology. The normalized-time control reads only elapsed time, yet its VOC of 1.000 exceeds every evaluator's, while its
$\tau_a$ and falling F1 are zero. Learned evaluators show the same
dissociation (Table~\ref{tab:bench-controls}). RoboReward pairs one of the
highest VOCs among specialized evaluators with the lowest falling F1, and Robometer and PRIMO-R1 follow suit. Across the single-view
specialized evaluators, VOC and falling F1 are only weakly related, far
less than across the VLMs. $\tau_a$ measures endpoint ranking and also orders the
specialized evaluators differently from falling F1, so neither metric
replaces a test of credit withdrawal. Outputs themselves lean toward elapsed time: on executions whose task state is non-monotone or ends below
completion, nearly two thirds of evaluators correlate more with elapsed time than with task state on most of them. Because nine in ten transitions rise,
the always-rising predictor beats half of the single-view specialized
evaluators on Macro-F1, so the falling class carries the discrimination.

\begin{rrmfindings}
\item Switching from Pair to Sequence queries trades falling recall for
falling precision, so the query is part of the evaluator.
\item VOC rewards chronology: a control that reads only elapsed time
exceeds every evaluator on it, so credit withdrawal needs a test of its
own.
\end{rrmfindings}

\subsection{Summary of Findings}
\label{sec:exp-summary}

Three findings separate capabilities that a single leaderboard merges.
\textbf{Endpoint judgment and dense credit:} RoboReward filters among the
best and responds to the fewest falls, whereas VLAC-Cut leads the
single-view specialized evaluators in falling F1 yet gives success-level credit to
half the failed executions of suites (A) and (B), so which evaluator to use
depends on the decision it serves. \textbf{Execution progression and task
fulfillment:} wrong placements, which look like progress unless checked
against the instruction, make broken constraints the hardest case for
dense judgments, and most evaluators follow elapsed time more than
task state where the two diverge; the procedure and constraint checks of
Section~\ref{sec:models-progress} target this gap. \textbf{Evidence and
its use:} no evaluator signals that the decisive view is missing, extra
cameras help where the interface fuses them, and the query style moves results by up to 0.17 Macro-F1.

\section{Challenges and Research Directions}
\label{sec:future}

Scale alone does not close the gaps of Section~\ref{sec:exp-summary}:
within a model series, the larger checkpoint scores much like the smaller
one (Table~\ref{tab:bench-main-results}). The gaps trace instead to what
evaluators learn from, how they represent the task, how they are queried,
what they observe, and how they are measured and used, and the directions
below follow this order.

\subsection{Learning from Real Failures}
\label{sec:future-failures}
Detecting drops and recognizing wrong destinations are different abilities: VLMs register the first far more often than the second
(Section~\ref{sec:exp-proxy}). Most task-state falls end a failed attempt
or follow a drop (Section~\ref{sec:bench-goal}), whereas reward models
learn mostly from successful demonstrations and synthetic negatives
(Section~\ref{sec:data-construction}).
RoboMIND~\cite{DBLP:journals/corr/abs-2412-13877} annotates real teleoperation failures by cause, and
VLAC-Cut~\cite{zhai2026help}, trained with signed progress on executions with failure, regression, and recovery, leads single-view specialized evaluators in falling F1 but recalls few lost
achievements (Section~\ref{sec:exp-temporal}). Closed-loop platforms and deployment studies already log such failures at scale
\cite{DBLP:journals/corr/abs-2510-17950,DBLP:journals/corr/abs-2506-18123,DBLP:journals/corr/abs-2606-08881}.
The open question is what supervision each failure mechanism needs: signed targets letting credit fall, labels by cause (failed attempt,
drop, wrong destination, lost achievement), and constraint labels for errors motion does not reveal. A gain would show as per-cause fall recall above the evaluator's fall rate, without more false falls
where the task advances.

\subsection{Task Structure Beyond the Instruction}
\label{sec:future-structure}
An instruction names the goal but rarely the phases that reach it or the
constraints that make it correct. Evaluators that register a drop still
miss a placement at the wrong destination (Fig.~\ref{fig:bench-contrasts}(B),
Section~\ref{sec:exp-proxy}), and goal-distance anchors can read necessary
detours as regressions (Section~\ref{sec:models-progress}). Structure has
entered reward models as stage-decomposed rewards and values
\cite{victor2024,DBLP:journals/corr/abs-2509-25358,DBLP:journals/corr/abs-2606-10305,stdr2026},
as stage reasoning before scoring
\cite{DBLP:journals/corr/abs-2603-28730,DBLP:journals/corr/abs-2605-08774,r2vlm2026},
and as post-conditions Thea~\cite{wang2026thea} checks at runtime.
Evaluators that state the current stage before scoring still predict a
fall on about a third of transitions where the task advances
(Supplementary Section~\ref{sec:supp-temporal}). The question is how to
combine judgments of physical state change with task-constraint checks. A structured specification of a task's units, constraints,
and order can condition progress within phases such as approach,
transport, and release, while post-condition checks verify what motion
does not reveal, such as the destination or required effect
(Fig.~\ref{fig:bench-contrasts}(A)--(B)). Lower false credit on suites (A)
and (B) at unchanged endpoint ranking would measure the gain.

\subsection{Query Design}
\label{sec:future-query}
Pair queries recall more falls with more false alarms, Sequence queries
fewer with fewer, and the same backbone scores differently under each
protocol (Section~\ref{sec:exp-legacy}). Existing work varies the query:
GVL~\cite{DBLP:conf/iclr/MaHFSLXK0DXBJ0Z25} asks for the completion of
shuffled frames, TOPReward~\cite{DBLP:journals/corr/abs-2602-19313} reads
token probabilities, and RoboRMBench~\cite{jeung2026robormbench} tests
consistency under paraphrased instructions. The prompt, query style, frame
selection, and output format therefore belong to an evaluator's
specification (Section~\ref{sec:models-crosscutting}) and to every
reported result. The research question is how local event comparison and
long-horizon task state can combine, as in protocols that compare adjacent observations within whole-video context, or prompts
that spell out a task's structure and test whether the evaluator uses it.
Continuous, calibrated outputs avoid coarse ordinal levels that miss
most changes. A protocol that combines the falling recall of Pair queries
with the falling precision of Sequence queries would be an advance.

\subsection{Evidence Sufficiency and Abstention}
\label{sec:future-evidence}
Reduced views lower Macro-F1 for nearly every evaluator, and where no view
shows the task, most Pair-queried evaluators still report progress
(Section~\ref{sec:exp-evidence}). An evaluator must recognize when it
cannot judge, since a judgment without evidence misleads filtering, shaping, and monitoring. Three capabilities supply this.
\emph{Abstention} reports insufficient evidence as a state separate from a
low score; the closest existing interface, the unanswerable response of
ProgressLM~\cite{progresslm2026}, flags a mismatch with the reference
demonstration rather than a missing view. \emph{Calibration} lets
confidence fall with the evidence:
WARP-RM~\cite{DBLP:journals/corr/abs-2606-28320} predicts a distribution
over progress and RARM~\cite{DBLP:journals/corr/abs-2606-22027} gates
uncertain matches, but neither has been tested against what the camera shows. \emph{Active evidence acquisition} selects, fuses, or
requests the decisive view: AnyviewMeter~\cite{anyviewmeter2026} adapts
Robometer~\cite{DBLP:journals/corr/abs-2603-02115} to multiview input via camera-ray conditioning,
Robo-Dopamine~\cite{DBLP:journals/corr/abs-2512-23703} reads synchronized
views, and extra cameras help where the interface fuses them. Training
with variable camera sets and missing views builds these capabilities, and per-view evidence labels, as in RRM-Bench-1K, make them testable. An
advance would show as abstention without a decisive view, confidence following the evidence label, and multiview gains above a vote over single views from whichever camera fills the primary slot.

\subsection{Measuring What Matters}
\label{sec:future-metrics}
VOC rewards chronology, and the always-rising predictor outscores half of
the single-view specialized evaluators on Macro-F1, because elapsed time
and task state mostly advance together (Section~\ref{sec:exp-legacy}).
Learned evaluators with a high VOC can keep their credit when the task
state collapses (Fig.~\ref{fig:bench-results}).
Progress metrics therefore need the normalized-time and always-rising
references beside them, the falling class in its own column, and credit
withdrawal tested per event, with $\tau_a$ kept as the measure of endpoint
ranking. Test sets need executions in which time and task state diverge,
such as failures, partial completions, and lost achievements, labeled from
what physically happened. Views without task evidence need their own
scoring: there an abstention or an unchanged output is correct, and a
score against the hidden task state penalizes the evaluator that
recognizes it cannot judge. RRM-Hub reports these capabilities separately,
so a new evaluator can show which of them it improves.

\subsection{From Evaluators to Decisions}
\label{sec:future-decisions}
Task-state fidelity and decision utility are different properties, because
each decision rule (Section~\ref{sec:models-pipeline}) consumes a
different capability. Selection among candidate actions or executions and
filtering of data need only the order of whole executions, which the strongest evaluators provide, including some that respond to few
falls (Section~\ref{sec:exp-proxy}). Monitoring needs failures detected as
they occur, and recovery needs to know whether a setback can still be
repaired, which a failure forecast or \(V^\pi\) answers and current-state
credit does not. Completion, time to go, and the chance of future success
respond differently to a setback (Section~\ref{sec:models-value}) and
serve acceptance, scheduling, and intervention, respectively. Shaping makes the strongest demand, since the policy optimizes against the reward and seeks executions where motion, time, or appearance
part from the task state, the cases of suites (A)--(C); optimization
against a frozen reward can raise both task success and wrong-object failures~\cite{critichack2026}. The fidelity a decision needs grows
with the optimization pressure it puts on the reward. Shaping and
monitoring need credit that falls with the task state, including at
placements that violate the instruction, and every rule must recognize missing evidence; no evaluated model meets these demands yet
(Sections~\ref{sec:exp-proxy}--\ref{sec:exp-evidence}). Joint models keeping these outputs separate and consistent can serve several rules.

Evaluators already act inside the policy loop, as a causal failure-risk
head in FIERCE~\cite{fierce2026}, a candidate verifier in
DiVeR~\cite{diver2026}, and an intervention gate in
Reward-DAgger~\cite{rewarddagger2026}, where a strong candidate pool can
mask a weak verifier (Section~\ref{sec:models-preference}). The offline
selection and filtering tests of Section~\ref{sec:exp-proxy} lead to
their closed-loop counterparts: candidate selection, rejection of
known-bad behavior, and controlled policy updates under deployment
conditions, including control deadlines, with each evaluator chosen for
its role and tested with its rule. Full policy optimization comes last,
because it places the most pressure on the reward.

\section{Conclusion}
\label{sec:conclusion}

In the foundation-model era, reward names several questions rather than
one quantity. This survey organizes robotic reward models into five
semantic families (Outcome, Absolute Progress, Relative, Temporal Value,
and Hybrid) by the question their native outputs answer and relates each
design route to the quantity it estimates, its supervision, and its
evidence. Relying on such an evaluator takes three answers: what its output
credits, whether its evidence suffices, and which decisions it can drive.
RRM-Bench-1K tests the first two on whole real-robot executions, with
contrast suites that separate the credited condition from its usual cues,
task-state labels that withdraw lost credit, and camera views labeled by
the evidence they show. The strongest final scores rank executions well
enough for selection and filtering, but dense credit follows the
progression of an execution more closely than its task state, VOC hides
this gap because it rewards chronology, and no evaluator signals when the
decisive view is missing. The query protocol and the evidence interface
change these results and so belong to the evaluator. The third answer
depends on the decision: the fidelity a rule needs grows with the
optimization pressure it places on the reward, so dense shaping and
monitoring await evaluators that withdraw credit when an achieved
condition is lost, verify constraints that motion does not reveal, and
abstain without the decisive view. The path runs through real failures in
training data, structured task specifications and queries, evidence-aware
interfaces, and metrics that separate task state from chronology, and
RRM-Hub lets new evaluators be tested on each of these capabilities. A
policy learns what its reward credits, so general-purpose robots will be
only as reliable as their evaluators.

{
    \small
    \bibliographystyle{IEEEtran}
    \bibliography{dblp,new_papers_2026_10}
}

\makeatletter
\def\@textbottom{\vskip\z@\@plus 1fill}\let\@texttop\relax
\makeatother

\patchcmd{\IEEEbiographynophoto}%
  {\vskip 4\baselineskip plus 1fil minus 0\baselineskip}%
  {\vskip 0.5\baselineskip plus 1fil minus 0\baselineskip}%
  {}{\PackageWarning{main}{biography spacing patch failed}}

\balance

\begin{IEEEbiographynophoto}{Guoqiang Liang}
is currently a Ph.D. student at Show Lab, National University of Singapore
(NUS). Liang's research interests include visual generation and embodied AI.
\end{IEEEbiographynophoto}

\begin{IEEEbiographynophoto}{Yiqi Lin}
is currently a Ph.D. student at Show Lab, NUS. Lin's research interests
include multimodal learning, video generation and editing,
vision--language models, and robot learning.
\end{IEEEbiographynophoto}

\begin{IEEEbiographynophoto}{Shangchen Zhou}
received the Ph.D. degree in computer science from Nanyang Technological
University (NTU), and then joined MMLab@NTU as a Research Assistant Professor. Zhou's research interests include image and video enhancement, generation, and editing.
\end{IEEEbiographynophoto}

\begin{IEEEbiographynophoto}{Wenzheng Zeng}
is currently a Ph.D. student at Show Lab, NUS. Zeng's research interests include video understanding, multimodal foundation models, and their applications to robotics.
\end{IEEEbiographynophoto}

\begin{IEEEbiographynophoto}{Yanzhe Chen}
is currently a Ph.D. student at Show Lab, NUS. Chen's research interests
include agents, robotics, and multimedia.
\end{IEEEbiographynophoto}

\begin{IEEEbiographynophoto}{Zechen Bai}
is currently a Ph.D. candidate at Show Lab, NUS. Bai's research interests
include multimodal foundation models and agents.
\end{IEEEbiographynophoto}

\begin{IEEEbiographynophoto}{Mike Zheng Shou}
received the Ph.D. degree in electrical engineering from Columbia
University, and was a Research Scientist at Facebook AI, before joining NUS, where Shou is currently an Associate
Professor and leads Show Lab. Shou is a Fellow of the National Research
Foundation (NRF) Singapore, serves as an Associate Editor of IEEE TPAMI,
and has served as an Area Chair for CVPR, ECCV, ICCV, and ACM MM. Shou's
research interests include computer vision and multimodal learning.
\end{IEEEbiographynophoto}

\onecolumn
\flushbottom
\section*{Supplementary Material}

\setcounter{section}{0}
\renewcommand{\thesection}{S\arabic{section}}
\renewcommand{\theHsection}{S\arabic{section}}
\setcounter{table}{0}
\renewcommand{\thetable}{S\arabic{table}}
\renewcommand{\theHtable}{S\arabic{table}}
\setcounter{figure}{0}
\renewcommand{\thefigure}{S\arabic{figure}}
\renewcommand{\theHfigure}{S\arabic{figure}}
\section{Method Inventory}
\label{sec:supp-inventory}
\begingroup
\makeatletter
\def\LT@makecaption#1#2#3{%
  \LT@mcol\LT@cols c{\hbox to\z@{\hss\parbox[t]\LTcapwidth{%
    \reset@font\footnotesize\sffamily\centering
    #1{#2\\}%
    #3%
    \endgraf\vskip0.5\baselineskip}%
  \hss}}}
\makeatother
\footnotesize
\setlength{\tabcolsep}{3pt}
\renewcommand{\arraystretch}{1.08}
\setlength{\LTcapwidth}{\textwidth}
\setlength{\LTleft}{0pt}
\setlength{\LTright}{0pt}
\begin{longtable}{@{}L{\dimexpr.185\textwidth-\tabcolsep\relax}L{\dimexpr.145\textwidth-2\tabcolsep\relax}L{\dimexpr.22\textwidth-2\tabcolsep\relax}L{\dimexpr.225\textwidth-2\tabcolsep\relax}L{\dimexpr.225\textwidth-\tabcolsep\relax}@{}}
\caption{Comparison of 114 robotic reward methods and evaluative protocols, one evaluative component per row, grouped by semantic family in the order of Section~\ref{sec:rrm_families} and ordered by first release within each family. $^{\ast}$~evaluates generated behavior; $^{\ddagger}$~marks a protocol or harness. Query abbreviations and component notes follow the table.}\label{tab:rrm-method-comparison}\\
\toprule
Method / component & Native output & Query evidence & Learning mechanism & Training / data source \\
\midrule
\endfirsthead
\multicolumn{5}{c}{\sffamily\tablename~\thetable\ (continued)}\\[3pt]
\toprule
Method / component & Native output & Query evidence & Learning mechanism & Training / data source \\
\midrule
\endhead
\midrule
\multicolumn{5}{r}{\footnotesize Continued on the next page.}\\
\endfoot
\bottomrule
\endlastfoot
\rowcolor{famO!18}\multicolumn{5}{l}{\rrmfamily{OUTCOME (12 methods)}} \\*
\addlinespace[3pt]
ZSRM (binary variant) \cite{pmlr-v162-mahmoudieh22a} & Binary goal condition & Image + Task & CLIP adaptation + spatial labels & Exploration images + descriptions \\
SuccessVQA (Flamingo detector) \cite{du2023successvqa} & Binary success answer & Video + Task & Flamingo VQA fine-tuning & Human success annotations \\
RoboCLIP \cite{DBLP:conf/nips/SontakkeZAPBSFI23} & Episode compatibility & Traj. + Ref./Task & Frozen video--text alignment & No RRM training \\
RoboFuME (reward classifier) \cite{robofume2023} & Binary completion & Image + Task & MiniGPT-4 fine-tuning & Demo endpoints + unsuccessful images \\
Sentinel (VLM monitor)$^{\ddagger}$ \cite{agia2024sentinel} & Failure forecast + rationale & Video + Task (+ Ref./Goal) & Prompted chain-of-thought video QA & Frozen VLMs; no training \\
SAFE \cite{gu2025safe} & Running failure score & Latent + Hist. & MLP/LSTM probe + functional conformal prediction & Successful/failed VLA rollouts \\
I-FailSense \cite{DBLP:journals/corr/abs-2509-16072} & Binary success/failure & Video + Task & Failure classification & Demos + synthetic data \\
RoboReward \cite{DBLP:journals/corr/abs-2601-00675} & End-episode progress (1--5) & Video + Task & Progress supervision & Human-scored rollouts + augmented demos \\
World-VLA-Loop$^{\ast}$ \cite{DBLP:journals/corr/abs-2602-06508} & Success score & Video + Init. + Action + Future & Reward regression & Demos + binary outcomes \\
VLAConf \cite{vlaconf2026} & Calibrated success confidence & Latent + Hist. & Support estimation + calibration & Success demos + outcome labels \\
OpenVLA warning probe~\cite{mahato2026openvlafailure} & Failure probability & Latent & Failure classification & Rollout latents + outcomes \\
FAVOR (Anchor Verifier) \cite{favor2026} & Deviation-event probability & Image + Ref. + Act.\,hist. & DINOv2 + local labels + episode-level multiple-instance learning & Clean/perturbed rollouts + outcomes \\
\midrule
\rowcolor{famAP!18}\multicolumn{5}{l}{\rrmfamily{ABSOLUTE PROGRESS (29 methods)}} \\*
\addlinespace[3pt]
VIP \cite{DBLP:conf/iclr/MaSJBK023} & Goal distance & Image + Goal & Implicit value & Human video \\
LIV \cite{DBLP:conf/icml/MaK0BJ23} & Value & Image + Task + Goal & Language--image value & Human video \\
VLM-RM \cite{DBLP:conf/iclr/RocamondeMNPL24} & Compatibility & Image + Task & Frozen image--text similarity & No RRM training \\
Adapt2Reward \cite{DBLP:conf/eccv/YangCQWWLGH24} & Compatibility & Video + Task & Failure-prompt alignment & Human + robot-failure video \\
Video-Language Critic \cite{DBLP:journals/tmlr/AlakuijalaMWFKM25} & Clip value & Video + Task & Contrastive temporal rank & Auto time labels \\
GVL \cite{DBLP:conf/iclr/MaHFSLXK0DXBJ0Z25} & Completion \% & Video + Task & Prompted frame order & Frozen VLM \\
PROGRESSOR \cite{DBLP:conf/iccv/AyalewZWJMW25} & Progress distribution & Image + Init. + Goal & Progress regression + RL & Demonstrations \\
ReWiND \cite{DBLP:journals/corr/abs-2505-10911} & Progress & Video + Task & Progress + rewind regression & Demonstrations + Open X-Embodiment \\
VITA \cite{vita2025} & Normalized progress & Image + Task + Hist. & CLIP adaptation + regression & Demonstration time labels \\
SPARTA-L (reward) \cite{sparta2025} & Object-state progress & Video + Task & VLM region labels + tracking & No RRM training \\
World-Env (RehearseVLA)$^{\ast}$ \cite{DBLP:journals/corr/abs-2509-24948} & Completion probability & Video + Task + Future & Success classification & Simulator frames \\
ProgressLM \cite{progresslm2026} & Progress / unanswerable & Image + Ref./Task & VLM SFT + GRPO & ProgressLM-45K \\
TOPReward \cite{DBLP:journals/corr/abs-2602-19313} & Progress (log-prob) & Video + Task & Truth-token probability & Frozen video VLM \\
Rewarding DINO \cite{DBLP:journals/corr/abs-2603-16978} & Reward & Multi-view + Task & Reward ranking & Analytic labels + frozen DINOv3 \\
R$^2$VLM \cite{r2vlm2026} & Completion \% + rationale & Video + Task + Hist. & Qwen2.5-VL SFT + multi-turn RL & ALFRED + Ego4D process labels \\
ProgressVLA \cite{DBLP:journals/corr/abs-2603-27670} & Progress & Image + Init. + Task + Future & Temporal regression & Robot video \\
SOLE-R1 \cite{DBLP:journals/corr/abs-2603-28730} & Progress + rationale & Video + Task + Init. & SFT + RLVR & Synthetic process reasoning \\
ProcVLM \cite{DBLP:journals/corr/abs-2605-08774} & Progress + proc. rationale & Video + Task & Procedure-grounded regression & Demos + synthetic labels \\
Feat2Go \cite{feat2go2026} & Binned structural progress & Image + Task & Qwen3-VL progress learning & Subgoal/feature-derived labels \\
NavWAM \cite{DBLP:journals/corr/abs-2606-13494} & Progress value & Image + Goal + Proprio. & Joint progress diffusion & Simulated + real navigation \\
RARM \cite{DBLP:journals/corr/abs-2606-22027} & Confidence-gated progress & Video + Reference & Temporal contrast & Unlabeled video \\
ReTVL \cite{DBLP:journals/corr/abs-2606-24633} & Value & Video + Task & Temporal + retry supervision & Outcomes + keypoints \\
WVM \cite{DBLP:journals/corr/abs-2606-24742} & Progress sequence & Video + Task & Progress-value learning & Robot + human video \\
WARP-RM \cite{DBLP:journals/corr/abs-2606-28320} & Progress (+ velocity) & Video & Warp-consistent progress & Successful robot video \\
LLM-as-a-Verifier (VLM verifier)$^{\ddagger}$ \cite{llmverifier2026} & Progress (score-token expectation) & Video (pair) + Task & Prompted score-token expectation + repetition & Frozen VLM (Qwen 3.6); no training \\
DenseReward \cite{DBLP:journals/corr/abs-2607-13033} & Progress & Video + Task & Per-step regression & Synthetic failures + outcomes \\
AnyviewMeter \cite{anyviewmeter2026} & Task progress & Video + Task + Geometry (+ Multi-view) & Robometer LoRA + ray conditioning & Labeled multiview trajectories \\
HiRE (potential) \cite{hire2026} & Success/failure support potential & Image + Ref. & Frozen DINOv2/SigLIP kernel score & Success frames + failed endpoints \\
ARS$^{\ddagger}$ \cite{ars2026} & Per-frame progress & Traj. + Task (+ Ref.) & VLM event proposal + verification & Frozen VLM; optional outcome labels \\
\midrule
\rowcolor{famR!18}\multicolumn{5}{l}{\rrmfamily{RELATIVE (23 methods)}} \\*
\addlinespace[3pt]
RL-VLM-F \cite{DBLP:conf/icml/WangSZXBHE24} & Scalar task utility & Image & Bradley--Terry preference & VLM pair labels \\
DecisionNCE-T \cite{decisionnce2024} & Transition-direction score & Pair + Task & CLIP implicit-preference learning & EPIC-KITCHENS video + language \\
Rank2Reward (ranking head) \cite{rank2reward2024} & Temporal-order utility & Image & Bradley--Terry on R3M features & Ordered passive-video frames \\
VLP (preference scorer) \cite{vlp2025} & Trajectory utility & Video + Task & CLIP cross-modal preference learning & Auto-generated video orderings \\
VARP (reward head) \cite{varp2025} & Preference utility & State & Preference learning + policy reg. & VLM trajectory-sketch labels \\
RoboMonkey \cite{DBLP:journals/corr/abs-2506-17811} & Rank & Image + Task + Action & Preference ranking & Synthetic action pairs \\
TimeRewarder \cite{DBLP:journals/corr/abs-2509-26627} & Signed temporal distance & Pair & Timestamp self-supervision & Robot + human video \\
RoVer \cite{DBLP:journals/corr/abs-2510-10975} & Rank + direction & Image + Task + Action & Ranking + direction & Synthetic pairs \\
EVE (selection verifiers)$^{\ddagger}$ \cite{eve2025} & Candidate selection & Image + Task + Action & Prompted specialist VLM selection & Frozen VLMs; no verifier fitting \\
Robo-Dopamine (incl.\ 2.0) \cite{DBLP:journals/corr/abs-2512-23703,robodopamine2_2026} & Relative progress & Pair + Multi-view + Task + Ref.\ (+ Hist.\ in 2.0) & Hop-label regression + signed-Hop replay (2.0) & Human keyframes + auto pairs + synthetic negatives (2.0) \\
CoVer \cite{DBLP:journals/corr/abs-2602-12281} & Alignment score & Image + Task + Act.\,hist. + Action & InfoNCE alignment & Demos + rephrases \\
MVR \cite{DBLP:journals/corr/abs-2603-01694} & State relevance utility & Privileged simulator state & Bradley--Terry order + embed. reg. & Frozen ViCLIP \\
GIFT (transferred reward) \cite{gift2026} & Intent-conditioned utility & Traj. + Criterion & GPT-4o alignment + MaxEnt IRL & Preferred/non-preferred demos \\
TapSampling \cite{DBLP:journals/corr/abs-2605-25547} & Progress delta & Image + Task + Action & Temporal delta & Ordered + reversed pairs \\
Small Stuff (pair critic) \cite{smallstuff2026} & Pairwise progress judgment & Pair + Task & VLM pairwise fine-tuning & Successful/failed robot rollouts \\
FPL \cite{DBLP:journals/corr/abs-2606-32027} & Criterion-cond. utility & Trajectory + Criterion & Preference utility & Human freeform pairs \\
VLM-AR3L (reward heads) \cite{vlmar3l2026} & Utility / pair preference & Image / Image pair & Bradley--Terry + Siamese learning & VLM preference labels \\
Dream2Reward \cite{dream2reward2026} & Signed transition alignment & Pair + Task + Hist. & Frozen DINOv3/Qwen3 + flow field & Positive demonstrations \\
Listwise VLM reward \cite{listwisevlm2026} & Preference utility & Image & Plackett--Luce reward learning & VLM-ranked robot behavior \\
RAPID (reward head) \cite{rapid2026} & Preference utility & Image & Bradley--Terry + adaptive queries & VLM-labeled rollout pairs \\
PreferenceFlow (reward) \cite{preferenceflow2026} & Action preference utility & Latent + Action & Preference + gradient regularization & Intervention pairs + expert demos \\
TeV (energy head) \cite{tev2026} & Action energy & Video + Task + Act.\,hist. + Action & Temporal contrastive ranking & Demos + constructed negatives \\
DiVeR (discriminator) \cite{diver2026} & Candidate utility & Latent + Action & Criticality-weighted classifier & Successful/failed VLA rollouts \\
\midrule
\rowcolor{famTV!18}\multicolumn{5}{l}{\rrmfamily{TEMPORAL VALUE (26 methods)}} \\*
\addlinespace[3pt]
V-PTR (robot Q) \cite{vptr2023} & Task Q-value & Image + Task + Action & Video TD pretraining + offline RL & Ego4D + multi-task robot data \\
CLIP-Motion (visual reward) \cite{clipmotion2023} & Motion stage / stage reward & Pair + Task & CLIP motion classification & Rollouts + state-derived motion labels \\
VICtoR \cite{victor2024} & Stage-aware motion progress & Image + Task & CLIP stage/progress learning & Action-free videos + instructions \\
REDS \cite{reds2025} & Subtask-conditioned reward & Video + Task & EPIC matching + CLIP alignment & Segmented expert/suboptimal demos \\
TeViR \cite{tevir2025} & Reached-reference stage & Image + Ref. + Hist. & CLIP-text-conditioned video matching & Task videos + generated references \\
Hume \cite{DBLP:journals/corr/abs-2505-21432} & Q-value & Image + Task + Action & Cal-QL (conservative TD) & Robot demos, sparse reward \\
T2-VLM$^{\ddagger}$ \cite{t2vlm2025} & Tracked subgoal reward & Image + Task + Hist. & VLM subgoals + Bayesian tracking & Frozen Qwen2-VL + SAM2 \\
Self-Improving EFM (steps-to-go) \cite{ghasemipour2025selfimproving} & Steps-to-go distribution & Pair / Multi-view + Task (+ Proprio.) & PaLI steps-to-go prediction + BC & Imitation data with hindsight step counts \\
SARM \cite{DBLP:journals/corr/abs-2509-25358} & Stage + progress & Video + Proprioceptive state & Stage/progress supervision & Process labels \\
RECAP (value function) \cite{physicalintelligence2025pistar} & Return distribution (steps to success) & Multi-view + Task + Proprio. & VLM-initialized distributional MC returns & Demos + labeled rollouts + interventions \\
PALM \cite{DBLP:journals/corr/abs-2601-07060} & Progress & Image + Task + Proprio. & Action--progress diffusion & Process-labeled demonstrations \\
Cosmos Policy \cite{DBLP:journals/corr/abs-2601-16163} & Return value & Multi-view + Task & Joint diffusion + MC returns & Demos + task rollouts \\
MARVL \cite{DBLP:journals/corr/abs-2602-15872} & Staged similarity & Image + Task + Goal & View-invariant CLIP tuning & Auto simulator frames \\
RISE$^{\ast}$ \cite{DBLP:journals/corr/abs-2602-11075} & Progress/TD value & Multi-view + Task + Future & Progress + TD value & Demos + success/failure rollouts \\
ViVa \cite{DBLP:journals/corr/abs-2604-08168} & Task value & Multi-view + Proprio. & Monte Carlo returns + dynamics & Task rollouts \\
Frozen-VLA value probes \cite{frozenvlaprobes2026} & Discounted outcome & Latent & Ridge regression on VLA features & Success/failure rollouts \\
SARM2 \cite{DBLP:journals/corr/abs-2606-10305} & Stage + value & Video + Multi-view + Task + Proprio. & Stage/value mixture & Primitive labels + rollouts \\
MV-WAM \cite{DBLP:journals/corr/abs-2606-21088} & Return value & Multi-view + Task + Action & MC return + joint dynamics & Robot demonstrations \\
DGAC \cite{DBLP:journals/corr/abs-2606-21406} & Value / candidate rank & Image + Action + Future & Value + future ranking & Human video + robot rollouts \\
STDR \cite{stdr2026} & Stage + within-stage progress & Video & Stage/progress + validity gates & VLM-segmented demos + negatives \\
Robo-ValueRL (value estimator) \cite{xia2026robovaluerl} & Normalized time-to-go value & Multi-view + Task + Hist. & Distributional remaining-time learning + failure penalty & Robot demos + online rollouts \\
WCM \cite{fei2026wcm} & History value & Video + Task & MC return + future latent & Demos + rollouts \\
RynnValue \cite{huang2026rynnvalue} & Temporal distance (absolute + signed) & Video + Task & Timestamp-distance learning & Embodied video \\
SeeQ (critic) \cite{seeq2026} & Subtask Q-value & Multi-view + Task + Action & Subtask supervision + TD & RoboCOIN + target-task data \\
eVTA (reward token) \cite{evta2026} & Expected future success & Video + Task & Terminal-outcome TD learning & Mixed-quality policy rollouts \\
MetaWAM \cite{liang2026metawam} & Return value & Multi-view + Task + Proprio. + Action & Joint diffusion + meta-frame value & Cosmos Policy RoboCasa data \\
\midrule
\rowcolor{famH!18}\multicolumn{5}{l}{\rrmfamily{HYBRID (24 methods)}} \\*
\addlinespace[3pt]
AHA \cite{DBLP:conf/iclr/DuanPKWTYKFMG25} & Sub-task verdict + failure explanation & Video + Multi-view + Task & LLaVA-1.5 instruction tuning & FailGen synthetic failures + VQA data \\
GRAPE (ranking reward) \cite{zhang2024grape} & Stage cost + self-likelihood + success & Traj. + Task + Latent & VLM stages + GPT-4o cost functions & No RRM training \\
VLAC \cite{DBLP:journals/corr/abs-2509-15937} & Relative progress + done & Pair + Task & Temporal-delta + done & Robot + human video \\
Guardian \cite{DBLP:journals/corr/abs-2512-01946} & Verdict + reasoning trace & Multi-view + Task & Failure-CoT tuning & Synthetic failures \\
ReWorld$^{\ast}$ \cite{DBLP:journals/corr/abs-2601-12428} & Hierarchical four aspects & Video + Task & Preference learning & GPT-4o-scored robot videos \\
Robometer \cite{DBLP:journals/corr/abs-2603-02115} & Progress + success + pref. & Video (pair) + Task & Multi-head temporal learning & Time + outcomes + auto prefs. \\
PRIMO-R1 \cite{DBLP:journals/corr/abs-2603-15600} & Progress + failure + CoT & Video + Task & Process SFT + RL & Process labels \\
Large Reward Models \cite{DBLP:journals/corr/abs-2603-16065} & Multi-part reward & Image (pair) + Task & Three LoRA interfaces (DPO + SFT) & Robot, human, and simulated demos (time labels) \\
PRM-as-a-Judge$^{\ddagger}$ \cite{DBLP:journals/corr/abs-2603-21669} & Outcome--process--diagnosis profile & Video + Task & Zero-shot PRM aggregation & No training \\
VLLR \cite{DBLP:journals/corr/abs-2604-00055} & Subgoal progress (critic init.) + self-certainty + outcome & Pair + Task + Graph + Latent & Prompted progress for critic warm-up + certainty & No RRM training (zero-shot LLM/VLM) \\
ARM \cite{DBLP:journals/corr/abs-2604-03037} & Advantage + completion & Video + Multi-view + Task & Dual-head classification & Human process labels \\
RoboAlign-R1$^{\ast}$ \cite{DBLP:journals/corr/abs-2605-03821} & Six scores & Video + Task & Teacher SFT + distillation & Rule + human rubric labels \\
Reward as an Agent$^{\ast}$ \cite{DBLP:journals/corr/abs-2606-19990} & 4-aspect scores + rationale & Video + Task & Prompted judging & Human audit \\
RoboGaze$^{\ast\ddagger}$ \cite{robogaze2026} & Localized failure events + severity & Video + Task + Init. & Grounded routing + specialist VLMs + critic & Frozen VLMs; no training \\
CoRe \cite{DBLP:journals/corr/abs-2607-01721} & Formal + residual reward & Image + Sim. + Action & Formal + preference residual & VLM labels \\
ACE-Brain-0.5 \cite{DBLP:journals/corr/abs-2607-04426} & Progress + preference & Video + Task & Process + preference learning & Labeled video \\
VLAC-Cut \cite{zhai2026help} & Signed progress + diagnosis + cut points & Video + Task (+ Ref.) & Process-annotation instruction tuning & Re-annotated robot video with failures and recoveries \\
Thea$^{\ddagger}$ \cite{wang2026thea} & Status + evidence + cause & Image + Criterion & Prompted post-hook & No training \\
TrustRobo\allowbreak{}Reward \cite{wang2026trustroboreward} & Scores / comparisons + rationale & Video (pair) + Task & Point/pair consistency & VLM + human correction \\
FIERCE \cite{fierce2026} & Progress + future failure risk & Video + Task + Proprio. + Action & Joint dynamics + reward learning & Robot data + outcome-labeled rollouts \\
EVO-WAM (verification)$^{\ast\ddagger}$ \cite{evowam2026} & Completion + action consistency & Multi-view + Video + Task + Init. + Proprio. + Action & VLM judging + inverse dynamics & Recorded video--action pairs \\
VeriFine-Judge \cite{verifine2026} & Rubric scores + diagnostics & Video + State + Task + Action + Rationale & Human-calibrated VLM distillation & Teacher scores + human corrections \\
Flora-Rewarder \cite{leng2026flora} & Episode grade (1--5) + step action quality + rubric scores & Video + Multi-view + Task & Full VLM fine-tuning with three prompt types & Human, recorded-success, and VLM-generated labels \\
RoboJudge-9B$^{\ast}$ \cite{robojudge2026} & Physical adherence + instruction alignment + rationale & Video + Task & Qwen3.5-9B SFT + GRPO & Human-verified scores + model-written rationales \\
\end{longtable}
\endgroup
\vspace{-1mm}
{\footnotesize\raggedright
\textit{Query evidence.} Image: one observation; Video: a temporal window;
Traj.: a complete trajectory (visual or structured state);
Pair: two observations or clips queried jointly; Multi-view: synchronized
camera views; Proprio.: proprioceptive state; Latent: policy-internal features
or distributions; Sim.: privileged simulator state; State: structured state
features; Future: a model-predicted observation. Task: instruction, task ID,
or evaluative question; Goal: goal image/configuration; Init.: initial observation;
Ref.: reference observations, demonstration, or support set;
Criterion: preference axis/post-condition; Graph: structured scene context;
Action: candidate action; Act.\,hist.: preceding actions;
Hist.: retained execution or reasoning context; Geometry: camera calibration/rays;
Rationale: the candidate policy's proposed reasoning.
A slash denotes alternatives; Pair is mandatory and (pair) is optional.
Task identity may be fixed in the fitted evaluator rather than supplied with
each query. Teacher inputs used only during training are omitted.\par
\smallskip
\textit{Component boundaries.}
Robo-Dopamine 2.0 is consolidated with the original method.
DecisionNCE-T and Rank2Reward's ranking head have Relative semantics;
the potential variant and density regularizer are separate constructions.
AnyviewMeter is a geometry-conditioned Robometer adaptation. FAVOR and
TeViR use generated references to assess actual execution. Small Stuff
records the observed-pair interface. HiRE uses fixed historical support
buffers; ARS requires complete rollouts. LLM-as-a-Verifier records the
robotics instantiation of a cross-domain protocol. VeriFine judges proposals and
reasoning rather than executed outcomes. VLLR's VLM progress only
initializes the critic; self-certainty and the sparse success reward drive
policy optimization.\par}

\begin{multicols}{2}[\section{Evaluation Details and Additional Results}%
\label{sec:supp-study}]
This section records the details behind Sections~\ref{sec:rrm-bench}
and~\ref{sec:experiments}. Section~\ref{sec:supp-settings} gives the
data sources, the annotation procedure, the input, scoring, and reporting rules, and the
settings fixed for each evaluated model, with the protocol of each
specialized evaluator (Table~\ref{tab:supp-roster}), the prompts of the
general-purpose VLMs, and the run conditions that differ between requests.
Sections~\ref{sec:supp-proxy}--\ref{sec:supp-metrics} give the per-model
values behind the findings of
Sections~\ref{sec:exp-proxy}--\ref{sec:exp-legacy}.

\subsection{Evaluation Settings}
\label{sec:supp-settings}

\begin{figure}[!tbp]
\centering
\includegraphics[width=0.96\linewidth]{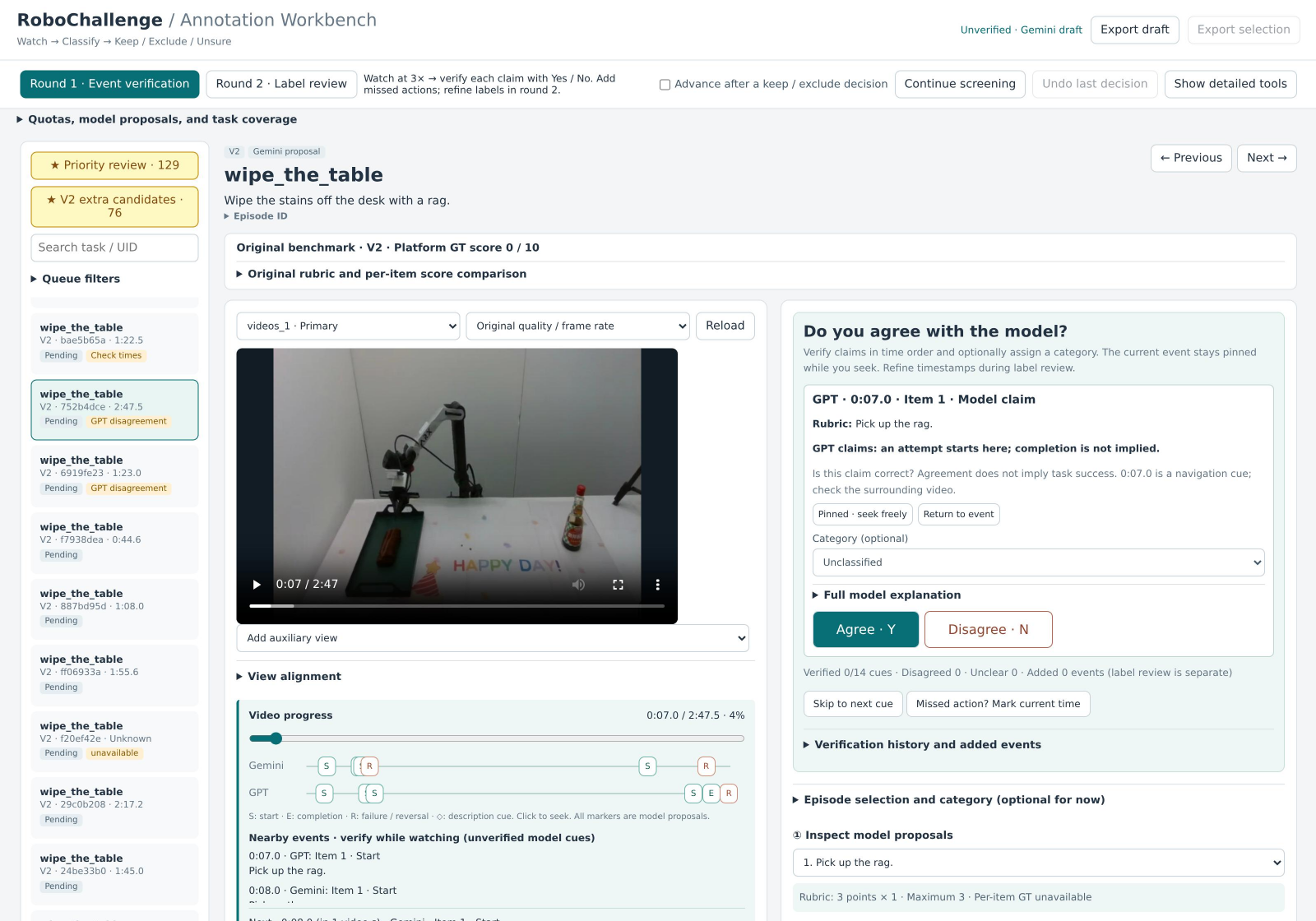}
\caption{Annotators verify model-proposed events against the video before
an execution enters the benchmark. In the annotation workbench, models
mark candidate events on the video timeline (S: start, E: completion, R:
failure or reversal). The annotator checks each claim in time order with
Agree or Disagree, may assign a diagnostic category, marks actions the
models missed, and then keeps, excludes, or flags the execution. Shown: an
execution of wipe the table at 7.0~s.}
\label{fig:supp-annotation}
\end{figure}

\emph{Data sources.} The RoboChallenge index lists 44{,}461 finished
rollouts on Table30 and Table30 V2, more than 120{,}000 camera videos. A
candidate pool of 1{,}500 rollouts, 750 per version and 4{,}228 videos, is
drawn across zero, partial, and full platform scores. Annotators screen this
pool, with a small set of added candidates, in the workbench of
Fig.~\ref{fig:supp-annotation} and annotate the executions they keep, whose
camera views and scope variants form the 1{,}117 videos.

\begin{figure}[!tbp]
\centering
\includegraphics[width=0.92\linewidth]{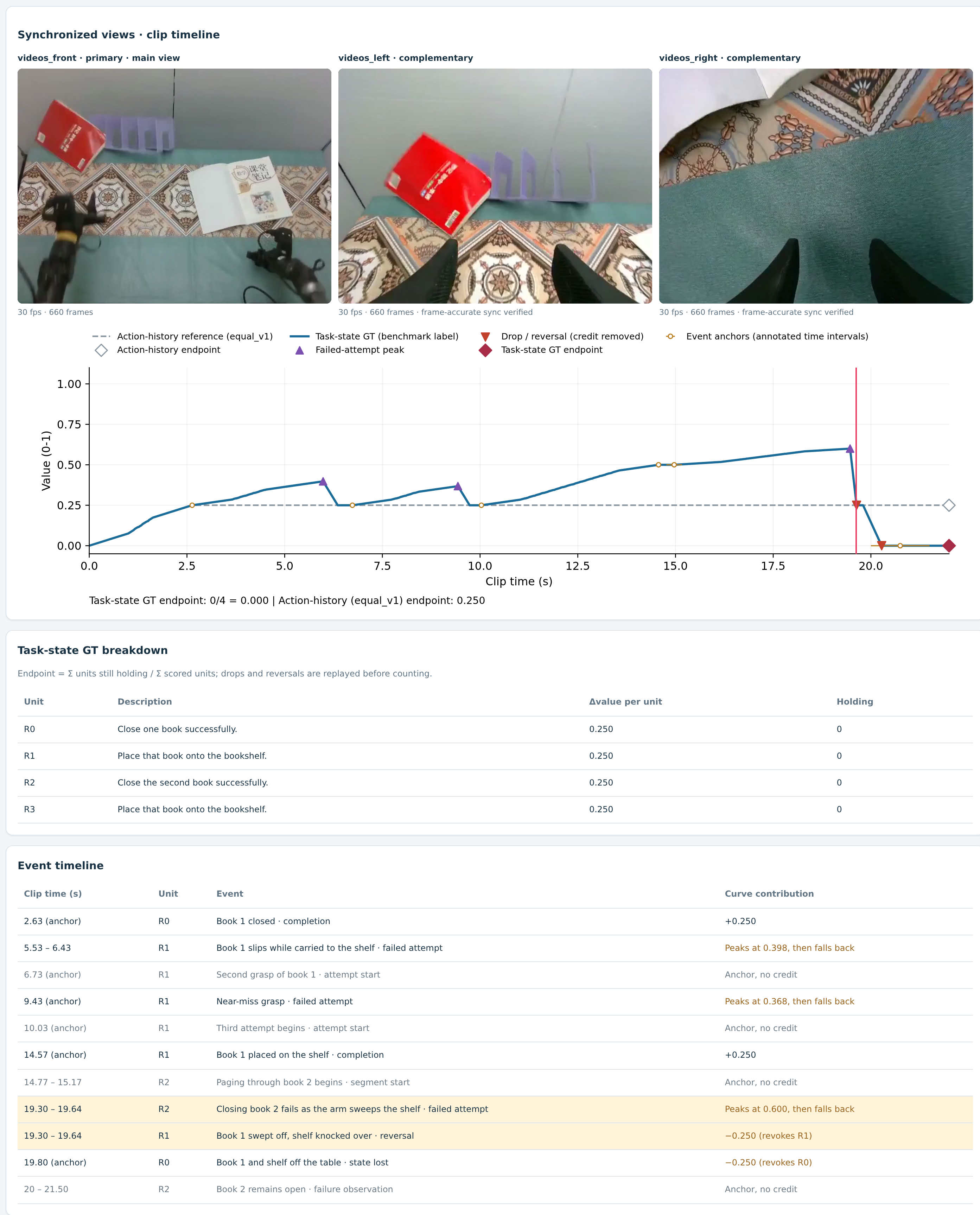}
\caption{The task-state label withdraws credit when an achieved condition
is lost, and every change traces to a time-stamped event linked to a task
unit. Synchronized views of the put-the-books-back execution of
Fig.~\ref{fig:bench-results} at 19.6~s, as the robot sweeps the shelved
book and the shelf off the table, above the compiled label (blue) and the
action-history reference (gray, dashed); below, the four task units and
the events behind the curve. Three failed attempts rise and fall back, and
the sweep revokes both completed units, so the label ends at 0 and the
reference at 0.25.}
\label{fig:supp-gt-page}
\end{figure}

\emph{Task-state labels.} Dense targets combine the human-reviewed events
of Section~\ref{sec:bench-reference} with rule-based state construction.
Annotators verify model-proposed events against the video and mark the
events the models miss (Fig.~\ref{fig:supp-annotation}).
Each event keeps its provenance (reviewed mark, approximate anchor, or
derived transition) and a timing-quality flag. The transient credit that
the compiler grants during an attempt is withdrawn if the attempt fails;
the provenance flag keeps such a rule-derived peak from being mistaken for
a physical reversal, which requires evidence that the condition was first
attained and then lost. Fig.~\ref{fig:supp-gt-page} shows the label of the execution of
Fig.~\ref{fig:bench-results} with its units and events. The normalized task-state score is an operational
score defined by the task rubric. Because the state label comes from the
views that show the decisive conditions, a placement that a wrist camera
misses keeps its label. The platform's rubric score is a separate target,
reported apart from the task-state label.

\emph{Inputs.} Sampling, image resolution, prompts, decoding, and score
transformations are fixed for each model. Each model receives its
documented evidence, including reference demonstrations or goal images
where required.

\emph{Scoring.} Unlike RoboPulse++~\cite{prmjudge15_2026}, which labels
annotated intervals, smooths continuous outputs, and drops steps near
interval boundaries, we score every non-flat transition of the task-state
reference without smoothing or boundary filtering. Because an unchanged
prediction counts as a miss for both classes, it lowers recall and
accuracy but not precision. VOC and $\tau_a$, defined in
Section~\ref{sec:bench-metrics}, are computed on native outputs with the
implementation of Robometer~\cite{DBLP:journals/corr/abs-2603-02115}.

\emph{Specialized evaluators.} The open-checkpoint roster comprises
ProcVLM~\cite{DBLP:journals/corr/abs-2605-08774},
VLAC~\cite{DBLP:journals/corr/abs-2509-15937} at 2B and 8B,
Robo-Dopamine~\cite{DBLP:journals/corr/abs-2512-23703} (GRM-3B, GRM-8B, and
the GRM-2.0 4B and 8B previews), Robo-ValueRL~\cite{xia2026robovaluerl},
Robometer~\cite{DBLP:journals/corr/abs-2603-02115},
RoboReward~\cite{DBLP:journals/corr/abs-2601-00675},
RynnValue~\cite{huang2026rynnvalue} at 4B and 8B,
PRIMO-R1~\cite{DBLP:journals/corr/abs-2603-15600}, Large Reward Models
(LRM)~\cite{DBLP:journals/corr/abs-2603-16065},
SOLE-R1~\cite{DBLP:journals/corr/abs-2603-28730},
TOPReward~\cite{DBLP:journals/corr/abs-2602-19313} applied to
Qwen3-VL-8B-Instruct~\cite{qwen2025qwen3vl}, VLAC-Cut~\cite{zhai2026help}, and
Flora-Rewarder~\cite{leng2026flora}.
GVL~\cite{DBLP:conf/iclr/MaHFSLXK0DXBJ0Z25} runs on the proprietary
Gemini 3.8 Flash~\cite{deepmind2026gemini38flash}.
Table~\ref{tab:supp-roster} gives the family, the input of one query,
the native output, and the source of the 1~FPS curve of each. For
models with several inference formulations, we evaluate RoboReward,
PRIMO-R1, and Flora-Rewarder on the episode endpoint and on successive
prefixes, and LRM through its completion, contrastive, and progress
interfaces. The open models receive single-view videos. Robo-ValueRL
requires several camera views at once and is scored only with multiview
input; Robo-Dopamine, SOLE-R1, and Flora-Rewarder are scored with both
(Section~\ref{sec:supp-evidence}).

Flora-Rewarder serves terminal, dense, and rubric prompts from one set of
weights. We use its terminal head, whose 1--5 episode grade $s$ gives the
progress value $(\mathbb{E}[s]-1)/4$ under the score-token probabilities,
and score every 1~FPS causal prefix as a complete episode with the
upstream frame selection, time metadata, and prompt. Its dense head grades
action quality rather than progress and its rubric head requires per-task
rubrics, so neither
is evaluated. Robo-Dopamine scores the change between
two observations, each filling three camera slots, against a start and a
goal image. A single view fills all three slots, with the video's first frame as the start and the official blank goal image, and the forward,
incremental, and backward estimates are averaged as in its paper. The
GRM-2.0 preview cards document this use; the GRM 1.0 cards show three
views with a goal image;
Section~\ref{sec:supp-evidence} reports all four checkpoints with the
other cameras in the remaining slots.

For direction tests, each model's native curve is differenced at 1~FPS;
curves saved only at sparse time points are interpolated and marked as
such, and specialized evaluators have no Pair-style result. LRM's contrastive interface compares the current frame with an
earlier one; we query it on adjacent source frames (about 30~FPS) and mark
these results accordingly. Terminal grades and LRM's completion interface
answer an outcome question and are not scored for direction.

\emph{General-purpose VLMs.} The VLMs share one system prompt and one user
template per query style, written after the method description of
RoboPulse++~\cite{prmjudge15_2026} and reproduced below. A request holds
the rendered template and then, for each observation, a label and its
image at native resolution; no timestamp, frame number, or camera name is
sent. In Sequence style, one request holds all 1~FPS observations of a
video in temporal order, a median of 15 and at most 110, and the model
returns a completion value from 0 to 100 for each. The predicted direction
of a transition is the sign of the difference between consecutive values,
and equal values are an unchanged prediction. In Pair style, the two
observations of each transition form their own request, labeled BEFORE and
AFTER and sent without earlier context, and the model answers $+1$ or
$-1$. Answers are parsed strictly, apart from format wrappers such as a
Markdown fence, and a request without a valid answer leaves its
transitions without a prediction. With three views
(Section~\ref{sec:supp-evidence}), each observation lists its three images
under the same templates. GVL keeps its own prompt: it receives the 1~FPS
frames in shuffled order with the first frame as a zero anchor, and its
scores are returned to temporal order.

\begin{rrmprompt}{System prompt, both styles}
\rrmpromptpar{You evaluate progress in robot manipulation tasks from visual observations.\newline
Use the given task instruction to identify the required physical goal and compare task-relevant states. Judge visible evidence rather than assuming that later observations must show more progress. Progress may increase, decrease, or remain unchanged, and the execution may end without success.\newline
Treat the task instruction as the goal to evaluate. Follow the requested output format exactly.}
\end{rrmprompt}
\begin{rrmprompt}{User prompt, Pair style}
\rrmpromptkeep\rrmpromptpar{Task instruction:\newline
\rrmtok{<task\_instruction>}\newline
\rrmtok{\$task\_instruction}\newline
\rrmtok{</task\_instruction>}}
\rrmpromptpar{You will receive two temporally adjacent observations, labeled BEFORE and AFTER. Images under the same observation label are simultaneous views, not additional time steps.}
\rrmpromptpar{Compare the physical task progress in AFTER with BEFORE. Consider the task-relevant objects, their relations to the goal, and any achieved state that is lost. Ignore appearance changes that do not affect the task. Do not use elapsed time alone as evidence of progress.}
\rrmpromptpar{Make a forced binary direction judgment:\newline
\rrmtok{+1}: AFTER is closer to satisfying the task than BEFORE.\newline
\rrmtok{-1}: AFTER is further from satisfying the task than BEFORE.\newline
If the visual evidence is ambiguous, choose the direction best supported by the observations. This binary baseline has no unchanged or abstention output.}
\rrmpromptkeep\rrmpromptpar{Return exactly one of the following, with no explanation:\newline
\rrmtok{<score>+1</score>}\newline
\rrmtok{<score>-1</score>}}
\end{rrmprompt}
\begin{rrmprompt}{User prompt, Sequence style}
\rrmpromptkeep\rrmpromptpar{Task instruction:\newline
\rrmtok{<task\_instruction>}\newline
\rrmtok{\$task\_instruction}\newline
\rrmtok{</task\_instruction>}}
\rrmpromptpar{You will receive \rrmtok{\$n\_observations} observations in chronological order, indexed from 0. Images under the same observation index are simultaneous views, not additional time steps.}
\rrmpromptpar{Estimate absolute physical task completion for every observation on a consistent scale from 0 to 100. Zero means no achieved task progress; 100 means that all requirements of the task are satisfied. Use intermediate values for partial completion.}
\rrmpromptpar{Base the scores on task-relevant physical states and the available temporal context. An achieved state that is undone can reduce the score. Unchanged states can receive equal scores. Do not force monotonic increases, do not force the first observation to 0, and do not force the last observation to 100. A clip can begin during an ongoing task and can end in failure. Elapsed time and robot motion alone do not establish completion.}
\rrmpromptpar{Return a JSON object with exactly one key, \rrmtok{"progress"}. Its value must be an array of exactly \rrmtok{\$n\_observations} finite numbers in [0, 100], in the supplied observation order. Return no Markdown or explanation.}
\end{rrmprompt}

GPT-6.1 Sol~\cite{openai2026gpt61sol} runs at low reasoning effort and
GPT-6 Luna~\cite{openai2026gpt6astra} and GPT-5.6
Terra~\cite{openai2026gpt56} at medium. Claude Opus
5.5~\cite{anthropic2026opus55}, Claude Sonnet
5.5~\cite{anthropic2026sonnet55}, and Claude Haiku
5.5~\cite{anthropic2026haiku55} run at medium effort and original image
scale, and Gemini 3.8 Flash runs
at its medium thinking level. A Sequence request takes a median of
8.8k--17.5k input tokens, depending on the VLM.

\emph{Decision model.} Jev~1.14~\cite{typesafe2026jev}, from TypeSafe AI,
returns probabilities over fixed answer options instead of text. It is
queried in both styles with the VLM task templates, and both styles use
the same frames, letterboxed to $280\times280$ pixels at JPEG quality 70.
In Sequence style, each video is sent once with all its 1~FPS observations
and one five-level completion question per observation (0, 25, 50, 75, or
100\% complete), and the expected level under the returned probabilities
is its progress value. In Pair style, it sees the two observations of each
transition and chooses between rise and fall; the more probable option is
its prediction.

\emph{RRM-Hub records.} Each adapter declares its source paper, the
official code it calls at a pinned commit, its weights, the camera views
it requires, its default frame sampling, and the type and range of its
native output, which is kept without rescaling. The benchmark labels give
each video's execution, task scope, camera view, evidence label, and
suites, so each result can be traced to the videos it covers. The licenses
of the wrapped code and weights are listed separately.

\emph{Reporting.} Table~\ref{tab:bench-main-results} reports
progress-direction classification for every evaluated model over 17{,}381
non-flat 1~FPS transitions, 1{,}702 of them falling; transitions
without a prediction, at most 1{,}323 per 1~FPS row, are counted
separately and excluded from its direction metrics. Within a model series, the larger checkpoint changes Macro-F1 by
$-0.007$ to $+0.027$ (Table~\ref{tab:bench-main-results}).
Table~\ref{tab:bench-controls} lists the 26 evaluators with a single-view
curve; Robo-ValueRL, LRM-8B (Contrastive), and the Pair-style queries
yield none.

\emph{Reward profiles.} Each panel of Fig.~\ref{fig:bench-results} and of
Figs.~\ref{fig:supp-profile-a}--\ref{fig:supp-profile-d} shows
the highest-VOC model (Table~\ref{tab:bench-controls}) of the group in its
subtitle at native timestamps, after a fixed unit conversion. Two
exceptions apply: the general-purpose VLM group shows its highest- and
lowest-VOC models, Claude Opus 5.5 and GPT-6 Luna, and the 4B panel shows
Robo-Dopamine-GRM-2.0-4B-Preview (third of four by VOC), the only 4B
model with dense outputs.

\subsection{Proxy Validity}
\label{sec:supp-proxy}

The first analysis covers the Action--Effect and Semantic/Ordered
Constraint suites and their controls, with Action--Effect executions scored on views that show the effect.

Endpoint judgments order executions more reliably than dense judgments
track change. In each of the 28 tasks whose executions differ in
final completion, we select the execution with the highest final score;
ties are averaged, and VLMs are queried in Sequence style. The selected
execution is the most complete one in 85\% of cases for GPT-6.1 Sol, 82\% for
GPT-5.6 Terra, and 75\% for Jev~1.14 and
Flora-Rewarder, against 44\% for a random choice, and TOPReward (38\%) falls
below chance. The 33
zero-completion executions of 11 tasks show how much endpoint credit
a failed execution receives. The share of them that reaches the median final
score of fully completed executions of the same task is 15\% for GPT-6.1
Sol and Claude Sonnet 5.5, 18\% for Flora-Rewarder, and 21\% for Robometer,
but 58\% for TOPReward, ProcVLM, and Robo-Dopamine-GRM-3B and 76\% for LRM-8B
(Progress), whose output rarely changes. Measured the same way on the 34
failed executions of the Action--Effect and Semantic/Ordered Constraint
suites, false credit is 3\% for Flora-Rewarder, 6\% for GPT-6.1 Sol, 9\% for
GPT-5.6 Terra and Robometer, and 12\% for Claude Opus 5.5 and Jev~1.14,
but 50\% for VLAC-Cut, 56\% for TOPReward, Robo-Dopamine-GRM-3B, and LRM-8B
(Progress), and 72\% for VLAC-8B.

Final scores also work as a filter. When the executions are ranked by final score and the top 10\% kept,
Robometer keeps only fully completed executions, RoboReward 93\%,
Flora-Rewarder and RynnValue-8B 90\%, and GPT-6.1 Sol 87\%, whereas a random
selection holds 47\%. RoboReward reaches this precision with the
lowest falling F1 among specialized evaluators (0.072). The GRM-2.0 previews
keep 50\%, and ProcVLM (45\%) and VLAC-8B (40\%) do no better than random.

Among the suites, semantic constraints are the hardest for dense
judgments (Table~\ref{tab:supp-suite-results}). On the transitions of the
Semantic/Ordered Constraint suite, 19 of the 26 evaluators score below an
always-rising predictor, more than on any other suite; the
Sequence-queried VLMs reach a falling F1 of at most 0.11 there, and the
best falling F1 is VLAC-Cut's 0.229. Even on the clean-success controls,
where nearly every predicted fall is an error, 21 of the 26 score below
the always-rising predictor.

General-purpose VLMs keep crediting placements at a wrong destination. On the 54 falls that follow a placement at a wrong destination
(Section~\ref{sec:supp-temporal}), no VLM row exceeds its overall rate of
predicting a fall by more than 0.04: Pair-queried VLMs recall 4--35\% of these falls and Sequence-queried
VLMs 0--6\%. On the 704 falls that follow a drop or a release failure, every
VLM row exceeds it by 0.10--0.36. Specialized evaluators
respond to wrong placements more often: Robometer and VLAC-2B exceed their
fall rate by 0.19, and ProcVLM, SOLE-R1, VLAC-Cut, and Flora-Rewarder by
0.10--0.15, although none recalls more than half of these falls.

\subsection{Temporal Validity}
\label{sec:supp-temporal}

The second analysis separates direction recognition from the response to
specific events. Rule-derived attempt peaks form a separate slice from
verified physical reversals, and an uncompleted final step counts as an
omission, not as the loss of a previously achieved state.

The falling class consists mostly of short local events. Of the 1{,}702
falling transitions of Table~\ref{tab:bench-main-results}, 863 (50.7\%) end a
failed attempt and 704 (41.4\%) follow a drop or a release failure. Another
54 (3.2\%) follow a placement at a wrong destination, and 15 follow an action
that completed without its effect. Only 66 (3.9\%) follow the loss of a
previously achieved condition, in seven executions.

Evaluators respond to these events in three ways. RoboReward and
LRM-8B (Progress) leave their output unchanged on 73\% of the
transitions. Most single-view specialized evaluators fluctuate instead: 15 of the 18
predict a fall on 13.9--44.3\% of the transitions, although only 9.8\% fall,
and 17 of the 18 have a falling precision of 0.111--0.209, 1.1--2.1 times the
base rate. On the transitions where the task state rises, the median
specialized evaluator predicts a fall 25\% of the time. Apart from the
near-constant RoboReward, this rate ranges from 7.6\% for VLAC-Cut to
44\% for TOPReward, with 34\% for ProcVLM and 31\% for SOLE-R1; Pair-queried
GPT-6.1 Sol predicts a fall on 8.0\% of these transitions. Sequence-queried
VLMs predict a fall on only 1.9--3.9\% of the
transitions, leave the score unchanged on 16--61\% of them, and recall
7.5--24.4\% of the falls. In Sequence queries, Jev~1.14 scores by an
expected level that varies almost continuously, so it leaves the score
unchanged on 4.3\% of the transitions; like most specialized
evaluators, it predicts a fall on 22.8\% of them, with a falling precision
of 0.183.

Switching from Pair to Sequence queries raises
falling precision and lowers falling recall for all seven VLMs. The query
protocol alone changes the result: Gemini 3.8 Flash reaches a Macro-F1 of
0.528 under the GVL protocol, 0.564 with Sequence queries, and 0.621 with
Pair queries, and the Pair and Sequence results of one VLM differ by up to
0.171 in either direction.
Jev~1.14 depends on the query as well: its Pair and Sequence answers agree
on only 65\% of the transitions, and unlike six of the seven VLMs it
scores higher with Sequence queries (Macro-F1 0.543 against 0.509), which
raise both its falling precision and its falling recall.

Fig.~\ref{fig:bench-results} shows these responses on one execution chosen from reversal-rich executions. GPT-6 Luna follows the shape of the reference at a
third of its height, and ProcVLM, SOLE-R1, and both Robo-Dopamine models
fluctuate while the task state rises. When the shelf falls, five of the
nine models drop to 0.05 or below, but PRIMO-R1 and
VLAC-Cut end higher than before and Flora-Rewarder stays near 0.5, although
all three have a higher VOC than Robo-Dopamine-GRM-3B (0.324), which drops.

\begin{figure}[!tbp]
\centering
\includegraphics[width=0.96\linewidth]{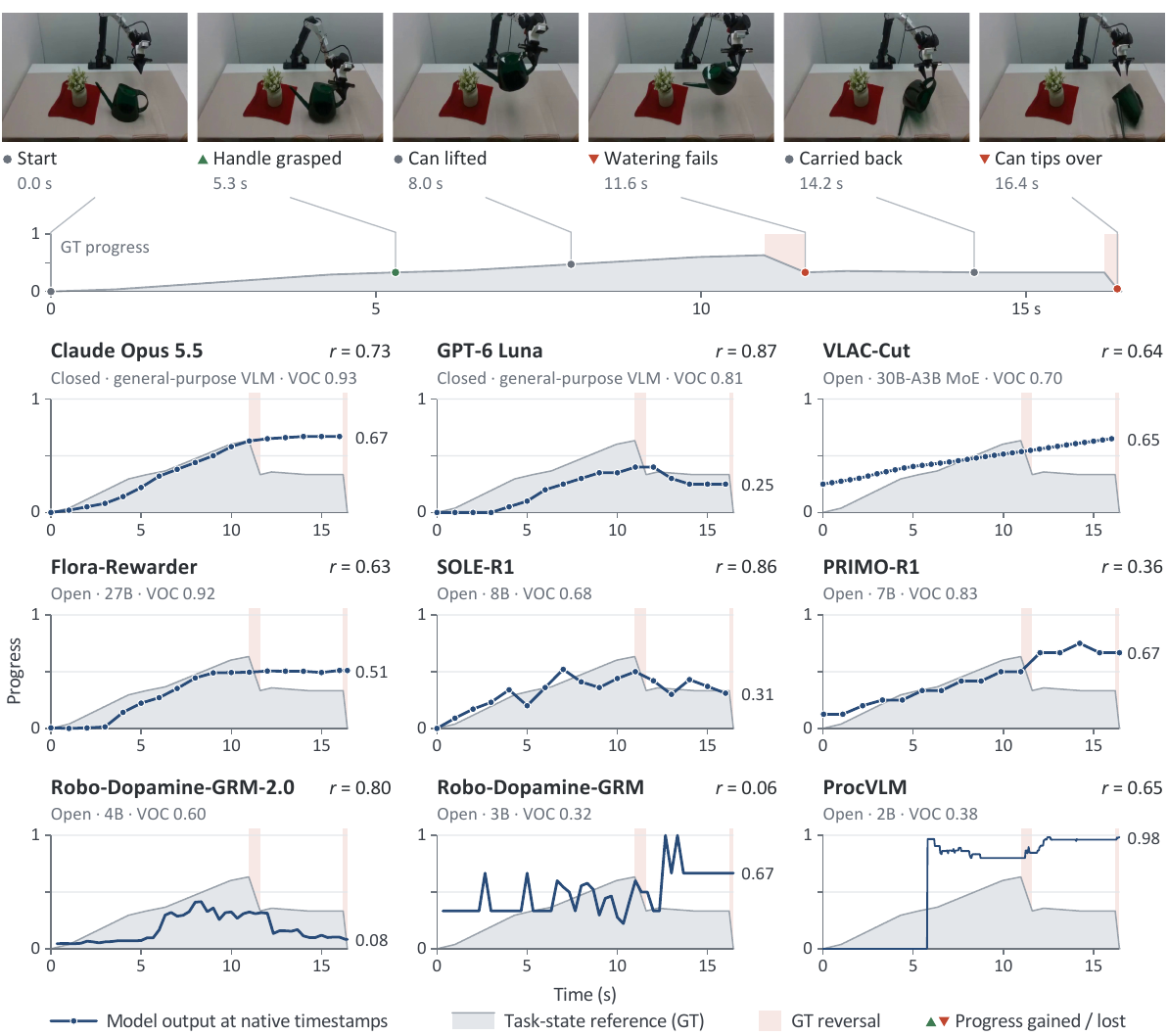}
\caption{Most evaluators credit the watering motion rather than its effect.
In this execution from the Action--Effect suite (A), the watering motion
ends at 11.6~s without watering the plant, and the can later tips over, so
the task-state reference (GT, gray) falls to zero. Claude Opus 5.5 and five
others keep or raise their credit after the failed watering; only GPT-6
Luna, SOLE-R1, and Robo-Dopamine-GRM-2.0 lower it. Layout as in
Fig.~\ref{fig:bench-results}.}
\label{fig:supp-profile-a}
\end{figure}

\begin{figure}[!tbp]
\centering
\includegraphics[width=0.96\linewidth]{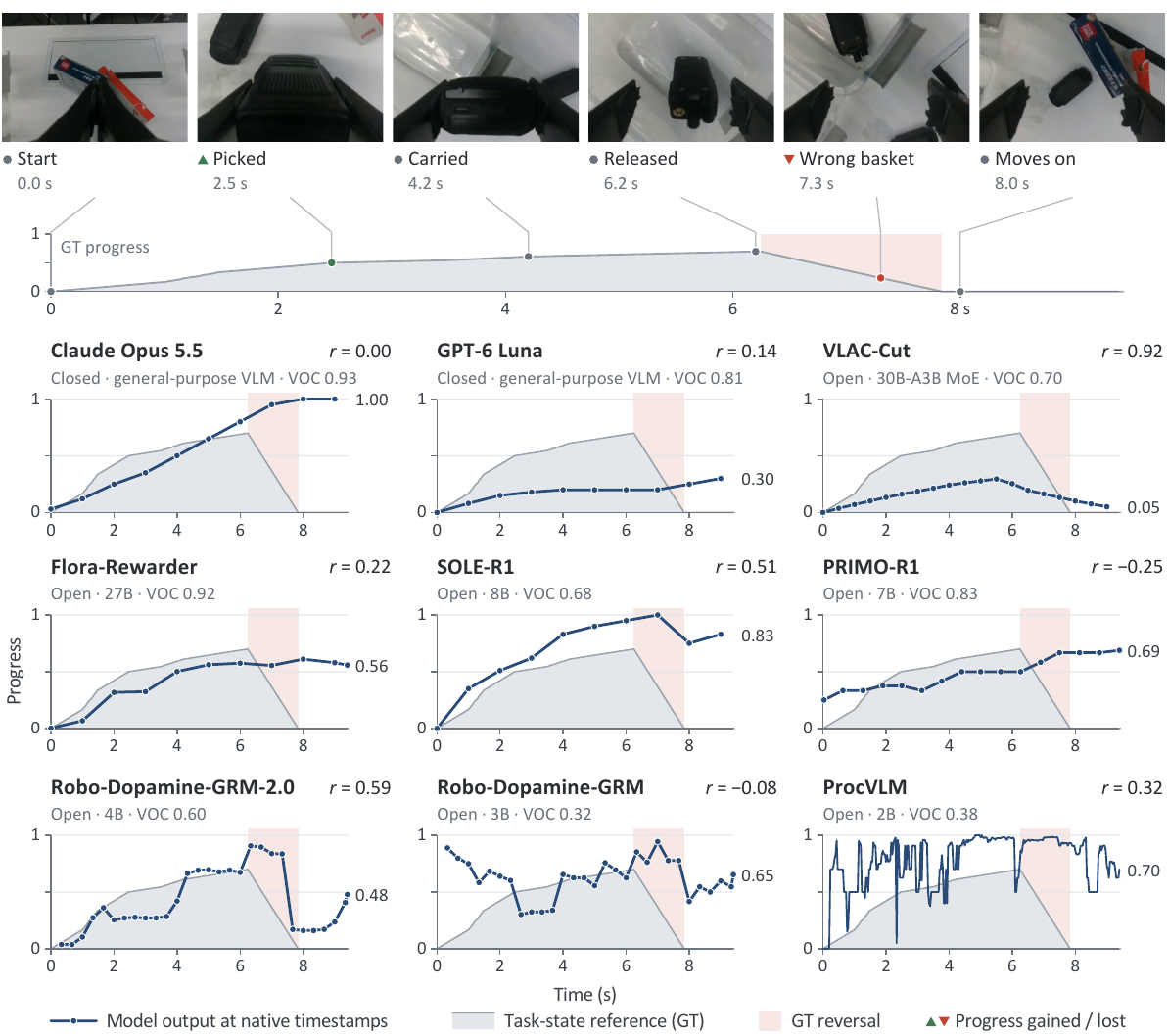}
\caption{Both general-purpose VLMs credit a placement in the wrong basket
as progress. In this execution from the Semantic/Ordered Constraint suite
(B), seen from the wrist camera, the robot releases the product into the
wrong basket, so the task-state reference returns to zero. Claude Opus 5.5
and GPT-6 Luna keep rising, Opus to full progress. Several specialized
evaluators dip at the release but rise again, and only VLAC-Cut ends near
zero.}
\label{fig:supp-profile-b}
\end{figure}

\begin{figure}[!tbp]
\centering
\includegraphics[width=0.96\linewidth]{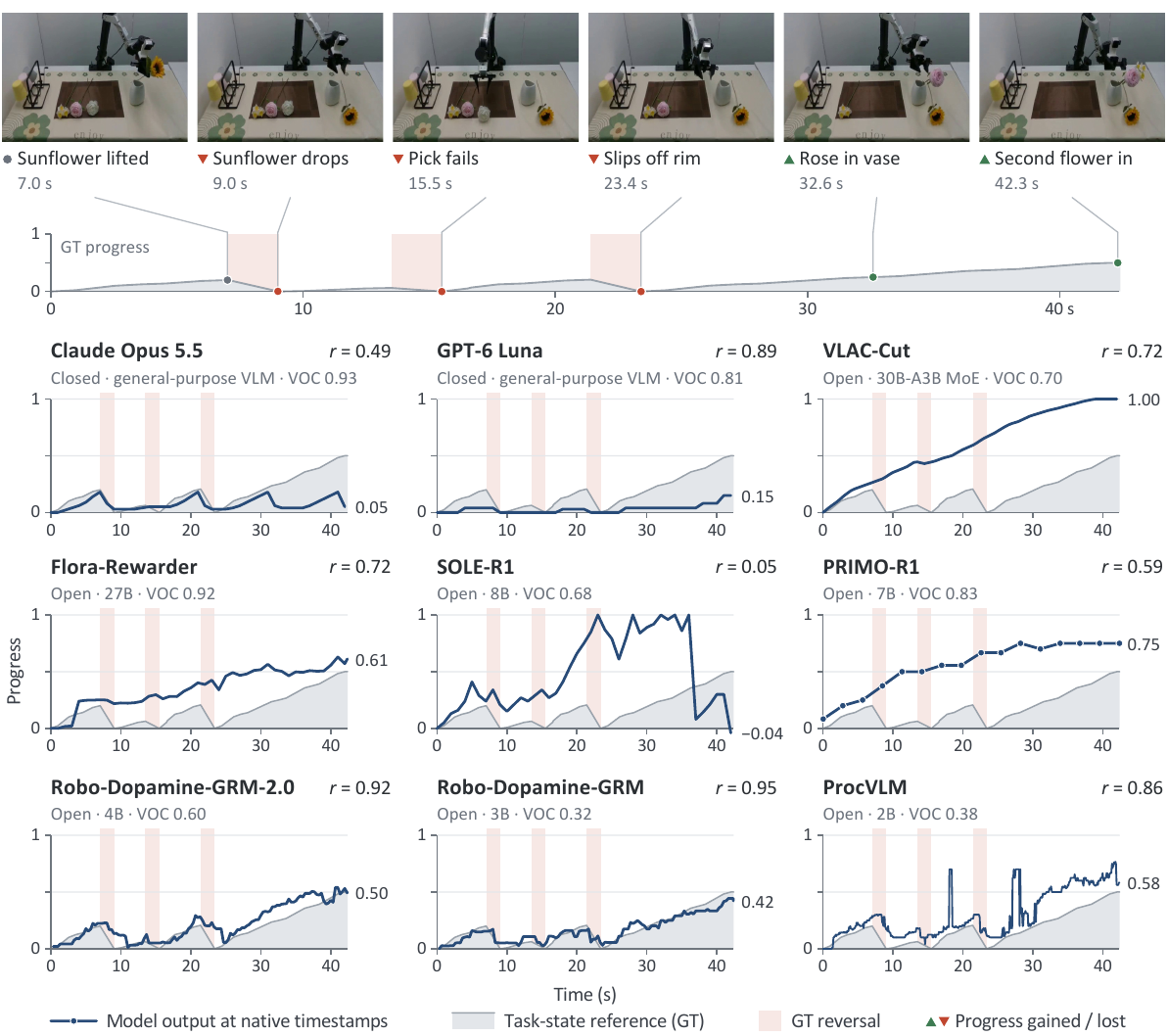}
\caption{Through repeated failures, only some evaluators follow the task
state. In this execution from the Persistence/Reversal suite (C), three
attempts fail before two flowers stay in the vase, so the task-state
reference ends at 0.5. Both Robo-Dopamine models withdraw their credit at
each failure and end near the reference, whereas VLAC-Cut and PRIMO-R1 rise
through every failure. Claude Opus 5.5 rises during each carry and falls
back when the carry ends, whether or not the flower stays, and so ends near
zero.}
\label{fig:supp-profile-c}
\end{figure}

Figs.~\ref{fig:supp-profile-a}--\ref{fig:supp-profile-c} show one more
execution from each of suites (A)--(C) in the same layout.

The task-state and action-history references end differently in 73
executions, and the task-state reference falls at least once in 122.
Where the two references disagree, the evaluators split. On the 922 transitions where the action history advances
but the task state does not, Pair-queried GPT-6.1 Sol,
Gemini 3.8 Flash, and Claude Opus 5.5 predict a rise 0.09--0.10 more often
than where both references are flat, so they follow the action history.
Flora-Rewarder, VLAC-Cut, and SOLE-R1 show no such difference ($-0.11$ to
$+0.01$). Instead, they predict rises at high rates throughout, including on
52\%, 72\%, and 51\% of the transitions where the task state falls;
Robometer does so on 74\%.

Responses to specific losses separate the evaluators further, again
measured against each row's overall rate of predicting a fall.
Sequence-queried VLMs recall 12--39\% of the 704 falls that follow a drop or
a release failure, although they predict a fall on only 1.9--3.9\% of
transitions. On the 66 falls that follow the loss of an achieved condition,
all 14 VLM rows and 15 of the 18 single-view specialized evaluators recall more than
their fall rate. Robometer recalls 55\% of them, 0.40 above its rate, and
GRM-2.0-8B-Preview 68\% (+0.35).
RoboReward recalls none, and VLAC-Cut, which leads the single-view
specialized evaluators in Macro-F1 and falling F1, recalls 15\%, 0.06
above its rate.

\subsection{Evidence Validity}
\label{sec:supp-evidence}

The third analysis pairs sufficient and reduced views of the same
execution and task scope (Table~\ref{tab:bench-evidence-results}). Of the
1{,}117 videos, 715 are labeled V, 374 P, and 28 X; the paired-evidence set
holds the 347 P/X videos with a V counterpart, all but six of them P, and
22 of the X videos, which show no task evidence, form a separate stress
condition.

\begin{figure}[!tbp]
\centering
\includegraphics[width=0.96\linewidth]{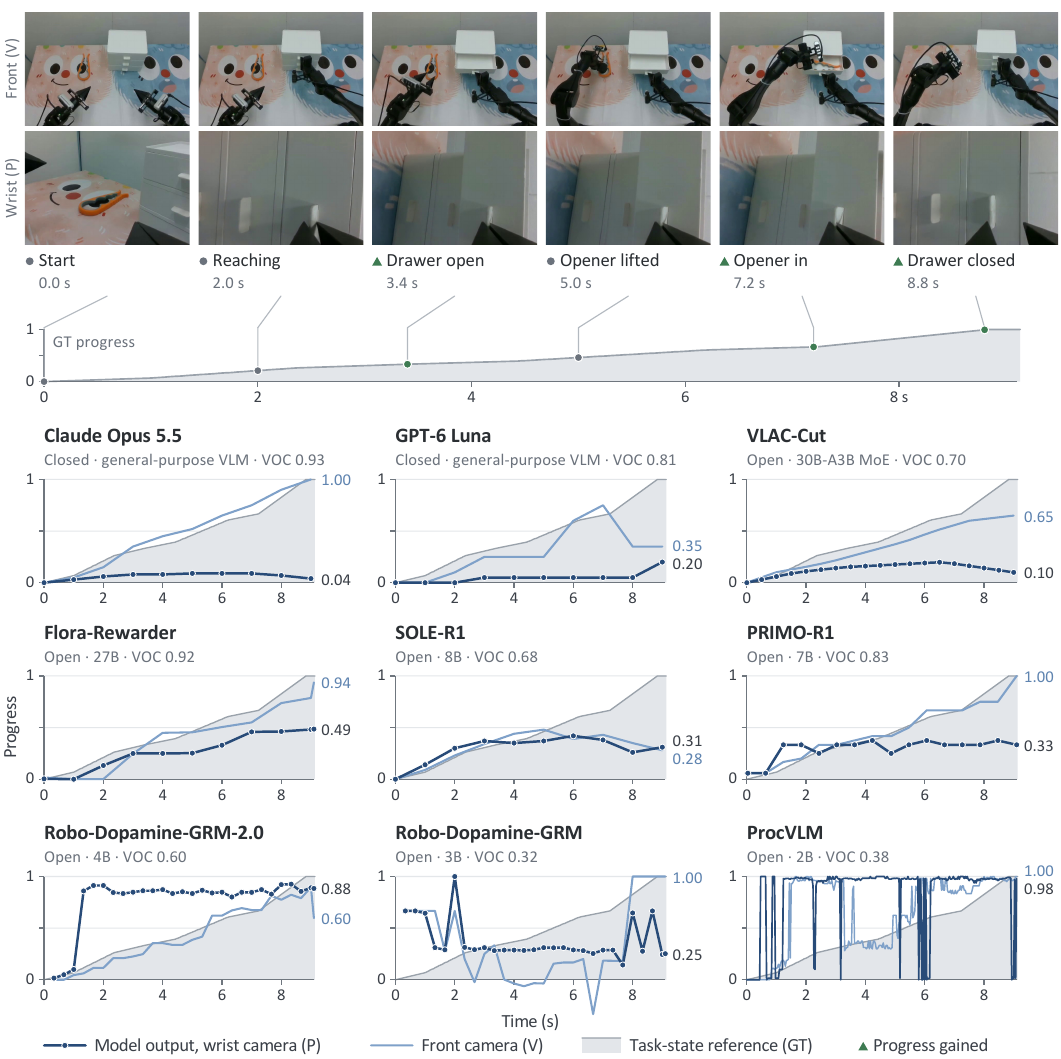}
\caption{Without the decisive view, most evaluators withhold the credit
they give the same execution when its outcome is visible. In this
successful execution from the Visibility suite (D), the robot opens a
drawer, places a bottle opener inside, and closes it; the wrist camera (P)
faces the drawer front and does not show the placement, whereas the front
camera (V) shows every step. Claude Opus 5.5 ends at 1.00 on the front view
but 0.04 on the wrist view, and PRIMO-R1 at 1.00 but 0.33;
Robo-Dopamine-GRM-2.0 instead ends higher on the wrist view. Layout as in
Fig.~\ref{fig:bench-results}, with each model's output on both views.}
\label{fig:supp-profile-d}
\end{figure}

Hiding the decisive evidence lowers Macro-F1 for 24 of the 26 single-view evaluators,
by at least 0.04 for 15 of them. The largest drops are 0.129 for
RynnValue-8B and GPT-6 Luna, 0.122 for RynnValue-4B, and 0.095 for GPT-5.6
Terra; Flora-Rewarder drops by 0.040. SOLE-R1 and LRM-8B (Progress) score
slightly higher on reduced views, the latter below 0.22 on both. The VLMs
and Jev~1.14 must answer every query, and specialized evaluators return a
score at every query, so no evaluated model can signal that the evidence is
insufficient, and answerability has no eligible model.
Fig.~\ref{fig:supp-profile-d} shows one successful execution under both
views: with the placement outside the wrist camera's view, most models end
far below the credit they give on the front view.

The 22 views without task evidence (194 transitions, 12 of them falling)
show how evaluators behave when the task is not visible at all. In Pair
queries, six of the seven VLMs still predict a rise on 55--99\% of these
transitions and Jev~1.14 on 96\%; Flora-Rewarder, VLAC-Cut, and SOLE-R1 do
so on 60--66\%. On these transitions the six VLMs reach a Macro-F1 of
0.41--0.53 and Jev~1.14 0.473, against 0.484 for a predictor that always
answers rising. In
Sequence queries, the VLMs return nearly constant curves, with at most 7.2\%
rises, which the direction test scores as errors because the task progresses
outside the view. Under the same query, Jev~1.14
predicts a rise on 59\% of these transitions, against 89\% on the V videos recorded alongside them, for a Macro-F1 of 0.415.

Adding views has the opposite effect for evaluators that can combine them.
On the 205 clips recorded by three cameras with verified synchronization,
from 111 executions (a clip is an execution or a scope variant with all
its camera views), the VLMs also receive the three views in one query,
scored on the same transitions and against the same reference as the three
videos. We compare the four VLMs whose single-view and multiview queries share an
access route. In Pair queries, GPT-6.1 Sol gains 0.092 Macro-F1 over each
clip's main view (from 0.635 to 0.727), which is a wrist camera in 129 of
the 205 clips. It also exceeds a majority vote over the three single views
by 0.048, so it combines the views. Its falling F1 rises the most, from
0.348 to 0.514. Sol gains 0.086 on clips whose three views are all labeled
V, and the gain persists without the 69 clips of arrange\_flowers. GPT-5.6
Terra gains 0.042 and Gemini 3.8 Flash 0.022 in Pair queries. In Sequence
queries, Sol gains 0.045, but for all four VLMs a majority vote over single
views exceeds the joint input.
Jev~1.14 does not gain in Sequence queries: its joint input scores 0.581,
against 0.584 for the main view and 0.593 for the vote. Seven clips exceed
its input limit as three separate images and are sent as one synchronized
three-view mosaic; the result holds without them.

Specialized evaluators that accept several cameras are scored on every
video recorded alongside other cameras ($^{\ddagger}$ in
Table~\ref{tab:bench-main-results}): the video's camera fills the primary
slot and the other cameras of its execution fill the remaining slots, so
the multiview and single-view results of a video share transitions and
reference. Of these 1,116 videos, 804 come from clips with three cameras,
189 of them with wrist cameras aligned by timestamp, and 312 from clips
with one fixed and one wrist camera, where Robo-Dopamine and Robo-ValueRL
repeat the primary view in the third slot. SOLE-R1 receives one external
and one wrist camera, and Flora-Rewarder the views tiled into one frame as
its documentation specifies. On the three-camera videos, the four
Robo-Dopamine checkpoints gain 0.072--0.101 Macro-F1 over the same single
view and SOLE-R1 gains 0.048, all with 95\% bootstrap intervals over
executions above zero, and their falling F1 rises by 0.045--0.092. The
Robo-Dopamine gains hold for V, P, and X views, for fixed and wrist
cameras, and in every suite. With a P or X view in the primary slot of a
clip with verified synchronization, all five exceed the single-view result
of a V view of the same clip, so the other cameras supply the evidence
that the primary view lacks. Flora-Rewarder does not gain ($-0.010$, with
an interval that includes zero); over all videos it loses 0.034 on V
views, which already show the decisive condition, holds on P views, and
gains on the 28 X views. Robo-ValueRL reaches a Macro-F1 of
0.416, against 0.474 for the always-rising predictor on the same
transitions. Because single-view use of Robo-Dopamine repeats one image in
slots built for different cameras, its gains also measure the cost of that
use.

The multiview judgment depends on which camera fills the primary slot. On
18 clips that were also rerun with identical input, swapping the primary
camera lowers the share of transitions with the same predicted direction
below that of the rerun for all seven evaluators, by 0.10 to 0.49. A
multiview result is therefore an average over choices of the primary
camera, not a property of the camera set.

\subsection{Task Scope and Metric Sensitivity}
\label{sec:supp-metrics}

Scope variants test whether the evaluator respects the declared scope of
completion. In 322 pairs from 132 scope variants of 39 executions, the
scope variant ends complete while the full task is not, with a median
full-task reference of 0.57 at that moment. Scored under its own
instruction, the scope variant receives more credit than the full task at
the same moment
in 68--90\% of pairs for the Sequence-queried VLMs, but in only 29--38\%
for the four Robo-Dopamine models, 39\% for VLAC-Cut, 45\% for SOLE-R1, and
17\% for LRM-8B (Progress), whose two scores tie in 47\% of pairs. For
these evaluators, completing the narrower instruction usually earns no
more than the credit of the unfinished task.

We then compare two conventional metrics, VOC and Kendall's $\tau_a$,
with falling F1 from the direction test (Table~\ref{tab:bench-controls}).
The normalized-time control never observes the video, yet its VOC of 1.000
exceeds that of every evaluator, while its $\tau_a$ and falling F1 are
zero: it assigns every execution the same final value and never predicts
a decrease. VOC therefore rewards chronology, not task state. Learned
evaluators show the same dissociation. RoboReward has the fourth-highest
VOC among specialized evaluators (0.807) and the lowest falling F1
(0.072), and Robometer and PRIMO-R1 follow the same pattern, with VOCs of
0.832 and 0.828 but falling F1 of only 0.214 and 0.154. Flora-Rewarder
leads the specialized evaluators on VOC (0.915) and $\tau_a$ (0.572) but
ranks second on falling F1 (0.252). Jev~1.14 has the third-highest VOC of
all evaluators (0.905) but ranks fifth on falling F1 (0.256) and seventh
on $\tau_a$ (0.496). In 52 of the 153 pairs of specialized evaluators, the
one with the higher VOC has the lower falling F1. VOC and falling F1 agree
more closely across the seven general-purpose VLMs (Spearman correlation
0.71, against 0.38 across the 18 specialized evaluators). On the executions alone, scored on one camera view each, the model rankings
are stable (Spearman correlation
0.95 for VOC and 0.83 for $\tau_a$ against the full set), and the
normalized-time control again attains a VOC of 1.000.

The direction test has its own trivial control. Because 90.2\% of the
transitions rise, a predictor that always answers ``rising'' reaches a
Macro-F1 of 0.474, above 9 of the 18 single-view specialized evaluators in
Table~\ref{tab:bench-main-results}, and the rising precision of every
single-view evaluator lies between 0.908 and 0.969; the falling class
carries the discrimination. Endpoint ordering and regression detection also
rank the evaluators differently. Among the specialized evaluators, $\tau_a$ and
falling F1 have a Spearman correlation of 0.45 on all videos and 0.32 on the executions; Flora-Rewarder, Robometer, and
RynnValue-4B lead on $\tau_a$, whereas VLAC-Cut, Flora-Rewarder, and
Robo-Dopamine-GRM-2.0-4B-Preview lead on falling F1.

Per-execution correlations show the same pull of elapsed time. On the 131 executions whose task-state reference is non-monotone or ends below completion,
the output of 17 of the 27 evaluators correlates more with elapsed time
than with the task state on most of them. The share is 75\% for Robometer and
69--70\% for RoboReward, PRIMO-R1, Flora-Rewarder, and Claude Haiku 5.5, all
with VOCs above 0.80, against 39\% for GPT-6.1 Sol and 37--39\% for the
GRM-2.0 previews; the median difference between the two correlations stays
within 0.05 for every evaluator.

\end{multicols}
\begin{table}[!t]
\centering
\caption{Specialized evaluators as run: family in
Table~\ref{tab:rrm-method-comparison}, input of one query, native output,
and source of the 1~FPS curve. Marks as in
Table~\ref{tab:bench-main-results}.}
\label{tab:supp-roster}
\footnotesize
\setlength{\tabcolsep}{4pt}
\renewcommand{\arraystretch}{1.12}
\begin{tabularx}{\textwidth}{@{}L{.23\textwidth}lYL{.245\textwidth}L{.125\textwidth}@{}}
\toprule
Evaluator & Family & Input of one query & Native output & 1~FPS curve \\
\midrule
ProcVLM-2B$^{\dagger}$
  & Absolute Progress
  & All frames (at most 512), eight-observation window
  & Progress, 0--100\%
  & Interpolated \\
VLAC-2B$^{\dagger}$, VLAC-8B$^{\dagger}$
  & Hybrid
  & 64 uniform frames, judged in adjacent pairs
  & Cumulative signed progress
  & Interpolated \\
Robo-Dopamine-GRM-3B, -8B,\newline
GRM-2.0-4B-Preview, -8B-Preview
  & Relative
  & Every tenth frame against the first frame and a blank goal image; the
    view fills all three camera slots
  & Relative progress, three modes averaged
  & Read at whole seconds \\
Robo-ValueRL$^{\ddagger}$
  & Temporal Value
  & Three views of every tenth frame, with one earlier frame of the base
    camera
  & Remaining steps $n$; value $1-n/5000$
  & Smoothed as in the official script \\
Robometer-4B$^{\dagger}$
  & Hybrid
  & Eight uniform frames in one pass
  & Progress in $[0,1]$ per frame
  & Interpolated \\
RoboReward-4B$^{\dagger}$
  & Outcome
  & 16 prefixes, each scored as an episode
  & Grade $s\in\{1,\dots,5\}$; $(s-1)/4$
  & Interpolated \\
RynnValue-4B$^{\dagger}$, -8B$^{\dagger}$
  & Temporal Value
  & Eight uniform frames in one pass
  & Negative remaining time in seconds
  & Interpolated \\
PRIMO-R1-7B$^{\dagger}$
  & Hybrid
  & 16 prefixes of 32 frames each
  & Progress, 0--100\%
  & Interpolated \\
LRM-8B (Contrastive)$^{\S}$
  & Hybrid
  & Two adjacent source frames
  & Signed change in $\{-1,0,+1\}$
  & None \\
LRM-8B (Progress)
  & Hybrid
  & One source frame
  & Progress in $[0,1]$
  & Read at whole seconds \\
SOLE-R1-8B
  & Absolute Progress
  & Frames at 1~FPS
  & Progress, 0--100\%, with a rationale
  & Native \\
TOPReward (Qwen3-VL-8B)$^{\dagger}$
  & Absolute Progress
  & 64 uniform frames, 15 prefixes, in Robometer's implementation
  & Log-probability of the answer token, min--max normalized
  & Interpolated \\
Flora-Rewarder-27B
  & Hybrid
  & Each 1~FPS prefix, scored as an episode
  & Grade $s\in\{1,\dots,5\}$; $(\mathbb{E}[s]-1)/4$
  & Native \\
VLAC-Cut-30B-A3B
  & Hybrid
  & Frames at 2~FPS in one pass
  & Keypoints of time and progress in \%
  & Read at whole seconds \\
GVL (Gemini 3.8 Flash)
  & Absolute Progress
  & All 1~FPS frames in shuffled order, with the first frame as a 0\% anchor
  & Completion, 0--100\%, per frame
  & Native \\
\bottomrule
\end{tabularx}
\end{table}

\begin{table}[!t]
\centering
\caption{Direction results by contrast suite and control group: Macro-F1 (M)
and falling F1 (F). Gray: Macro-F1 below the always-rising predictor on the
same transitions. Bold: best in each panel. $^{\dagger}$Curve interpolated
to 1~FPS.}
\label{tab:supp-suite-results}
\footnotesize
\setlength{\tabcolsep}{3pt}
\providecommand{\rrmbelow}[1]{\textcolor{black!45}{#1}}
\begin{tabularx}{\textwidth}{@{}l*{12}{>{\centering\arraybackslash}X}@{}}
\toprule
& \multicolumn{2}{c}{(A)} & \multicolumn{2}{c}{(B)} & \multicolumn{2}{c}{(C)} & \multicolumn{2}{c}{(D)} & \multicolumn{2}{c}{Control} & \multicolumn{2}{c}{Control} \\
& \multicolumn{2}{c}{Action--Effect} & \multicolumn{2}{c}{Semantic/Ordered} & \multicolumn{2}{c}{Persistence/Reversal} & \multicolumn{2}{c}{Visibility} & \multicolumn{2}{c}{Clean success} & \multicolumn{2}{c}{Obvious failure} \\
\cmidrule(lr){2-3}\cmidrule(lr){4-5}\cmidrule(lr){6-7}\cmidrule(lr){8-9}\cmidrule(lr){10-11}\cmidrule(lr){12-13}
Model & M & F & M & F & M & F & M & F & M & F & M & F \\
\midrule
\rowcolor{black!6}\multicolumn{13}{l}{\textbf{\textit{Specialized evaluators, 2B--4B}}} \\
ProcVLM-2B$^{\dagger}$ & \rrmbelow{0.459} & 0.247 & \rrmbelow{0.407} & 0.145 & 0.469 & 0.257 & \rrmbelow{0.411} & 0.168 & \rrmbelow{0.369} & 0.014 & \textbf{0.572} & \textbf{0.487} \\
VLAC-2B$^{\dagger}$ & \rrmbelow{0.241} & 0.146 & \rrmbelow{0.269} & 0.136 & \rrmbelow{0.395} & 0.212 & \rrmbelow{0.322} & 0.162 & \rrmbelow{0.352} & 0.025 & \rrmbelow{0.269} & 0.337 \\
Robo-Dopamine-GRM-3B & \rrmbelow{0.372} & 0.203 & \rrmbelow{0.374} & 0.145 & \rrmbelow{0.423} & 0.263 & \rrmbelow{0.351} & 0.170 & \rrmbelow{0.266} & 0.019 & \rrmbelow{0.342} & 0.344 \\
Robo-Dopamine-GRM-2.0-4B-Preview & 0.476 & 0.235 & \rrmbelow{0.440} & \textbf{0.169} & 0.518 & \textbf{0.312} & \rrmbelow{0.452} & \textbf{0.212} & \rrmbelow{0.398} & 0.033 & 0.544 & 0.467 \\
Robometer-4B$^{\dagger}$ & \textbf{0.519} & 0.144 & \textbf{0.515} & 0.147 & \textbf{0.572} & 0.262 & \textbf{0.529} & 0.205 & \textbf{0.507} & \textbf{0.067} & 0.556 & 0.306 \\
RoboReward-4B$^{\dagger}$ & \rrmbelow{0.283} & 0.107 & \rrmbelow{0.205} & 0.038 & \rrmbelow{0.244} & 0.082 & \rrmbelow{0.195} & 0.052 & \rrmbelow{0.232} & 0.000 & \rrmbelow{0.196} & 0.000 \\
RynnValue-4B$^{\dagger}$ & 0.494 & \textbf{0.264} & \rrmbelow{0.460} & 0.161 & 0.542 & 0.299 & \rrmbelow{0.456} & 0.187 & \rrmbelow{0.434} & 0.027 & 0.567 & 0.429 \\
\rowcolor{black!6}\multicolumn{13}{l}{\textbf{\textit{Specialized evaluators, 7B--8B}}} \\
PRIMO-R1-7B$^{\dagger}$ & \rrmbelow{0.431} & 0.154 & \rrmbelow{0.428} & 0.144 & \rrmbelow{0.428} & 0.171 & \rrmbelow{0.424} & 0.157 & \rrmbelow{0.395} & \textbf{0.045} & \rrmbelow{0.383} & 0.174 \\
LRM-8B (Progress) & \rrmbelow{0.237} & 0.153 & \rrmbelow{0.176} & 0.081 & \rrmbelow{0.206} & 0.175 & \rrmbelow{0.213} & 0.138 & \rrmbelow{0.151} & 0.014 & \rrmbelow{0.319} & 0.235 \\
Robo-Dopamine-GRM-8B & \rrmbelow{0.402} & 0.212 & \rrmbelow{0.389} & 0.152 & \rrmbelow{0.445} & 0.275 & \rrmbelow{0.386} & 0.187 & \rrmbelow{0.307} & 0.019 & \rrmbelow{0.378} & 0.396 \\
Robo-Dopamine-GRM-2.0-8B-Preview & 0.481 & 0.246 & \rrmbelow{0.436} & 0.158 & 0.504 & 0.289 & \rrmbelow{0.448} & 0.201 & \rrmbelow{0.400} & 0.029 & 0.532 & 0.460 \\
RynnValue-8B$^{\dagger}$ & 0.526 & \textbf{0.265} & \rrmbelow{0.445} & 0.162 & \textbf{0.546} & \textbf{0.296} & \rrmbelow{0.449} & 0.188 & \rrmbelow{\textbf{0.436}} & 0.023 & 0.542 & 0.417 \\
SOLE-R1-8B & \textbf{0.532} & 0.257 & \textbf{0.480} & \textbf{0.170} & 0.523 & 0.278 & \textbf{0.498} & \textbf{0.208} & \rrmbelow{0.421} & 0.028 & \textbf{0.594} & \textbf{0.483} \\
TOPReward (Qwen3-VL-8B)$^{\dagger}$ & \rrmbelow{0.445} & 0.206 & \rrmbelow{0.406} & 0.122 & \rrmbelow{0.461} & 0.237 & \rrmbelow{0.427} & 0.169 & \rrmbelow{0.367} & 0.021 & 0.477 & 0.391 \\
VLAC-8B$^{\dagger}$ & \rrmbelow{0.293} & 0.163 & \rrmbelow{0.248} & 0.113 & \rrmbelow{0.426} & 0.212 & \rrmbelow{0.348} & 0.153 & \rrmbelow{0.384} & 0.034 & \rrmbelow{0.381} & 0.440 \\
\rowcolor{black!6}\multicolumn{13}{l}{\textbf{\textit{Specialized evaluators, 27B--30B}}} \\
Flora-Rewarder-27B & 0.555 & \textbf{0.274} & 0.501 & 0.185 & 0.568 & 0.317 & 0.511 & \textbf{0.208} & \rrmbelow{0.459} & 0.037 & \textbf{0.605} & \textbf{0.478} \\
VLAC-Cut-30B-A3B & \textbf{0.568} & 0.232 & \textbf{0.568} & \textbf{0.229} & \textbf{0.623} & \textbf{0.335} & \textbf{0.561} & 0.205 & \textbf{0.506} & \textbf{0.045} & 0.414 & 0.034 \\
\rowcolor{black!6}\multicolumn{13}{l}{\textbf{\textit{Specialized evaluators, proprietary backbone}}} \\
GVL (Gemini 3.8 Flash) & 0.532 & 0.253 & 0.486 & 0.187 & 0.556 & 0.295 & 0.506 & 0.227 & \rrmbelow{0.449} & 0.013 & 0.602 & 0.476 \\
\rowcolor{black!6}\multicolumn{13}{l}{\textbf{\textit{General-purpose VLMs, Sequence style}}} \\
GPT-6.1 Sol & 0.561 & \textbf{0.352} & \rrmbelow{0.412} & \textbf{0.114} & 0.589 & \textbf{0.438} & 0.494 & \textbf{0.291} & 0.556 & 0.308 & 0.410 & 0.105 \\
GPT-6 Luna & \rrmbelow{0.425} & 0.198 & \rrmbelow{0.276} & 0.073 & \rrmbelow{0.446} & 0.308 & \rrmbelow{0.343} & 0.182 & \rrmbelow{0.413} & 0.147 & \rrmbelow{0.400} & 0.140 \\
GPT-5.6 Terra & \rrmbelow{0.421} & 0.140 & \rrmbelow{0.254} & 0.007 & \rrmbelow{0.402} & 0.271 & \rrmbelow{0.302} & 0.136 & \rrmbelow{0.362} & 0.087 & \rrmbelow{0.346} & 0.073 \\
Claude Opus 5.5 & \textbf{0.617} & 0.333 & \textbf{0.492} & 0.112 & \textbf{0.655} & 0.419 & \textbf{0.563} & 0.243 & \textbf{0.651} & \textbf{0.357} & \textbf{0.472} & \textbf{0.172} \\
Claude Sonnet 5.5 & 0.530 & 0.204 & \rrmbelow{0.416} & 0.042 & 0.526 & 0.221 & 0.479 & 0.137 & \rrmbelow{0.469} & 0.000 & 0.432 & 0.107 \\
Claude Haiku 5.5 & \rrmbelow{0.441} & 0.103 & \rrmbelow{0.354} & 0.034 & \rrmbelow{0.393} & 0.151 & \rrmbelow{0.372} & 0.084 & \rrmbelow{0.408} & 0.030 & 0.462 & 0.140 \\
Gemini 3.8 Flash & 0.503 & 0.194 & \rrmbelow{0.401} & 0.031 & 0.610 & 0.386 & 0.533 & 0.268 & 0.556 & 0.222 & 0.407 & 0.037 \\
\rowcolor{black!6}\multicolumn{13}{l}{\textbf{\textit{Decision model, Sequence style}}} \\
Jev~1.14 & 0.551 & 0.253 & 0.496 & 0.159 & 0.566 & 0.318 & 0.525 & 0.232 & \rrmbelow{0.457} & 0.041 & 0.564 & 0.412 \\
\bottomrule
\end{tabularx}
\end{table}

\end{document}